\documentclass{article} 
\usepackage{Aux/iclr2027_conference,times}

\usepackage{amsmath,amsfonts,bm}

\def\eqref#1{equation~\ref{#1}}

\def\1{\bm{1}}

\DeclareMathAlphabet{\mathsfit}{\encodingdefault}{\sfdefault}{m}{sl}
\SetMathAlphabet{\mathsfit}{bold}{\encodingdefault}{\sfdefault}{bx}{n}

\usepackage{graphicx}
\usepackage{svg}
\usepackage{overpic}
\usepackage{float}
\usepackage{wrapfig}
\usepackage[most]{tcolorbox}
\usepackage{subcaption}

\usepackage{multirow}
\tcbuselibrary{skins, breakable}

\usepackage{adjustbox}
\usepackage{booktabs}
\usepackage{refcount}

\usepackage[most]{tcolorbox}
\usepackage{thm-restate}
\usepackage{amsthm}

\usepackage{hyperref}
\usepackage{array}
\usepackage{booktabs}
\usepackage{longtable}
\usepackage{makecell}
\usepackage{tabularx}
\usepackage[table]{xcolor}
\usepackage[svgnames]{xcolor}

\usepackage{booktabs}

\definecolor{OIblue}{HTML}{0072B2}
\definecolor{OIorange}{HTML}{E69F00}
\definecolor{OIgreen}{HTML}{009E73}

\newcolumntype{B}{>{\columncolor{OIblue!10}}c}
\newcolumntype{O}{>{\columncolor{OIorange!12}}c}
\newcolumntype{G}{>{\columncolor{OIgreen!10}}c}

\colorlet{blue}{black}

\usepackage{tikz}
\usetikzlibrary{backgrounds, calc, fit, positioning, shapes.geometric, decorations.pathmorphing, decorations.pathreplacing, arrows.meta}
\usepackage{tikz-cd}

\usepackage{extpfeil}
\usepackage{url}
\usepackage{comment}
\usepackage{parskip}
\usepackage{pifont}
\usepackage{mathrsfs}
\usepackage{scalerel}
\usepackage{makecell}

\usepackage{enumitem}

\newcommand{\repeatpropositionname}{Proposition}

\newtheorem*{repeatpropositioninner}{\repeatpropositionname}

\newtheorem{theorem}{Theorem}

\newtheorem{proposition}[theorem]{Proposition}
\newtheorem{definition}[theorem]{Definition}
\newtheorem{corollary}[theorem]{Corollary}
\theoremstyle{remark}
\newtheorem{remark}{Remark}
\newtheorem{example}{Example}
\newtheorem{subexample}{Subexample}
\theoremstyle{plain}

\newcommand{\inarrow}[1]{%
  \mathrel{\tikz[baseline=-0.55ex]{%
    \draw[->] (0,0) -- (1.25,0)
      node[midway,fill=white,inner sep=0.5pt] {$\scriptstyle #1$};
  }}%
}

\newcounter{casestudy}

\newcommand{\casestudyparagraph}[1]{%
  \refstepcounter{casestudy}%
  \paragraph{Case Study \thecasestudy: #1}%
}

\definecolor{CaseLink}{HTML}{0057B8}

\newcommand{\caseref}[1]{%
  \hyperref[#1]{\textcolor{CaseLink}{\ref*{#1}}}%
}

\title{\centering Symmetry Slippage in Equivariant ML: \\ Creation, Computation, and Corrections}

\author{
\normalfont
\begin{tabular}{@{}c@{\hspace{1em}}c@{\hspace{1em}}c@{}}
\textbf{Jacob Hume} &
\textbf{Jacob Bamberger} &
\textbf{Michael M. Bronstein} \\
Mathematical Institute &
Department of Computer Science &
Department of Computer Science \\
University of Oxford &
University of Oxford &
University of Oxford \\
Oxford, UK & Oxford, UK & AITHYRA \\
\texttt{jmh274@cantab.ac.uk} & &
\end{tabular}
}

\iclrfinalcopy 
\begin{document}

\maketitle
\lhead{}

\begin{abstract}


Symmetry-based inductive biases are prevalent throughout machine learning for science and beyond. Often, the symmetries realized by an architecture differ from those prescribed. This mismatch can occur in two ways: when prescribed symmetries are not realized (\emph{under-equivariance}) and when \emph{non}-prescribed symmetries \emph{are} realized; we call this \emph{symmetry slippage}. Although famous examples of symmetry slippage abound, a systematic study of its origin, estimation, and corrections has yet to exist. This work aims to initiate such a study. We formulate symmetry slippage in terms of subarchitectural decompositions, and use this framing to identify three common mechanisms by which contemporary techniques can unintentionally create it. These creation mechanisms recover the familiar examples, theoretically ground popular heuristics, and identify symmetry slippage across nine active research areas.
While we find that guaranteeing an absence of symmetry slippage is hard in general, we prove conditions under which our decomposition framing is exact. We use this result to exactly compute, for the first time, the realized symmetries of a classical family subsuming e.g. convolutional architectures. Finally, we study \emph{grippage encodings} (GEs), simple symmetry-correcting features that restore expressivity. We find that GEs improve empirical performance across our nine case studies.

\end{abstract}

\section{Introduction}

The ubiquity of symmetry in science
 motivates machine learning architectures $\mathscr{A}$ that leverage obedience to a prescribed transformation law $\mathrm{A}$ as inductive bias~\citep{bronstein2021geometric, weber2025geometric}. Precisely, $\mathscr{A}$ is designed so that the transformation law it realizes, denoted $\operatorname{Sym}\mathscr{A}$, satisfies $\mathrm{A}=\operatorname{Sym}\mathscr{A}$. Equality requires two conditions: that every prescribed symmetry is realized ($\operatorname{A} \subset \operatorname{Sym}\mathscr{A}$), and that every realized symmetry was prescribed $(\mathrm{A} \supset \operatorname{Sym}\mathscr{A})$.  The first condition, $\operatorname{A} \subset \operatorname{Sym}\mathscr{A}$, is known as \emph{equivariance} of $\mathscr{A}$ with respect to $\mathrm{A}$; its enforcement bedrocks the eponymous \emph{equivariant machine learning} literature and Geometric Deep Learning \citep{bronstein2021geometric} more broadly. The other condition, $\mathrm{A} \supset \operatorname{Sym}\mathscr{A}$, is subtler but relevant, with famous failure cases in fields ranging from molecular learning~\citep{dumitrescu2025eequivariant, weaver2025chi} to the theory of graph neural networks~\citep{xu2018how, morris2019weisfeiler}. We call it \emph{symmetry slippage}. Despite its prominence, the principles of creating, computing, and correcting symmetry slippage have yet to be systematically investigated in contemporary geometric ML.

The present work aims to initiate such a study. We develop a calculus for studying end-to-end symmetry slippage from that of local, modular architectural components, and from this derive three general mechanisms, called \emph{group-closure}, \emph{over-quotienting}, and \emph{extension-mistyping}, by which design choices in equivariant ML can produce symmetry slippage. The first mechanism, group-closure, describes how the class of functions used to build $\mathscr{A}$ can impose a specific symmetry slippage which may be called \emph{over-equivariance}. Group-closure examples include ``automorphism group-closure", which includes (but is not limited to) a classical finite group setting, and ``Zariski group-closure", which occurs e.g. when polynomial features incidentally extend equivariance from orientation-sensitive Lie subgroups such as $\operatorname{SO}^{+}(1,d)$, $\operatorname{GL}^{+}(d)$, or $(\mathbb{R}^{+})^k$. The second mechanism, over-quotienting, describes architecture operations which can identify inputs more coarsely than prescribed. The symmetry slippage induced by over-quotienting is more general than the over-equivariance induced by group-closure; for instance, we show that over-quotienting in message-passing graph neural networks converts $S_{n}$-invariance into \emph{Weisfeiler-Lehman} ($\operatorname{WL}$)-invariance \citep{Weisfeiler1968, xu2018how, morris2019weisfeiler}. The third mechanism, extension-mistyping, induces a still more general symmetry slippage. It occurs when a transformation law on a subgroup admits multiple extensions to a larger group, and certain architectural design choices incidentally choose the `wrong' one.  

To demonstrate the applicability of these creation mechanisms, we present nine case studies (three per mechanism) of current research areas in which symmetry slippage occurs but has been un- or incompletely-recognized. In some of these case studies, we locate existing heuristics that partially and implicitly address symmetry slippage in certain domains but not others. Because these heuristics are often motivated by domain-specific intuition, we hope that our creation mechanisms' theoretical, domain-agnostic grounding of them will boost their transferability. 

The ease with which symmetry slippage can be unintentionally created leads naturally to inquiry into how its absence can be proven. While we show that exactly computing $\operatorname{Sym}\mathscr{A}$ is GI-hard in general, even when promised $\mathrm{A} \leq \operatorname{Sym}\mathscr{A}$, we prove conditions under which $\operatorname{Sym}\mathscr{A}$ can be exactly computed by composing the transformation laws of $\mathscr{A}$'s subarchitectures. To exhibit this structure theorem, we apply it to compute $\operatorname{Sym}\mathscr{A}$ exactly for the classical setting of sparse parameter-sharing equivariant MLPs (e.g. convolutional architectures~\citep{cohen2016group}), by composing per-layer transformation law computations. Although this classical setting is quite well-studied~\citep{ravanbakhsh2017equivarianceparametersharing, maron2018invariant, maron2019universality}, this constitutes to our knowledge its first exact computation of $\operatorname{Sym}\mathscr{A}$.

To correct symmetry slippage and restore expressivity, we introduce \emph{grippage encodings (GEs)}, a simple type of positional encoding~\citep{vaswani2017attention, dwivedi2023benchmarking, rampavsek2022recipe} tailored to our three creation mechanisms. For each creation mechanism, we derive a recipe for building GEs that repair symmetry slippage. GEs may be incorporated from the outset into a (re)design, especially when methods to avoid the symmetry slippage are bespoke or otherwise nonobvious. They may also be used to tweak the transformation law of an existing architecture when retraining would be expensive or the prescribed law varies with task, akin to how symmetry-breaking techniques have been recently used to modulate foundation model transformation laws \citep{goel2026any}. We empirically show GEs improve performance across each our nine case studies.

To summarize, our contributions are:
\begin{itemize}[leftmargin=1.5em]
\item (Creation) We develop a calculus for studying symmetry slippage via architectural decompositions. Our framing identifies three symmetry slippage creation mechanisms: \emph{group-closure}, \emph{over-quotienting}, and \emph{extension-mistyping}. These mechanisms expose previously unrecognized or incompletely-characterized symmetry slippage across nine active areas of research.
\item (Computation) While we show that exactly computing $\operatorname{Sym}\mathscr{A}$ is hard in general, we prove conditions under which $\operatorname{Sym}\mathscr{A}$ can exactly be computed from subarchitectures. We apply this to compute $\operatorname{Sym}\mathscr{A}$ for a well-studied classical setting subsuming e.g. convolutional architectures.
\item (Corrections) We introduce \emph{grippage encodings}: lightweight symmetry-correcting features that restore expressivity. We empirically verify that grippage encodings improve performance on our nine case studies. 
\end{itemize}

\section{Background and Motivation}
\label{sec:background}

\begin{wrapfigure}{r}{0.4\textwidth}
\vspace{-22pt}
    \centering
\includegraphics[width=\linewidth]{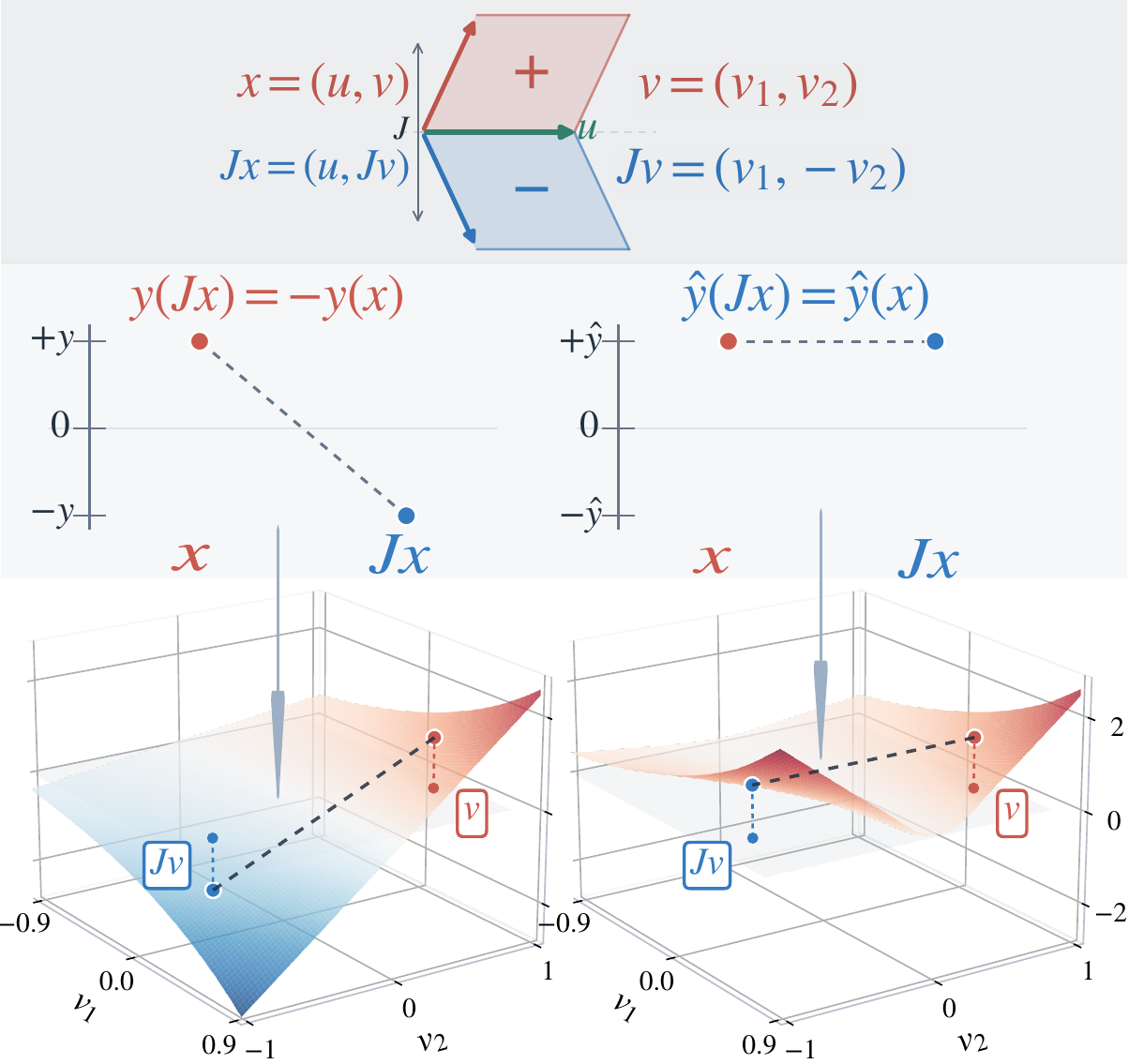}
\vspace{-20pt}
\caption{\textcolor{blue}{\textbf{Symmetry Slippage}: No \(\operatorname{O}(2)\)-invariant architecture can learn a reflection-sensitive target, providing a classical example of symmetry slippage. Here, we trained the architecture $f^{\theta}:(\mathbb{R}^{2} )^{2}\xrightarrow{\gamma}\mathbb{R}^{3} \xrightarrow{\operatorname{MLP}^{\theta}} \mathbb{R}$, $(\boldsymbol u,\boldsymbol v) \xmapsto{\gamma}(\|\boldsymbol u\|^{2}, \|\boldsymbol v\|^{2}, \boldsymbol u \cdot \boldsymbol v )$ against the target $y(\boldsymbol u,\boldsymbol v)=e^{\boldsymbol u \cdot \boldsymbol v}\det[\boldsymbol u\, \, \boldsymbol v]$. To illustrate the expressivity loss, we fix $\boldsymbol u=(1,0)$ and plot $\boldsymbol v=(v_{1},v_{2})$ against the target $y$ and learned $\hat{y}$. The two agree for $v_{2}>0$ but differ by a sign for $v_{2}<0$. This reflects symmetry slippage of the sign representation $\mathrm A_{\mathrm{}}:=\underline{\operatorname{O}(2)}_{\text{}}^{\text{sign}}=\{\big({Q,\det(Q)I\big): Q \in  \operatorname{O}(2)}\}$, as well as symmetry slippage of $\mathrm A_{}:=\underline{\operatorname{SO}(2)}=\{{(R,I_Y): R\in \operatorname{SO}(2)}\}$.} 
}
\vspace{-35pt}
\label{fig:motivational-figure}
\end{wrapfigure}

\paragraph{Transformation Laws.} By an \textit{architecture} $\mathscr{A}:X \rightsquigarrow Y$ we mean a map $\Theta \to Y^{X}$, where $\Theta$ is a space of parameters. We make precise the notion of ``symmetries of $\mathscr{A}$" using transformation laws. \textcolor{blue}{Roughly speaking, by a \textit{transformation law} we mean any rule $\mathrm{L}$ assigning certain transformations $\sigma$ of the input space $X$ to certain transformations $\tau$ of the output space $Y$. We then say $\mathscr{A}$ \textit{obeys} the transformation law $\mathrm{L}$ if any function $f^{\theta}$ realized by $\mathscr{A}$ satisfies the equivariance condition $f^{\theta} \sigma = \tau f^{\theta}$ for all pairs $(\sigma, \tau)$ comprising $\mathrm{L}$. For example, the architecture in Figure~\ref{fig:motivational-figure}, \(f^{\theta}:(\mathbb{R}^{2} )^{2}\xrightarrow{\gamma}\mathbb{R}^{3} \xrightarrow{\operatorname{MLP}^{\theta}} \mathbb{R}\), \((u,v) \xmapsto{\gamma} (\|u\|^{2}, \|v\|^{2}, u \cdot v )\), is manifestly invariant under the natural action $\alpha_{X}:\operatorname{SO}(2) \to \operatorname{GL}((\mathbb{R}^{2})^{2})$ and thus obeys the law\footnote{Here we have standardly abused notation by writing $Q$ in place of $\alpha(Q)$.} $\mathrm{A}=\{ (Q,I): Q \in \textcolor{SteelBlue}{\operatorname{SO}(2)} \}$. It also obeys the larger law $\operatorname{L}=\{ (Q, I): Q \in \textcolor{SteelBlue}{\operatorname{O}(2)}  \}$, however. We call the \textit{largest} transformation law obeyed by an architecture $\mathscr{A}$ its \textit{realized transformation law}, and denote it $\operatorname{Sym}\mathscr{A}$. To make this all precise:} with $J_{X}$ and $J_{Y}$ denoting ambient transformation groups of $X$ and $Y$, a \textit{transformation law} is a subgroup $\mathrm{L} \leq J_{X} \times J_{Y}$, and the \textit{realized transformation law} of $\mathscr{A}$ is $$\operatorname{Sym}\mathscr{A}=\{ (\sigma, \tau) \in J_{X} \times J_{Y} : f ^{\theta}\sigma = \tau f^{\theta} \text{ for all } \theta \in \Theta\}.$$\textcolor{blue}{In the example of Figure~\ref{fig:motivational-figure}, $J_{X}=\operatorname{GL}(X)$ and $J_{Y}=\operatorname{GL}(Y)$. Other common examples we will encounter are when $J_{X},J_{Y}$ are isomorphic to finite permutation groups $S_{n}$ or are even just all bijections of $X$ and $Y$ with no additional structure imposed.}

\paragraph{Symmetry Slippage}
Equivariant learning designs architectures $\mathscr{A}:X \rightsquigarrow Y$ obeying a transformation law $\mathrm{A} \leq J_{X} \times J_{Y}$ prescribed by group actions $\alpha_{X}:H \to J_{X}$, $\alpha_{Y}:H \to J_{Y}$. Specifically, $\mathrm{A}:=\{ (\alpha_{X}(h), \alpha_{Y}(h)) \}_{h \in H}$. (We also use the notation $\mathrm{A}=\underline{H}_{\alpha_{X}}^{\alpha_{Y}}$, omitting the subscript or superscript as context permits.) \textcolor{blue}{Usually $H$ is defined in terms of an ambient group $\Gamma \geq H$, with $\alpha_{X}$ arising as the restriction of an ambient action $\alpha:\Gamma \to J_{X}$. (To return to the example of Figure~\ref{fig:motivational-figure}, there $\Gamma=\operatorname{O}(2)$ and $H=\operatorname{SO}(2)$.)} The goal is to achieve $\mathrm{A}=\operatorname{Sym}\mathscr{A}$. One way equality fails is $\operatorname{Sym}\mathcal{}{\mathscr{A}} \not \supset \mathrm{A}$, which can be called \textit{under-equivariance}. Under-equivariance is extensively studied, insofar as the most fundamental motivation for equivariant learning is to avoid it. The other way equality fails is $\operatorname{Sym} \mathscr{A} \not \subset \mathrm{A}$, when $\mathscr{A}$ realizes transformations that were not prescribed. We call this \textit{symmetry slippage}. Symmetry slippage contains \textit{over-equivariance} as the special case when $\mathrm{A}:=\underline{H}_{\alpha|_{H}}^{\alpha_{Y}}$ is prescribed for some $H \leq \Gamma$, but a group $\Gamma \geq G \gneq H$ exists and $\alpha_Y$ extends to an action of $G$ on $Y$ such that $\underline{H}_{\alpha|_{H}}^{\alpha _{Y}} \subsetneq \underline{G}_{\alpha|_{G}}^{\alpha _{Y}}\leq \operatorname{Sym}\mathscr{A}$. \textcolor{blue}{For example, the latter symmetry slippage described in Figure~\ref{fig:motivational-figure},  \(\underline{\operatorname{SO}(2)} \subsetneq \underline{\operatorname{O}(2)}=\operatorname{Sym}\mathscr{A}\), is via over-equivariance.} The notion of symmetry slippage is altogether more general than over-equivariance, however. \textcolor{blue}{For example, the former symmetry slippage described in Figure~\ref{fig:motivational-figure} is \(\underline{\operatorname{O}(2)}^{\text{sign}} \not \subset \underline{\operatorname{O}(2)} =\operatorname{Sym}\mathscr{A}\) with no inclusion in either direction.} A detailed contextualization of the relationship between symmetry slippage and over-equivariance may be found in Appendix~\ref{sec:related-work}. 

This work aims to systematically discuss how standard choices in building an architecture $\mathscr{A}$ can and do create symmetry slippage across active fields, how symmetry slippage can be (dis)proven, and how symmetry slippage can be repaired in practice. 
\textcolor{blue}{In the simple scenario of Figure~\ref{fig:motivational-figure}, the existence of symmetry slippage follows from inspection. There is a rather self-evident repair as well, given merely by appending a feature that is \(\operatorname{SO(2)}\)-invariant but not \(\operatorname{O}(2)\)-invariant such as \(c(u,v)=\det[u\,\,v]\). In general, however, there may be complicated groups, actions, and architectures interacting in complicated ways, and understanding the relationship between a prescribed law \(\mathrm{A}\) and the realized law \(\operatorname{Sym}\mathscr{A}\) becomes subtle. For example, suppose that in Figure~\ref{fig:motivational-figure} the group $\operatorname{SO}(2)$, consisting of maps that preserve orientation and the dot product $u \cdot v=u_{1}v_{1}\textcolor{SteelBlue}{+}u_{2}v_{2}$, was replaced with the Lorentz group $\operatorname{SO}^{+}(1,1)$ of maps that preserve orientation and the \textit{Minkowski form} $\langle u,v \rangle=u_{1}v_{1}\textcolor{SteelBlue}{-}u_{2} v_{2}$. We will find in this case that not only does appending an orientation-sensitive feature such as $\det[u \, \, v]$ fail to prevent symmetry slippage, but \textit{every analytic feature} fails to prevent symmetry slippage. This is an instance of our \textit{group-closure} creation mechanism (Section~\ref{sec:creation-group-closure}), and we find it causes symmetry slippage in several popular Lorentz-equivariant architectures (Case Study \ref{case:lorentz}). Our other eight case studies (Section~\ref{sec:correcting}) identify further examples of unrecognized symmetry slippage in contemporary work. Symmetry slippage's subtleties are witnessed also by certain \textit{well}-recognized examples such as famous results linking GNN expressivity to the WL isomorphism test (an instance of our \textit{over-quotienting} creation mechanism; Section~\ref{sec:over-quotienting}). The situation complicates further when, instead of asking if symmetry slippage exists, one asks if it does \textit{not} exist. Answering this question amounts to computing $\operatorname{Sym}\mathscr{A}$ exactly, a task that can be decidedly nontrivial (Section~\ref{sec:computing}).} 

\textcolor{blue}{To accommodate this diversity of architectures $\mathscr{A}$ to which symmetry slippage applies,} our analysis is based on (i) understanding the symmetries of subarchitectures, and (ii) understanding in general how ``local" subarchitectural symmetries relate to ``global", end-to-end symmetries. For example, to understand the symmetries of an \(\mathscr{A}=\operatorname{MLP}\), we seek to understand the extent to which the symmetries of its individual layers (subarchitectures) can be ``composed" into symmetries of \(\mathscr{A}\) itself. \textcolor{blue}{To precisely reason about the subarchitectural pieces of more general $\mathscr{A}$, we adopt the language of computation graphs. For instance, an MLP can be viewed as a computational graph of neurons \(v \in V\) and weights; the layers \(L\) of an MLP correspond to node cuts \(C \subset V\) in this graph, splitting the MLP into upstream and downstream computational subgraphs.} In general, we identify an architecture $\mathscr{A}$ with its computation graph $\mathscr{A}=\big( V, E ,  (Z_{v})_{v \in V}, (\phi_{v})_{v \in V} \big)$, and studying cut-induced subgraphs of the latter. Here, $(V,E)$ is a DAG having input nodes $I \subset V$ and output nodes $O \subset V$, the $Z_{v}$ are vector spaces, and $\phi_{v}=\{ \phi_{v}^{\theta}:Z_{\{ u \to v \}} \to Z_{v} \}_{\theta \in \Theta}$ is the parameterized family of \textit{node maps} at $v$, where we have set the notation $Z_{B}:=\prod_{v \in B}^{}Z_{v}$ for $B \subset V$ and $\{ u \to v \}:=\{ u \in V : (u, v) \in E \}$).

\section{Creating Symmetry Slippage}
\label{sec:creating}

\subsection{Decomposing Transformation Laws}
\label{sec:decomposing-transformation-laws}
We now make precise the local-to-global viewpoint described in Section~\ref{sec:background}. Say a pair $(C_{1},C_{2})$ of node cuts of $V$ is \emph{good} if every path from an input $i \in I$ to an output $o \in O$ meets $C_{1}$ before $C_{2}$. A good pair $(C_{1},C_{2})$ uniquely determines a computation graph $\mathscr{A}_{(C_{1}, C_{2})}:Z_{C_{1}} \rightsquigarrow Z_{C_{2}}$, which we call the \emph{subarchitecture} of $\mathscr{A}$ corresponding to $(C_{1},C_{2})$. Every good pair $(C_{1},C_{2})$ induces, for each $\theta \in \Theta$, a factorization $f^{\theta}: X \inarrow{a^{\theta}} Z_{C_{1}} \inarrow{\Phi^{\theta}} Z_{C_{2}} \inarrow{e^{\theta}} Y$, where $a^{\theta}$, $\Phi^\theta$, and $e^{\theta}$ are realized by $\mathscr{A}_{(I, C_{1})}$, $\mathscr{A}_{(C_{1},C_{2})}$, and $\mathscr{A}_{(C_{2}, O)}$, respectively. In particular, given a node cut $C \subset V$, the pairs $(I, C)$ and $(C, O)$ are always good, whence $C$ splits $\mathscr{A}$ into subarchitectures
$\mathscr{A}: X \overset{\mathscr{A}_{(I, C)}}{\rightsquigarrow} Z_{C}
\overset{\mathscr{A}_{(C, O)}}{\rightsquigarrow} Y$.
At the level of hypothesis classes, however, this decomposition is only \textit{lax}: in general,
$\mathcal{F}_{\mathscr{A}} \subset
\mathcal{F}_{\mathscr{A}_{(C, O)}}\circ
\mathcal{F}_{\mathscr{A}_{(I ,C)}}$ but equality need not hold.
\begin{restatable}{proposition}{decomposingtlaws}
\label{prop:decomposing-tlaws}
Any subarchitecture $\mathscr{A}_{(C_{1} ,C_{2})}$ gives rise to a (lax) decomposition
\begin{equation}
\operatorname{Sym}\mathscr{A}_{(C_{2}, O)} \circ \operatorname{Sym}\mathscr{A}_{(C_{1} , C_{2})} \circ \operatorname{Sym} \mathscr{A}_{(I, C_{1})} \leq \operatorname{Sym}\mathscr{A}.
\label{eq:compos}
\end{equation}

Equality need not hold in general.
\end{restatable}


Proposition~\ref{prop:decomposing-tlaws} implies that to identify symmetry slippage in an architecture $\mathscr{A}$, it is sufficient to find a subarchitecture $\mathscr{A}_{(C_{1}, C_{2})}$ whose realized law contains an unprescribed transformation preserved by upstream and downstream computation. At the same time, it implies $\operatorname{Sym}\mathscr{A}$ cannot in general be computed from subarchitectures. Our creation mechanisms in Section~\ref{sec:local-symmetry-slippage-creation-mechanisms} exploit the first fact. Our computational results in Section~\ref{sec:computing} investigate conditions under which equality in (\ref{eq:compos}) does hold.

\begin{wrapfigure}{r}
{0.7\textwidth}
\vspace{-19pt}
    \centering
    \includegraphics[width=\linewidth]{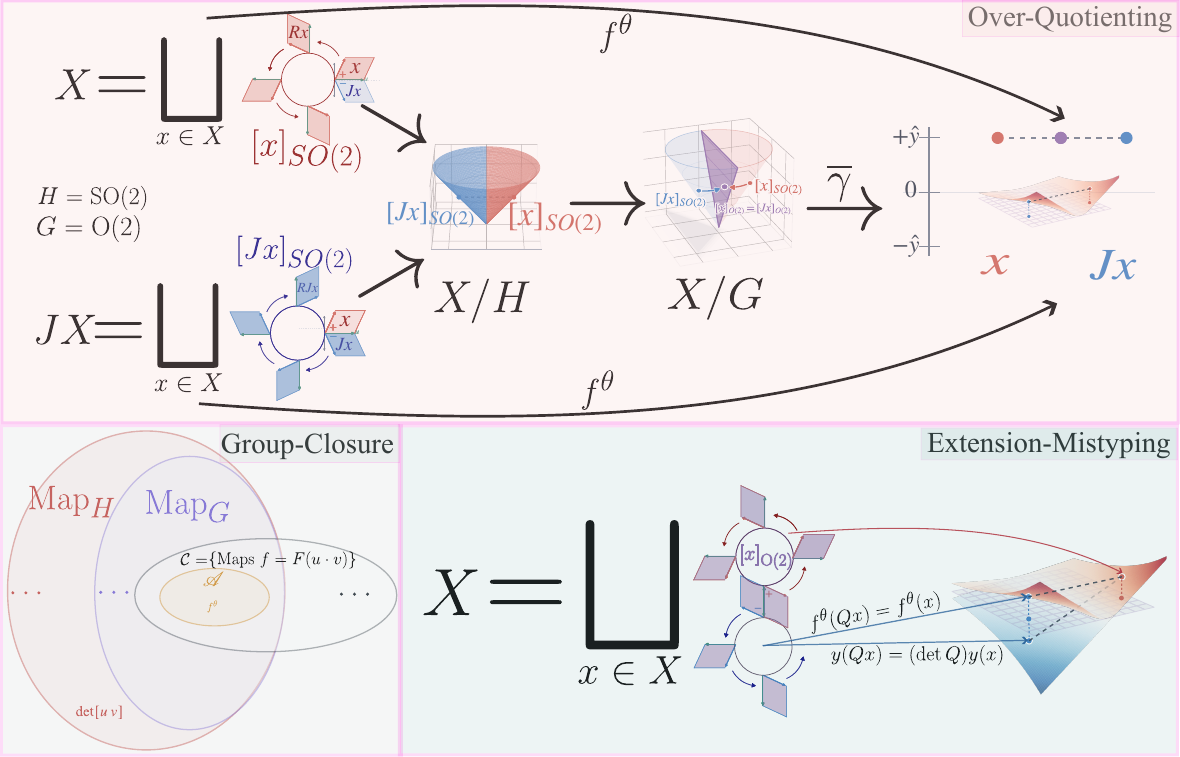}
    \vspace{-2em}
    \caption{\textcolor{blue}{Three creation mechanisms for symmetry slippage.}}
    \vspace{-15pt}
    \label{fig:creation-mechanisms-toy}
\end{wrapfigure}

\subsection{Local Symmetry Slippage}
\label{sec:local-symmetry-slippage-creation-mechanisms}

With the framing of Section~\ref{sec:decomposing-transformation-laws} in mind, we describe three mechanisms by which symmetry slippage may be locally created. \textcolor{blue}{To motivate these creation mechanisms, we first return to architecture $f^{\theta}:(\mathbb{R}^{2} )^{2}\xrightarrow{\gamma}\mathbb{R}^{3} \xrightarrow{\operatorname{MLP}^{\theta}} \mathbb{R}$ outlined in Figure~\ref{fig:motivational-figure}. We can view its symmetry slippage in this example as arising for three distinct reasons, which, while aligning in this very special case, will extend in three different directions to three distinct \textit{creation mechanisms} for symmetry slippage. We call these \textit{group-closure}, \textit{over-quotienting}, and \textit{extension-mistyping} respectively.}

\textcolor{blue}{
Roughly,
\begin{itemize}[leftmargin=1.5em]
    \item \textit{group-closure }occurs when the function class from which an architecture is built has too much symmetry, causing over-equivariance. For example, $f^{\theta}$ parameterizes functions of dot products of feature vectors, which are not only $\operatorname{SO}(2)$-invariant but $\operatorname{O}(2)$-invariant. 
    \item \textit{over-quotienting} occurs when an architecture incidentally identifies inputs. For example, $\gamma$ identifies inputs related by reflections $J$, collapsing two $\operatorname{SO}(2)$-orbits into a single $\operatorname{O}(2)$ orbit.
    \item \textit{extension-mistyping} occurs when a transformation law of $\mathscr{A}$ has multiple extensions to a larger group, and an architecture incidentally realizes the wrong one. For example, the $\operatorname{O}(2)$-equivariant target $y$ is a \textit{pseudoscalar}, meaning it extends from \(\operatorname{SO}(2)\)-invariance to $y(Qx)=-y(x)$ for $Q \in \operatorname{O}(2)-\operatorname{SO}(2)$; but $f^{\theta}$ instead extends from \(\operatorname{SO}(2)\)-invariance to $\operatorname{O}(2)$-invariance. 
\end{itemize}
See Figure~\ref{fig:creation-mechanisms-toy}. }Below, we make precise these notions and how they cause symmetry slippage. 

\subsubsection{Group-Closure}
\label{sec:creation-group-closure}
Group-closure gives local conditions under which over-equivariance (Section~\ref{sec:background}; App. \ref{sec:related-work}) occurs.

\begin{definition}
\label{def:group-closure}\textcolor{blue}{
Let $\mathscr{B}$ be a subarchitecture whose node maps $\phi_{v}^{\theta}$ belong to a class $\mathcal{C}$ of functions. Let $H\leq G \curvearrowright Z_{v}$ be an inclusion of group actions. We say \emph{$\mathcal{C}$ closes $H$ to $G$ on $\mathscr{B}$} if for every node $v$ of $\mathscr{B}$, \(\mathcal{C} \cap \operatorname{Map}_{H}(Z_{\{ u \to v \}}, Z_{v}) = \mathcal{C} \cap \operatorname{Map}_{G}(Z_{\{ u \to v \}}, Z_{v})\).}

\end{definition}

\begin{proposition}[Informal]
\textcolor{blue}{
     Suppose $\mathcal{C}$ closes $H$ to $G$ on a subarchitecture $\mathscr{B}$ whose nodes maps are $H$-equivariant and belong to $\mathcal{C}$, and whose upstream and downstream components are both $G$-equivariant. Then $\underline{H}_{\alpha_{X}}^{\alpha _{Y}} \lneq \underline{G}_{\alpha_{X}}^{\alpha _Y} \leq \operatorname{Sym}\mathscr{A}$.   }
\end{proposition}
In other words, group-closure causes symmetry slippage, in particular via over-equivariance.

\vspace{.4em}

\begin{example}[Automorphism closure]
\label{example:automorphism-closure}
A common strategy is to view elements of $H$ as automorphisms of some object $\Sigma$, then constrain $\mathscr{A}$ to a class $\mathcal{C}$ of $\operatorname{Aut}(\Sigma)$-equivariant maps. This strategy only guarantees $H \leq \operatorname{Aut}(\Sigma)$, not equality. Consequently, $\mathcal{C}$ closes $H$ to $\operatorname{Aut}(\Sigma)$, with symmetry slippage occurring when $H \lneq \operatorname{Aut}(\Sigma)$.
\end{example}

\begin{subexample}
\label{subex:motivational-toy}
\textcolor{blue}{
Letting $\Sigma:= \big( \mathbb{R}^{2}, \_\cdot \_ \big)$, so that $\operatorname{Aut}(\Sigma)=\operatorname{O}(2)$, and letting $\mathcal{C}=\{ f:F(u \cdot v) : u,v \in \mathbb{R}^{2}\}$, we recover the group-closure example depicted in Figure~\ref{fig:creation-mechanisms-toy}. }
\end{subexample}

\begin{subexample}
\label{subexample:aut-closure-finite}
The classical setting of building $\mathscr{A}$ from morphisms of finite-group permutation representations, as is done e.g. in G-CNNs \citep{cohen2016group}, GCNs \citep{bruna2013spectral,henaff2015deep,kipf2016semi}, and DeepSets \citep{ravanbakhsh2016deep,zaheer2017deep}, is contained in Example~\ref{example:automorphism-closure}. Indeed, let $I$ and $O$ be the sets of input and output neurons of a single linear $\operatorname{MLP}$ layer $\mathscr{A}_{(I,O)}$. Let $H \curvearrowright I$, $H \curvearrowright O$ be finite actions, inducing permutation representations $H \to J:=S_{I} \leq \operatorname{GL}(\mathbb{R}^{I})$, $H\to K:=S_{O} \leq \operatorname{GL}(\mathbb{R}^{O})$. One can check that, in this setting, $\operatorname{Hom}_{H}(\mathbb{R}^{I},\mathbb{R}^{O})=\mathcal{C}$, where $\mathcal{C}$ is the space of matrices constant on index orbits. Thus $H$-equivariance can be implemented by orbitwise parameter-sharing. The orbit construction, however, only yields $\underline{H}_{\alpha_X}^{\alpha_Y}\leq\operatorname{Aut}(\Sigma)$ in general, with $\mathcal{C}$ closing $H$ to $\operatorname{Aut}(\Sigma)$, where $\Sigma$ is a coloring of the edges $E=O \times I$ of $\mathscr{A}_{(I,O)}$ by $H$-orbit. This perspective was studied by \citet{ravanbakhsh2017equivarianceparametersharing}, who introduced the term \emph{uniquely equivariant} for what is in our present notation equality $\underline{H}_{\alpha_X}^{\alpha_Y}=\operatorname{Aut}(\Sigma)$, i.e. the absence of symmetry slippage by over-equivariance. 

If the actions on $I$ and $O$ are in fact free, there is a (noncanonical) bijection $\Psi:(O \times I) / H \to H \times O / H \times I / H$ identifying $H$-orbits in $O\times I$ with triples of the form $(a, Ho, Hi)$ for $a \in H$, $o \in O$, $i \in I$. This makes it possible to sparsify the above orbitwise parameter-sharing construction, by fixing $A \subset H$ and retaining only those $H$-orbits $H(o,i)$ for which $\Psi_{1}\big( H(o, i) \big) \in A$ \citep{ravanbakhsh2017equivarianceparametersharing}.\footnote{For instance, if $H=C_{n}$ acts by circular translation on $I=O=C_{n}$, then $O / H$, $I / H$ are singletons (the action is transitive) and so $H$-orbits in $O \times I$ are in bijection with relative translations $a \in C_{n}$ between input and output indices. Retaining only $a \in A$ then recovers circular convolution with a kernel having support $A$.} An $\operatorname{MLP}$ which implements equivariance by sharing parameters according to $A$ may be called an \emph{$A$-$\operatorname{MLP}$}; in Section~\ref{sec:computing}, we will exactly compute $\operatorname{Sym}(A\text{-}\operatorname{MLP})$.

\end{subexample}

Example \ref{example:automorphism-closure} is earnestly more general than the classical finite setting above. For instance, our Case  Study~\ref{case:intaut} (Section~\ref{sec:case-studies-main-group-closure}) applies Example~\ref{example:automorphism-closure} with $\Sigma=\mathfrak{g}$ a Lie algebra rather than a colored graph.

\medskip

\begin{example}[Zariski closure]
\label{ex:zariski-closure}
\textcolor{blue}{For certain groups $G$, including $\operatorname{SO}(p,q)$ for $p,q \geq 1$, $\operatorname{GL}(d, \mathbb{R})$, and $(\mathbb{R}^{\times})^{d}$, equivariance under the identity component $G_{0}$ already implies equivariance under all of $G$ for polynomial maps, $\mathcal{C}=\mathbb{R}[X]$, and rational maps, $\mathcal{C}=K(X)$, whence $\mathcal{C}$ closes $G_{0}$ to $G$. The general criterion is algebro-geometric, being based on the density of $G_{0}$ in $G$ with respect to the Zariski topology, $\overline{G_{0}}^{\text{Zariski}}=G$; see Appendix~\ref{app:Zariski-group-closure}. 
To our knowledge, this  ``Zariski (group) closure" phenomenon has not been discussed in equivariant learning. Nevertheless, it causes symmetry slippage in several existing architectures (cf. Case Studies \ref{case:lorentz} and \ref{case:units}).} In linear settings, we are able to upgrade Example~\ref{ex:zariski-closure} from $\mathcal{C}=\mathbb{R}[X]$ to the analytic class $C^{\omega}(X)$.
\end{example} \begin{restatable}{proposition}{groupclosureanalyticprop}
\label{prop:group-closure-poly-to-analytic}
If the spaces $Z_{v}$ for $v$ in $\mathscr{B}$ are finite-dimensional $G$-representations and polynomials $\mathbb{R}[X]$ close $H$ to $G$ on $\mathscr{B}$, then $\mathcal{C}:=C^{\omega}(X)$ closes $H$ to $G$ on $\mathscr{B}$. 
\end{restatable} 
%
%
%

\subsubsection{Over-Quotienting}
\label{sec:over-quotienting}
While group-closure investigates the setting where the function class $\mathcal{C}$ from which $\mathscr{A}$ is built induces over-equivariance and hence symmetry slippage, \textit{over-quotienting} investigates the setting where an operation in $\mathscr{A}$ itself can induce symmetry slippage. 

\begin{definition}
\label{def:over-quotienting}
We say that $\mathscr{A}$ \emph{over-quotients} $\mathrm{A}$ if there exist a subarchitecture $\mathscr{A}_{(C_1,C_2)}$, an equivalence relation $\sim$ on $Z_{C_1}$, \textcolor{blue}{and $s\in J_X$ such that $(s,I_Y)\notin\mathrm{A}$ and, for all $\theta\in\Theta$, there exists a unique $\overline{\Phi}^{\theta}$ making the following diagram commute:}
\vspace{-1.8em}

\[
\scalebox{0.74}{$
\begin{tikzcd}[
  ampersand replacement=\&,
  row sep=0.4em,
  column sep=1.5em
]
          \& \& X \arrow[r,"a^\theta"]
          \& Z_{C_1}
            \arrow[rd]
            \arrow[rrrd,bend left=8,end anchor=north west,"\Phi^\theta"]
          \& \& \& \& \& \\
f^\theta: \& X
            \arrow[ru,"I"]
            \arrow[rd,"s"']
          \& \&
          \& Z_{C_1}/{\sim}
            \arrow[rr,dashed,
              "{\scriptstyle\exists!\,\overline{\Phi}^\theta}" description]
          \& \hspace{0.8em}
          \& Z_{C_2}
            \arrow[rr,"e^\theta" description]
          \&
          \& Y \\
          \& \& X \arrow[r,"a^\theta"']
          \& Z_{C_1}
            \arrow[ru]
            \arrow[rrru,bend right=8,end anchor=south west,"\Phi^\theta"']
          \& \& \& \& \&
\end{tikzcd}
$}
\]

\vspace{-1.4em}
\end{definition}
\textcolor{blue}{Unpacking this diagram}, the over-quotienting condition is verified by checking $a^{\theta}s(x)\sim a^{\theta}(x)$ for all $x \in X$ and $\theta \in \Theta$, then checking that $\Phi^{\theta}$ is constant on $\sim$-classes and so preserves this identification. Finally, one checks $(s, I_{Y} )\not \in \mathrm{A}$, which verifies the resulting invariance was not prescribed by $\mathrm{A}$. The next result follows immediately:

\begin{restatable}{proposition}{overquotientingprop}
\label{prop:over-quotienting}
If $\mathscr{A}$ over-quotients $\mathrm{A}$, then $f^\theta s=f^\theta$ for all $\theta$. Therefore, $(s,I_Y)\in \operatorname{Sym}\mathscr{A}-\mathrm{A}$ and $\mathscr{A}$ exhibits symmetry slippage.
\end{restatable}


\begin{example}
\label{ex:expositional-oq-example}
\textcolor{blue}{
    The architecture $f^{\theta}=\overbrace{ e^{\theta} }^{ \operatorname{MLP}^{\theta} }\circ \overbrace{ \Phi }^{ \gamma}$  of Figure~\ref{fig:motivational-figure} overquotients $\underline{\operatorname{SO}(2)}$, as is witnessed by taking $\sim:= \sim_{\operatorname{O(2)}}$ the orbit relation  of $\operatorname{O}(2)$ and $s:=J:=\operatorname{diag}(1,-1)$ in Definition~\ref{def:over-quotienting} (Figure~\ref{fig:creation-mechanisms-toy}).}  
\end{example}
\textcolor{blue}{Comparing with Example~\ref{subex:motivational-toy}, Example~\ref{ex:expositional-oq-example} shows that, like group-closure, overquotienting can cause over-equivariance. }
Unlike group-closure, however, not all over-quotienting instances are over-equivariance. For example, over-quotienting in vanilla MPNNs \citep{gilmer2017neural} converts $S_{n}$-invariance into constancy on $\operatorname{WL}$-classes \citep{xu2018how, morris2019weisfeiler}, even though these additional identifications are not induced by any $S_{n} \lneq G \leq \Gamma$ (see Appendix~\ref{sec:related-work}). Case Study~\ref{case:glycan}  reinforces that over-quotienting beyond over-equivariance also occurs unrecognized into recent published work.

\subsubsection{Extension-Mistyping}
\label{sec:extension-mistyping}
In both group-closure and over-quotienting, symmetry slippage $\operatorname{Sym} \mathscr{A} \not \subset \mathrm{A}$ generally occurs because more transformations are realized than prescribed, $\mathrm{A} \subsetneq \operatorname{Sym}\mathscr{A}$. We next introduce a creation mechanism, \emph{extension-mistyping}, for which $\mathrm{A}$ and $\operatorname{Sym}\mathscr{A}$ may be earnestly incomparable. 
\textcolor{blue}{To motivate extension-mistyping, recall that the symmetry slippage in Figure~\ref{fig:motivational-figure} may be viewed as the architecture extending the transformation law $\underline H=\underline{\operatorname{SO}(2)}_{\rho_X}^{I}$ to $\underline G=\underline{\operatorname{O}(2)}_{\widetilde\rho_X}^{I}$ rather than the prescribed law $\mathrm A=\underline{\operatorname{O}(2)}_{\widetilde\rho_X}^{\mathrm{sign}}$, where $\rho_X$ is the $H=\operatorname{SO}(2)$-action on $X$ and $\widetilde\rho_X$ is its standard extension to $G=\operatorname{O}(2)$. In other words, the prescribed extension of $\boldsymbol\rho:=(\rho_X,I)$ was $\widetilde{\boldsymbol\rho}^{\mathrm p}:=(\widetilde\rho_X,\mathrm{sign})$, but the realized extension was $\widetilde{\boldsymbol\rho}^{\mathrm r}:=(\widetilde\rho_X,\eta_Y)$, where $\eta_Y$ is the trivial $G$-action on $Y$, differing from $\mathrm{sign}$ under reflections. More generally,} consider a subarchitecture $\mathscr{B}$ with $H$-action $\boldsymbol \rho$ on its spaces. Suppose certain nodes $T \subset V$ come equipped with $G$-actions $\eta_v$ extending $\rho_v$ (in the motivating example above, $T=\{ Y \}$), while $\widetilde{\boldsymbol\rho}^{\mathrm p}$ denotes the prescribed extension of $\boldsymbol\rho$ to $G$.


\begin{definition}
\label{def:extension-mistyping}
Say  $\mathscr{A}$ \emph{extension-mistypes $\tilde{\rho}_{}^{\text{p}}$ on $\mathscr{B}$} if there is an extension $\tilde{\rho}^{\text{r}}$ of $\rho$ such that $\tilde{\rho}_{v}^{\text{r}}=\eta_{v}$ for \emph{all} $v \in T$, $\eta_{v}\neq \tilde{\rho}_{v}^{\text{p}}$ for \emph{some} $v \in T$, and the node maps $\phi_{v}^\theta$ of $\mathscr{B}$ are \(G\)-equivariant with respect to \(\tilde{\boldsymbol\rho}^{\mathrm r}\).
\end{definition}

 \begin{restatable}{proposition}{extensionmistypingprop}
\label{prop:extension-mistyping}
Suppose $\mathscr A$ extension-mistypes $\widetilde{\boldsymbol\rho}^{\mathrm p}$ on $\mathscr{B}=\mathscr{A}_{(C_{1},C_{2})}$ via $\widetilde{\boldsymbol\rho}^{\mathrm r}$. Suppose further that there exist actions $\alpha_X^{\mathrm r}: G \to J_{X}$ and $\alpha_Y^{\mathrm r}:G \to J_{Y}$ such that $a^\theta$ and $e^\theta$ are $G$-equivariant with respect to $(\alpha_X^{\mathrm r},\widetilde\rho_{C_1}^{\mathrm r})$ and $(\widetilde\rho_{C_2}^{\mathrm r},\alpha_Y^{\mathrm r})$. Then $\underline G_{\alpha_X^{\mathrm r}}^{\alpha_Y^{\mathrm r}}\leq\operatorname{Sym}\mathscr A$. In particular, if $\underline G_{\alpha_X^{\mathrm r}}^{\alpha_Y^{\mathrm r}}\not\subseteq\mathrm A$, then $\mathscr A$ exhibits symmetry slippage.
\end{restatable}

\textcolor{blue}{For index-two extensions $H\triangleleft G$, extension-mistyping admits a particularly simple sufficient criterion; see Proposition~\ref{prop:extension-mistyping-special} in Appendix~\ref{sec:creation-proofs}. }

\section{Computing Symmetry Slippage}
\label{sec:computing}

Identifying symmetry slippage as in Section~\ref{sec:creating} does not require exactly computing $\operatorname{Sym}\mathscr{A}$, just the verification that $\operatorname{Sym}\mathscr{A} \not \subset \mathrm{A}$. Nevertheless, exact computation of $\operatorname{Sym}\mathscr{A}$ remains of interest in principle, for it is the only way to guarantee absence of symmetry slippage.

\begin{theorem}
\label{thm:hardness-main}
For every finite group \(H\), computing additional generators \(B\) satisfying \(\operatorname{Sym} \mathscr{A}= \langle \mathrm{A}, B \rangle\) is GI-hard, even when \(\mathrm{A}=\underline{H}_{\alpha_{X}}^{\alpha_{Y}} \leq \operatorname{Sym} \mathscr{A}\) is promised.
\end{theorem}


 The first step to proving Theorem~\ref{thm:hardness-main} returns to Subexample~\ref{subexample:aut-closure-finite} and shows the inclusion described therein is exact. The following lightly strengthens Theorem 2.1 in \cite{ravanbakhsh2017equivarianceparametersharing}.  

\begin{proposition}
\label{prop:graphaut-main}
With the setting of Subexample~\ref{subexample:aut-closure-finite}, $\operatorname{Sym}_{}(\mathscr{A})=\operatorname{Aut}(\Sigma)$ for one linear layer. The result maintains for a nonlinear layer $x \mapsto \mathrm{a}(W x)$ if $(r \in \mathbb{R}) \mapsto (t   \mapsto \mathrm{a}(tr))$ is injective.
\end{proposition}

Hardness makes us expect that deducing whether $\mathrm{A}=\operatorname{Sym}\mathscr{A}$ can in general necessitate architecture-specific knowledge. Since general architectures are usually built from modular pieces, it is natural to ask when $\operatorname{Sym}\mathscr{A}$ can be exactly computed from subarchitectures $\operatorname{Sym} \mathscr{A}_{(C_{i}, C_{j})}$, that is, when a converse to Proposition~\ref{prop:decomposing-tlaws} holds. The following result offers such conditions. Let $\mathscr{A}=\mathscr{A}_{(C_{L-1}, O)} \circ \dots \circ \mathscr{A}_{(I, C_{1})}$ be a decomposition of $\mathscr{A}$ for which $\mathcal{F}_{\mathscr{A}}=\mathcal{F}_{\mathscr{A}(C_{L-1}, {O})} \circ \dots \circ \mathcal{F}_{\mathscr{A}_{(I, C_{1})}}$. Assume $\boldsymbol \beta=(\beta_{v}:G \curvearrowright Z_{v})$ is an action on the spaces of $\mathscr{A}$ such that $\operatorname{Sym}\mathscr{A} \subset \beta_{I}(G) \times \beta_{O}(G)$. For $\mathcal F\subset Y^X$, write
$\mathcal F^\top:X\to Y^{\mathcal F}$ for the map
$x\mapsto(f(x))_{f\in \mathcal F}$.


\begin{theorem}

\label{thm:composition-converse-main}
For each $\ell$ suppose (i) there exists $\mathcal{F}_{\ell}' \subset \mathcal{F}_{\mathscr{A}_{(C_{\ell-1}, C_{\ell})}} \cap \operatorname{Map}_{G}(Z_{C_{\ell-1}}, Z_{C_{\ell}})$ such that $\mathcal{F}_{\ell}^{\prime\top}$ is injective, and (ii) for $\ell>1$, $fg \equiv gf$ for all $g \in G$ whenever $fgp=gfp$ for all $f \in \mathcal{F}_{\mathscr{A}_{(C_{\ell-1}, C_{\ell})}}, p \in \mathcal{F}'_{\ell - 1} \circ \dots \circ \mathcal{F}'_{1}$. Then  $\operatorname{Sym}_{}\mathscr{A}=\operatorname{Sym}_{}\mathscr{A}_{(C_{L-1}, O)} \circ \dots \circ \operatorname{Sym}_{}\mathscr{A}_{(I, C_{1})}$.
\end{theorem}

Conditions (i) and (ii) are complementary, roughly corresponding to appropriate notions of injectivity and surjectivity. If the $fg-gf$ were linear, the first condition is equivalent to $\bigcap_{f \in \mathcal{F}'_{\ell}}^{}\operatorname{ker } f=(0)$ and the second is $\operatorname{span}(\bigcup_{p \in \mathcal{F}'_{\ell-1} \circ \dots \circ \mathcal{F}_{1}'}\operatorname{im }p)=Z_{C_{\ell -1}}$. Of course, linearity basically never holds, and the stated hypotheses are the generalizations of these natural notions to the nonlinear setting. 

To exemplify Theorem~\ref{thm:composition-converse-main}, we apply it to compute $\operatorname{Sym}_{}(A\operatorname{-MLP})$ (Subexample~\ref{subexample:aut-closure-finite}) in closed-form. Before applying Theorem~\ref{thm:composition-converse-main}, we first compute $\operatorname{Sym}_{}\mathscr{A}$ in closed form for $\mathscr{A}$ a single equivariant layer. 

\begin{proposition}
\label{prop:AMLP-single-layer-main}
Let $\mathscr{A}:\mathbb{R}^{I} \to \mathbb{R}^{O}$ be a linear $A$-$\operatorname{MLP}$ layer as in Subexample~\ref{subexample:aut-closure-finite}, $A\neq \emptyset$. Put $D_{A}:=\langle A A ^{-1} \rangle$, $k:=[H :D_{A}]$. Then $\operatorname{Sym} \mathscr{A}\cong  D_{A} \wr S_{k}$. The conclusion maintains for a nonlinear layer $x \mapsto \mathrm{a}(W x)$ provided $(r \in \mathbb{R}) \mapsto (t   \mapsto \mathrm{a}(tr))$ is injective. 
\end{proposition}

Here, $\wr$ denotes the \emph{wreath product} $D_{A} \wr S_{k}:=D_{A}^{k} \rtimes S_{k}$ where $S_{k}$ acts on $D_{A}^{k}$ by permuting its factors. In the dense special case $A=H$, $D_{A}=H$ and $k=1$ and so $\operatorname{Sym}\mathscr{A} \cong H$.  The injectivity condition on $\mathrm{a}$ in Propositions~\ref{prop:graphaut-main} and \ref{prop:AMLP-single-layer-main} is not stringent; it is satisfied e.g. by $\operatorname{ReLU}$, sigmoid, $\operatorname{\tanh}$, leaky $\operatorname{ReLU}$, $\operatorname{ELU}$, etc.

\begin{proposition}
\label{prop:AMLP-multilayer-main}
Let $\mathscr{A}=\mathscr{A}_{(C_{L-1}, O)}\circ \dots \circ \mathscr{A}_{(I, C_{1})}$ be an $A$-$\operatorname{MLP}$ as in Subexample~\ref{subexample:aut-closure-finite} with activation $\mathrm{a}$, such that $(r \in \mathbb{R}) \mapsto (t   \mapsto \mathrm{a}(tr))$ is injective. Put $D_{A}:=\langle A A ^{-1}\rangle$, $k:=[H :D_{A}]$. Then $$\operatorname{Sym}\mathscr{A}\cong D_{A} \wr S_{k}.$$
\end{proposition}

It is of future interest to derive further closed-form characterizations of $\operatorname{Sym}\mathscr{A}$ in other settings.

\section{Correcting Symmetry Slippage}
\label{sec:correcting}

\subsection{Grippage Encodings}
\label{sec:GEs}
We next discuss a simple architectural modification that restores expressivity loss incurred by symmetry slippage. Let us say that a modification $\mathscr{A}'$ of $\mathscr{A}$ \emph{grips} a slipped transformation $s=(\sigma_{s}, \tau_{s}) \in \operatorname{Sym}\mathscr{A}-\mathrm{A}$ if $s \notin \operatorname{Sym}\mathscr{A}'$, i.e., there exists  $f' \in \mathcal{F}_{\mathscr{A}'}$ such that $f' \sigma_{s} \neq \tau_{s} f'$. In this section, we study a lightweight intervention, \emph{grippage encodings (GEs)}, for gripping slipped transformations by appending particular features. Throughout, we assume there is a ``usual way" to append feature maps to $\mathscr{A}$. To be precise, we assume that for any cut $C$, there is a rule assigning any $c:X \to Z$ to an architecture $\mathscr{A} \oplus_{C} c$ whose functions factor as $X \inarrow{(b,c)}Z_{C} \times Z \inarrow{e'}Y$ and for which $\mathscr{A}\oplus_{C} 0=\mathscr{A}$. This rule is usually self-evident. For example, if $C$ is a layer of $\mathscr{A}=\operatorname{MLP}$, the rule replaces $z \mapsto \phi(Wz)$ with $\big( (b(x), c(x)) \big) \xmapsto{}\phi(Wb(x) + Uc(x))$ for $U$ trainable. 

\begin{definition}
\label{def:GE}
A \emph{grippage encoding for $S \subset \operatorname{Sym}\mathscr{A}-\mathrm{A}$ at $C$} is a map $c:X \to Z$ for which $\mathscr{A}\oplus_{C} c$ grips every $s \in S$. 
\end{definition}



GEs are of course not the only method by which slipped transformations can be gripped. For certain over-quotienting and extension-mistyping examples in particular, there are bespoke repairs which are just as simple. GEs remain attractive, however, in that their analysis is amenable to general principles. We codify these principles in the following proposition, which greatly streamlines proving GEs in our case studies, where their implementation can be subtle depending on the creation mechanism for $S$.





\begin{proposition}[Informal]
\label{prop:main-GE-prop}
    $c:X \to Z$ is a GE for $S$ if any of the following hold:
\begin{enumerate}[leftmargin=1.5em]
    \item Group-closure created $S$ and \(c \alpha_{X}(g) \neq gc\) for every \(s=(\alpha_{X}(g), \alpha_{Y}(g)) \in S\).\footnote{Recall that transformations slipped by group-closure take the form $s =\big(\alpha_{X}(g), \alpha_{Y}(g) \big)$ for $g \in G-H$, }
    
    \item \textcolor{blue}{Over-quotienting created $S$ and $c\sigma\neq c$
for every $s=(\sigma,I_Y)\in S$.}

    \item Extension-mistyping created $S$ and there exist $\kappa^{\text{p}},\kappa^{\text{r}}:G\curvearrowright Z$, compatible with $\mathrm{A}$ and $\mathscr{A}$ respectively, such that
    $\operatorname{im}c\not\subset\operatorname{Eq}\bigl(\kappa^{\text{p}}(g),\kappa^{\text{r}}(g)\bigr)$
    for every $s_g\in S$. If in fact $[G:H]=2$, $\kappa^{\text{r}}|_H=\kappa^{\text{p}}|_H$, and $\kappa^{\text{p}}(s)=-\kappa^{\text{r}}(s)$ for some $s\in G-H$, it suffices that $c\neq0$.
\end{enumerate}
\end{proposition}

\begin{remark}
\label{rmk:ge-remark}
\vspace{.2em}
\leavevmode
\vspace{-.5em}
\begin{itemize}[leftmargin=1.5em]
    \item  GEs can also be used to naturally modify an architecture's realized law with minimal retraining when the aforementioned bespoke methods often cannot; Case Study~\ref{case:ecg-normalization} includes an example. 
\item Abstractly, the purpose of GEs is to modify one obeyed transformation law \(\mathrm{L}_{1}\) into another transformation law \(\mathrm{L}_{2}\). For us, the ambient interpretation is that  \(\mathrm{L}_{1}=\operatorname{Sym}\mathscr{A}\) is a realized transformation law, while \(\mathrm{L}_{2}=\mathrm{A}\) is a prescribed one. However, in principle \(\mathrm{L_{1}}\) and \(\mathrm{L_{2}}\) can be any pair of laws, and GEs can be used to modulate between them. This may be useful e.g. for adapting the equivariance properties of pretrained models to different tasks or targets \cite{goel2026any}. Case Study~\ref{case:ecg-normalization} includes an example of this principle.
\end{itemize}

 \end{remark}

\subsection{Experiments}
\label{sec:experiments}

We offer 9 case studies in which our creation mechanisms identify, and GEs provably repair, symmetry slippage within an active area. We overview each case study here, deferring details to Appendix~\ref{app:case-studies}. Table~\ref{tab:ge-summary} summarizes the effects of GEs across the case studies. 

\subsubsection{Case Studies for Group-Closure}
\label{sec:case-studies-main-group-closure}
\casestudyparagraph{Time-Orientation in Lorentz-Equivariant ML.} \label{case:lorentz}In areas such as experimental high-energy particle physics (HEP), architectures seek equivariance to the (proper orthochronous) Lorentz group $\operatorname{SO}^{+}(1,d)$ of spacetime-orientation-preserving maps that respect the Minkowski metric on $\mathbb{R}^{1,d}$ \citep{bogatskiy2020lgn,gong2022lorentznet,spinner2024lgatr,bogatskiy2024pelican,batatia2023gmace}. Existing techniques for $\operatorname{SO}^{+}(1,d)$-equivariant learning use exclusively polynomial features; since $\operatorname{SO}^{+}(1,d)$ is Zariski-dense in $\operatorname{SO}(1,d)$, this can be shown to entail symmetry slippage under Example~\ref{ex:zariski-closure} and Proposition~\ref{prop:group-closure}. Conceptually, these architectures $\mathscr{A}$ discard time-orientation information. While this information is not relevant on standard particle physics tasks, it is relevant on any task involving directed relations such as displacements in $\mathbb{R}^{1,d}$. To demonstrate this, we compare four Lorentz-equivariant architectures with and without GEs on a neural PDE task, namely, learning the retarded solution to the wave equation $\square u= s$ in 2+1 dimensions \citep{LuEtAl2021, LiEtAl2021, KovachkiEtAl2023}. 

\casestudyparagraph{Signs in Units-Equivariant ML.} \label{case:units}Physical laws should not depend on arbitrary choices of measurement units, motivating units-equivariant learning. A common construction, based on the Buckingham-$\Pi$ theorem~\citep{Buckingham1914}, first maps measured quantities to dimensionless Laurent-monomial features and then applies an unconstrained architecture \citep{villar2023dimensionless}. The physical group consists of positive changes of base units, $H=(\mathbb{R}^{+})^{k}$, but $H$ is Zariski-dense in $G=(\mathbb{R}^{\times})^{k}$; consequently, such algebraic dimensionless features are also invariant to nonphysical sign reversals of the base units, giving symmetry slippage under Example \ref{ex:zariski-closure} and Proposition \ref{prop:group-closure}. To demonstrate this, we compare units-equivariant architectures with and without GEs on a forecasting task with the Mauna Loa $\operatorname{CO}_{2}$ dataset~\citep{Keeling1976,RasmussenWilliams2006}, using previous concentration changes to predict future ones. 

We further observe that Zariski-density of $(\mathbb{R}^{+})^{k}$ in $(\mathbb{R}^{\times})^{k}$ appears also in weight-space learning for ReLU networks, causing polynomial path representations \citep{meng2019gsgd,gonon2025rescaling} to inherit unprescribed sign-invariance by Example~\ref{ex:zariski-closure}. This gives a group-closure interpretation of existing heuristics; in particular, the practice of explicitly retaining sign information acts as a GE for the entailed over-equivariance \citep{meng2019gsgd,yi2019positively,gonon2025rescaling}.

\casestudyparagraph{Inner Automorphisms in Lie Algebra Learning.} \label{case:intaut} Lie-algebra-valued data arise in geometric learning problems such as dynamical-system modeling, registration, and homography-based vision
\citep{LinEtAl2024LieNeurons}. It is often desirable that architectures learning on the Lie algebra $(\mathfrak{k}, [\_, \_])$ of a Lie group $K$ be equivariant to the \emph{Adjoint representation} $\operatorname{Ad}:K \to \operatorname{Aut}\mathfrak{k}$; conceptually, this encodes a form of coordinate-independence. This turns out equivalent to being $H:=\operatorname{Int}(\mathfrak{k})$-equivariant, where $\operatorname{Int}(\mathfrak{k})$ is the \emph{inner automorphism group} of $\mathfrak{k}$. An attractive approach to implementing $\operatorname{Ad}$-equivariance is offered by Lie Neurons \citep{LinEtAl2024LieNeurons, KimEtAl2026ReLN}. However, Lie Neuron architectures turn out to be built from operations that are not only $H$-equivariant, but $G$-equivariant, where $G=\operatorname{Aut}(\mathfrak{k})$ is the full automorphism group of $\mathfrak{k}$, yielding symmetry slippage under Example~\ref{example:automorphism-closure} and Proposition~\ref{prop:group-closure}. We compare the architectures with and without GEs on a standard task in data-driven constitutive modeling \citep{dammass2025invariants, holthusen2026treloar}. 

\subsubsection{Case Studies for Over-Quotienting}

\casestudyparagraph{Roots in Glycan Learning.} \label{case:glycan} Glycans are carbohydrate molecules made of simple sugars linked by chemical bonds. Many glycans have a \emph{reducing end}, a distinguished sugar that contains a free anomeric carbon. These glycans are abstracted as trees whose nodes are sugars and whose edges are bonds, who are rooted at the reducing end, and whose nodes are colored by sugar type and edges by bond type. Recent glycan-learning architectures such as GlycanML \citep{xu2025glycanml} and GlycanAA \citep{xu2025glycanaa} use graph-based representations of these molecules that over-quotient the reducing end position, causing symmetry slippage per Proposition~\ref{prop:over-quotienting} (although, unlike the group-closure case studies above, it is not over-equivariance). We compare GlycanML and GlycanAA with and without GEs on the domain prediction benchmark found in both papers. 

\casestudyparagraph{Normalization in ECG ML.}\label{case:ecg-normalization} Amplitude normalization, which may be viewed as (at least) $G:=\mathbb{R}^{+}$-invariance, is a common practice in ECG ML \citep{ma2021ecg,weimann2024federated,zeng2026stabilizing,hu2026one,li2025ecgfounder}. While deliberate and often desirable, normalization over-quotients certain targets that prescribe a trivial transformation law $\mathrm{A}=\underline{H}_{\alpha}^{I}=\underline{( e )}_{\alpha}^{I}=\{ (I_{X}, I_{Y}) \}$, incurring symmetry slippage via Proposition~\ref{prop:over-quotienting}. To demonstrate this, we evaluate the models \citep{weimann2024federated, li2025ecgfounder} with and without GEs on the well-known PTB-XL benchmark \citep{wagner2020ptbxl, strodthoff2021ptbxl} on which they evaluate. Our first experiment demonstrates the slippage on a multilabel classification task restricted to two known amplitude-sensitive labels (Table~\ref{tab:ge-summary}). Our second experiment appends GEs to the papers' released pretrained architectures for the full multilabel task that includes targets for which normalization is desirable and for which it is not; GEs modestly but replicably improve performance here (see Appendix~\ref{app:case-studies}). In this way, this case study draws upon our secondary GE perspective (Remark~\ref{rmk:ge-remark}), that appending GEs to a pretrained foundation model allows the model's transformation law to be adjusted based on the task. 

\casestudyparagraph{Polarity in Time-Frequency Signal Processing.} \label{case:cwd}
The Choi-Williams distribution is a classical quadratic time-frequency representation widely applied to nonstationary signal analysis \citep{choi1989improved}. Being quadratic, $\operatorname{CWD}(x)$ is invariant under polarity reversal, over-quotienting the prescribed $\operatorname{Aff}^{+}(1, \mathbb{R})$-invariance of an established framework for $\operatorname{CWD}$-based ECG analysis \citep{kayikcioglu2020time} to $\operatorname{Aff}(1, \mathbb{R})$-invariance. This becomes problematic e.g. when the task is sensitive to ST elevation and ST depression, which are clinically distinct phenomena \citep{wagner2009aha}. This is the case for the task in \cite{kayikcioglu2020time}. We evaluate the architecture with and without GEs on the European ST-T database \citep{taddei1992european} task of distinguishing ST-elevation and ST-depression episodes. 

\subsubsection{Case Studies for Extension-Mistyping}

\casestudyparagraph{(Pseudo)scalar mistyping in neural operators.} \label{case:G-FNO}$H:=C_{4}$ acts on functions $\omega:\mathbb{R}^{2}\to \mathbb{R}$ as $g \omega(x):=\omega(g ^{-1} x)$. This action extends in two natural ways to $G:=D_{4}$. One is via $g \omega(x):=\omega(g ^{-1} x)$, in which case $\omega$ is a \textit{scalar field }such as pressure; another is via $g\omega(x):=\det(g)\omega(g ^{-1} x)$, in which case $\omega$ is a \textit{pseudoscalar}, such as vorticity ($\operatorname{curl}$). Popular $G$-equivariant neural operator approaches such as \citet{helwig2023group}, however, mistype the extension so that e.g. vorticity is treated as an ordinary scalar field, creating symmetry slippage under Proposition \ref{prop:extension-mistyping}. We evaluate on two Navier–Stokes benchmarks, including NS-SYM \citep{helwig2023group}.

\casestudyparagraph{Tetrahedral Chirality in Molecular Graph Learning.} \label{case:QGeoGNN}  A common source of molecular chirality is a tetrahedral atom bonded to four distinct ligands, admitting two configurations conventionally labeled $R$ and $S$. Since ligand ordering is arbitrary, $S_{4}$ acts by permuting them with even permutations $A_{4}$ preserving handedness and odd permutations reversing it, so a stereochemical encoding defined relative to this ordering should transform according to the sign representation \citep{pattanaik2020message}. However, QGeoGNN, a popular GNN approach for chromatographic retention-time prediction \citep{xu2023retention}, extends the representation on $A_{4}$ to the trivial representation on $S_{4}$, creating symmetry slippage by extension mistyping per Proposition~\ref{prop:extension-mistyping}. We compare QGeoGNN with and without GEs on its all-column retention-time benchmark \citep{xu2023retention}. 

\casestudyparagraph{Pseudovector Mistyping in Magnetic Equivariant ML.} \label{case:magnet} Magnetic moments and spins are pseudovectors, transforming under  $G:=\operatorname{O}(3)$ as $\widetilde{\rho}^{\text{p}}(Q)w:=\det(Q) Qw$. The distinction vanishes when only $\operatorname{SO}(3)=:H \triangleleft G$ is meaningfully acting, but becomes relevant to magnetic equivariant learning, where pseudovectors appear alongside ordinary vectors like atomic positions and displacements~\citep{yuan2024magnet,yu2024spingnnpp,cho2026sparsitypromoting}. However, such architectures often extend the $\operatorname{SO}(3)$-action to the vector rather than pseudovector action of $\operatorname{O}(3)$, causing extension-mistyping per Proposition~\ref{prop:extension-mistyping}. We compare two architectures~\citep{yuan2024magnet,cho2026sparsitypromoting} with and without GEs on a noncollinear anomalous-Hall prediction task \citep{kipp2024hall}. 

\begin{table*}[t]
\centering
\small
\setlength{\tabcolsep}{1.5pt}

\caption{
Effect of GEs across \(9\) case studies.
For each metric $M$, we define
$\Delta M := M_{\mathrm{GE}}-M_{\mathrm{baseline}}$. The two $+\mathrm{GE}$ rows give the range of effects across the evaluated architectures.}
\vspace{-10pt}
\label{tab:ge-summary}

\begin{tabular}{@{}l BBB OOO GGG@{}}
\toprule

Case Study
& \caseref{case:lorentz}
& \caseref{case:units}
& \caseref{case:intaut}
& \caseref{case:glycan}
& \caseref{case:ecg-normalization}
& \caseref{case:cwd}
& \caseref{case:G-FNO}
& \caseref{case:QGeoGNN}
& \caseref{case:magnet}
\\

\midrule

$\Delta$ Metric
&
NMSE $\downarrow$
&
Acc. (pp) $\uparrow$
&
{\(\mathcal{L}\) $\downarrow$}
&
Macro-F1 $\uparrow$
&
AUROC $\uparrow$
&
AUROC $\uparrow$
&
Rel.\! MSE $\downarrow$
&
RMSE $\downarrow$
&
RMSE $\downarrow$
\\[2pt]

$+\mathrm{GE}$ (min)
&
$-0.571$
&
$+23.6$
&
$-0.268$
&
$+0.036$
&
$+0.015$
&
$+0.117$
&
$-0.662$
&
$-0.145$
&
$-0.041$
\\

$+\mathrm{GE}$ (max)
&
$-0.458$
&
$+23.6$
&
$-0.242$
&
$+0.046$
&
$+0.066$
&
$+0.127$
&
$-0.662$
&
$-0.145$
&
$-0.039$
\\

\bottomrule
\end{tabular}
\vspace{-20pt}
\end{table*}

\section{Discussion}
\label{sec:discussion}

This paper identified three mechanisms creating symmetry slippage across nine research areas and established conditions for computing \(\operatorname{Sym}\mathcal A\)from subarchitectures, despite the problem's general hardness. We applied these conditions to obtain the first exact symmetry characterization of \(A\)-\(\operatorname{MLPs}\), a class subsuming e.g. \(G\)-CNNs \citep{ravanbakhsh2017equivarianceparametersharing, cohen2016group}. 
To correct symmetry slippage, we introduced grippage encodings, and demonstrated them as a repair across our 9 case studies. A limitation of GEs is that they do not come with expressivity guarantees beyond promising that a targeted subset $S \subset \operatorname{Sym}\mathscr{A}-\mathrm{A}$ has been removed, $S \cap  \operatorname{Sym}\mathscr{A}'=\emptyset$. In particular, they do not guarantee $\operatorname{Sym}\mathscr{A}'=\mathrm{A}$, perhaps unsurprisingly in light of Theorem~\ref{thm:hardness-main}. Furthermore, while the all-parameter nature of symmetry slippage makes existence a strong statement about an architecture, the GE condition is inversely weak, requiring only that \emph{some} functions $f^{\theta} \in \mathcal{F}_{\mathscr{A}'}$ disobey $S$. Thus not all GEs are equivalent, and their efficacy can depend on how $C$ and $e^{\theta'}$ are chosen. Nevertheless, we find GEs remain natural, non-bespoke, lightweight, and effective in restoring performance lost by symmetry slippage in many cases. This includes cases where they modify the transformation law of a pretrained model for which retraining is costly or otherwise undesirable, as when symmetry requirements are task-dependent \citep{goel2026any}.

Scientific machine learning is entering a differentiation phase \citep{polikarov1995concerning} in which increasingly specalized architectures, data modalities, and scientific tools are being assembled into increasingly modular workflows \citep{smidt2026physics}. In this era, we anticipate symmetry slippage to become both more salient and more subtle. While we expect that general principles for proving symmetry slippage's absence are rarer than general principles proving its creation, it is possible to envisage a broad program in exactly computing $\operatorname{Sym}\mathscr{A}$ for diverse architecture classes $\mathscr{A}$, allowing for provably exact equivariance to be deployed in complicated scientific workflows.

\newpage
\subsection*{AI use statement}

We used generative AI to generate synthetic data sets, implement methods, and clean and reformat datasets. We used generative AI to assist in proofwriting by way of proofreading and polishing. We largely did not use generative AI to provide critical ingredients for proving mathematical claims; an exception is Theorem~\ref{thm:hardness-main}, for which our initial approach was flawed and generative AI provided some useful, albeit noisy, insights that ultimately helped repair it.  We additionally used generative AI to assist in literature search. All LLM-generated code was checked by the authors, and no paper (LLM-retrieved or not) was referenced without the authors' understanding of its content. We did not use generative AI to help develop theoretical models or conceptual frameworks, formulate mathematical claims, to assist with translation, to propose or refine hypotheses, to design or provide feedback on research methodology or experiments, support qualitative and thematic data analysis, or interpret results. We take responsibility for the final content of this work.




\subsection*{Reproducibility statement}

The proofs of all theoretical results may be found in Appendix~\ref{app:theory}. Full case study discussions may be found in Appendix~\ref{app:case-studies}. Experimental details may be found in Appendix~\ref{app:experimental-details}. 



{\footnotesize
\bibliography{references}
}
\bibliographystyle{references}

\newpage
\appendix
\tableofcontents
\newpage
\section{Related Work}

\label{sec:related-work}
\paragraph{Over-Equivariance} The most-studied example of symmetry slippage occurs when $\mathrm{A}:=\underline{H}_{\alpha_{X}}^{\alpha_{Y}}$ is prescribed for some $H \leq \Gamma$, but a group $\Gamma \geq G \gneq H$ exists such that $\underline{H}_{\alpha_{X}}^{\alpha _{Y}} \subsetneq \underline{G}_{\alpha _{X}}^{\alpha _{Y}}\leq \operatorname{Sym}\mathscr{A}$; forms of this principle have have been called e.g. \textit{excessive equivariance} \citep{han2022learning,zheng2024relaxing} and \textit{nonunique equivariance} \cite{ravanbakhsh2017equivarianceparametersharing, maron2019deep}. We adopt the term \emph{over-equivariance}. Of our three creation mechanisms, group-closure always entails symmetry slippage by over-equivariance, over-quotienting sometimes does, and extension-mistyping never does. A familiar story of over-equivariance involves chirality in molecular learning: \citep{dumitrescu2025eequivariant, weaver2025chi}. The majority of such works follow a contour in which over-equivariance is identified by domain- and architecture-specific means, and a repair is tailored accordingly. In contrast, the present work considers the complementary question of how domain- and architecture-agnostic design choices can create over-equivariance (and more general symmetry slippage) to begin with, together with characterizing computational considerations and establishing general repair principles. The clearest precedent for such a program is the lineage \citep{ravanbakhsh2017equivarianceparametersharing, maron2019universality}, who systematically examined over-equivariance in the classical setting where $\mathscr{A}$ is built from morphisms of finite-group permutation representations. This paper may be viewed as a generalization of these works to the setting of contemporary equivariant ML. In particular, we identify the setting of \citet{ravanbakhsh2017equivarianceparametersharing} as a doubly-special case of group-closure in Section~\ref{sec:local-symmetry-slippage-creation-mechanisms}, and in Section~\ref{sec:computing} we strengthen their computational results. Other precedents include \citet{finzi2021practical}, who characterize necessary (but, as we shall see, not sufficient) conditions for arbitrarily matrix-group equivariance, and \citet{wang2023general}, who analyze the learning-theoretic costs of over-equivariance. 

When symmetry slippage is caused by over-equivariance, our GEs may be viewed as symmetry-breaking features~\citep{kaba2023symmetry,xie2024equivariant,goel2026any,lawrence2025improving,smidt2021finding}
. The two nevertheless concern different aspects of symmetry and neither subsumes the other. On the one hand, symmetry slippage is an "architecture-level property", in the sense that belonging to $\operatorname{Sym}\mathscr{A}$ entails transformation-law obedience of every $f^{\theta}$, $\theta \in \Theta$. In contrast, much symmetry breaking literature instead studies the refined question of how symmetry behaves pointwise under a fixed $f^{\theta}$. This refined notion also underlies recent \emph{symmetry increase} studies \citep{cen2024high, lin2026reducing}, which explore how equivariance forces individual inputs to map to outputs with strictly larger stabilizers, and how this increase can be reduced. On the other hand, not all symmetry slippage is over-equivariance, and accordingly not all GEs may be viewed as symmetry-breaking. To see an example of a well-known instance of symmetry slippage that is not over-equivariance, we turn to graph representation learning \cite{scarselli2009graph, gilmer2017neural, hamilton2018inductiverepresentationlearninglarge}. 

\paragraph{Symmetry Slippage Beyond Over-Equivariance}
There is a natural action $\alpha_X$ of $\Gamma:=S_n$ on the space $X=X_n$ of $n$-node graphs by node relabeling, and the most basic class of graph neural networks $\mathscr A:X\rightsquigarrow Y$ are designed to be invariant under $\alpha_X$, i.e. invariant under graph isomorphism. In our notation, the transformation law $\mathrm A:=\underline H_{\alpha_X}^{I_Y}$, where $H:=\Gamma=S_n$, is prescribed. Equivalently, $\mathrm A$ prescribes that every realization $f^\theta:X\to Y$ is constant on the classes of the orbit relation $\sim_{S_n}$, and hence factors uniquely through $X/S_n$. It is known that message-passing graph neural networks $\mathscr A$ \citep{gilmer2017neural} are no more expressive than a coarser equivalence relation on $X$ given by the Weisfeiler-Lehman Isomorphism Test \citep{Weisfeiler1968}; we write $G\sim_{\mathrm{WL}}G'$ when WL fails to distinguish $G$ and $G'$. Isomorphic graphs are WL-indistinguishable, so $\sim_{S_n}\subset\sim_{\mathrm{WL}}$, inducing a map $q:X/S_n\to X/{\sim_{\mathrm{WL}}}$ that is surjective but not generally injective. If $G\sim_{\mathrm{WL}}G'$, then $f^\theta(G)=f^\theta(G')$ for every realization $f^\theta\in\mathcal F_{\mathscr A}$. Thus every realization factors uniquely through $X/{\sim_{\mathrm{WL}}}$, and hence as $X\to X/S_n\overset q\to X/{\sim_{\mathrm{WL}}}\to Y$.

We claim this is over-quotienting in the sense of Definition~\ref{def:over-quotienting}, but not over-equivariance. Take $C_1=I$, $C_{2}=O$, and take $\sim:=\sim_{\mathrm{WL}}$. Choose graphs $G_0,G_1\in X$ with $G_0\sim_{\mathrm{WL}}G_1$ but $G_0\not\sim_{S_n}G_1$. With $J=\operatorname{Bij}(X)$, define $\sigma\in J$ to swap $G_0$ and $G_1$ and fix every other graph. Then $G\sim_{\mathrm{WL}}\sigma(G)$ for every $G\in X$. Since every $f^\theta$ is constant on WL-classes, $f^\theta\sigma=f^\theta$, and hence $(\sigma,I_Y)\in\operatorname{Sym}\mathscr A$. On the other hand, $(\sigma,I_Y)\notin\mathrm A$. Indeed, every input component of an element of $\mathrm A$ is a vertex relabeling $\alpha_X(\pi)$, which can send a graph only to an isomorphic graph, whereas $\sigma$ sends $G_0$ to the nonisomorphic graph $G_1$. Thus $\sigma\notin\alpha_X(S_n)$, and therefore $(\sigma,I_Y)\in\operatorname{Sym}\mathscr A-\mathrm A$. Therefore, vanilla message-passing GNNs exhibit symmetry slippage by over-quotienting in light of Proposition~\ref{prop:over-quotienting}. This slippage is not over-equivariance, however. Indeed, here one has $H=\Gamma=S_n$, while over-equivariance requires group extension $H\lneq G\leq\Gamma$. 

Our case studies provide further examples where symmetry slippage beyond over-equivariance occurs.

\section{Elaborations Upon and Proofs of Results}
\label{app:theory}

In this appendix section, we provide the precise statements and proofs of results referenced in the main text. The majority of results from Appendices~\ref{sec:creation-proofs} and \ref{sec:GE-proofs} are straightforward once the relevant definitions are in place. The results from Section~\ref{sec:computation-proofs} are considerably less straightforward. For a family $\mathcal F\subset Y^X$, recall that we write
$\mathcal F^\top:X\to Y^{\mathcal F}$ for the map
$x\mapsto(f(x))_{f\in \mathcal F}$; for an architecture $\mathscr A$, write
$\mathscr A^\top:=\mathcal F_{\mathscr A}^\top$.

\subsection{Zariski Group-Closure}
\label{app:Zariski-group-closure}
{\color{blue}
We substantiate Example~\ref{ex:zariski-closure} assuming no background in algebraic geometry. The punchline of Example~\ref{ex:zariski-closure} was that equivariance of e.g. polynomial or rational maps under the identity component of groups such as \((\mathbb{R}^\times)^d\), \(\operatorname{SO}(p,q)\), and \(\operatorname{GL}(d, \mathbb{R})\) immediately implies equivariance under all of \(G\). Any architecture prescribed to be equivariant solely on these identity components (this usually corresponds to preserving some notion of orientation), then, exhibits symmetry slippage if it is built from e.g. polynomial features. 

To motivate, consider the fact that any polynomial identity in $t$ that holds for all $t>0$ (i.e. for $t \in \mathbb{R}^{+}$) in fact holds for all $t$.\footnote{Indeed, if $P(t)=Q(t)$ is the identity in question, then $R(t):=P(t)-Q(t)$ is zero for all $t>0$. Thus $R$ is a polynomial with infinitely many roots, so $R \equiv 0$. This property reflects the fact that polynomials are incredibly rigid; even simple non-polynomial equations such as $|t|=t$  do not have this property.} \textit{Zariski closure} generalizes this observation to more complicated groups. A\textit{ (real, linear) algebraic group} is, roughly, a group $G$ described by polynomial equations. Letting $V(S)$ denote the set of simultaneous zeros of a polynomial collection $S$, examples of real algebraic groups include $\operatorname{O}(n)=V(\{ X^{\top}X - I \})$, $\operatorname{SO}(p,q)=V(\{ A^{\top}J A - J  , \det A - 1\})$, $J=\text{diag}(I_{p}, -I_{q})$, $\mathbb{R}^{\times}\cong V(\{ ts- 1 \})$. A \textit{regular function} on such a group is a rational function with no (complex) poles on the group. For example, regular functions on $\operatorname{SO}(p,q)$ are just (the restrictions of) polynomials in the ambient coordiantes. Regular functions on $\mathbb{R}^{\times}$ are \textit{Laurent polynomials}, that is polynomials in the variables $t, s=t^{-1}$ (there is a pole at $t=0$, but $0$ is not in the group). 

A subgroup $H \leq G$ is \textit{Zariski-dense} in $G$ if for every regular function $f$ on $G$, $f(h)=0$ for all $h \in H$ implies $f(g)=0$ for all $g \in G$. Put differently, $H$ satisfies no additional polynomial identities beyond those satisfied by $G$. As the name suggests, Zariski density is equivalent to density $\overline{H}^{\text{Zariski}}=G$ in the \textit{Zariski topology (of varieties)}, defined as the topology whose closed sets are all of the $V(S)$. 

Example \ref{ex:zariski-closure} is based on the following fact.

\begin{proposition}[Zariski group-closure]
Suppose $G$ is the group of real points of a real algebraic group
$\mathbf{G}$, acting on $X$ and $Y$, with $Y$ a vector space. Let $H\leq G$ and $\mathcal{C} \subset  Y^X$ be a class of functions such that, for every $\phi\in\mathcal{C}$ and $z\in X$, the map $g\mapsto \phi(gz)-g\phi(z)$ is regular on $G$. If $\overline{H}^{\mathrm{Zariski}}=G$, then
$\mathcal{C}$ closes $H$ to $G$: $\mathcal{C}\cap\operatorname{Map}_{H}(X, Y)=\mathcal{C}\cap\operatorname{Map}_G(X,Y).$ 
\end{proposition}

\begin{proof}
Let $\phi\in\mathcal{C}\cap\operatorname{Map}_H(X,Y)$. Fix $z\in X$ and consider the regular map
$D_{\phi,z}:G\to Y,g\mapsto\phi(gz)-g\phi(z)$. Since $\phi$ is $H$-equivariant, $D_{\phi,z}$ vanishes on $H$. By Zariski density, it therefore vanishes
on all of $G$. Since $z$ was arbitrary, $\phi$ is $G$-equivariant, proving
$$
\mathcal{C}\cap\operatorname{Map}_H(X,Y)
\subseteq
\mathcal{C}\cap\operatorname{Map}_G(X,Y).
$$
The reverse inclusion is immediate from $H\leq G$.
\end{proof}

A useful source of Zariski-dense subgroups is the identity component $G_0$ of a Zariski-connected real algebraic group $G$. Relevant examples include $\operatorname{SO}_{0}(p,q) \subset \operatorname{SO}(p,q)$, $p,q \geq 1$, (including the proper orthochronous Lorentz group $\operatorname{SO}^{+}(1,d)s \subset \operatorname{SO}91,d$), $\operatorname{GL}^{+}(d, \mathbb{R}) \subset \operatorname{GL}(d,\mathbb{R})$, and $(\mathbb{R}^{+})^{d} \subset (\mathbb{R}^{\times})^{d}$. A nonexample is $\operatorname{SO}(n) \subset \operatorname{O}(n)$; and indeed, $\operatorname{O}(n)$ is not Zariski-connected. We do not prove these facts; see \citet{borel1991linear} for a reference. 

Proposition \ref{prop:group-closure-poly-to-analytic} extends this group-closure phenomenon from polynomial maps to real-analytic maps for finite-dimensional representations.}

\subsection{Proofs from Section~\ref{sec:local-symmetry-slippage-creation-mechanisms}}
\label{sec:creation-proofs}

\decomposingtlaws*

\begin{proof}
Let $(\sigma, \tau) \in\operatorname{Sym}\mathscr{A}_{(C_{2}, O)} \circ \operatorname{Sym}\mathscr{A}_{(C_{1} , C_{2})} \circ \operatorname{Sym} \mathscr{A}_{(I, C_{1})}$, so that there exist $\delta_{1}$, $\delta_{2}$ such that $(\sigma,\delta_{1})\in\operatorname{Sym}\mathscr{A}_{(I,C_{1})}$, $(\delta_{1},\delta_{2})\in\operatorname{Sym}\mathscr{A}_{(C_{1},C_{2})}$, and $(\delta_{2},\tau)\in\operatorname{Sym}\mathscr{A}_{(C_{2},O)}$. As described in Section~\ref{sec:decomposing-transformation-laws}, any $\theta \in \Theta$ gives rise to $f^{\theta}=e^{\theta}\Phi^{\theta}a^{\theta}$, where $a^{\theta}$, $\Phi^{\theta}$, and $e^{\theta}$ are the maps realized by $\mathscr{A}_{(I,C_{1})}$, $\mathscr{A}_{(C_{1},C_{2})}$, and $\mathscr{A}_{(C_{2},O)}$, respectively. Then $f^{\theta}\sigma=e^{\theta}\Phi^{\theta}a^{\theta}\sigma=e^{\theta}\Phi^{\theta}\delta_{1}a^{\theta}=e^{\theta}\delta_{2}\Phi^{\theta}a^{\theta}=\tau e^{\theta}\Phi^{\theta}a^{\theta}=\tau f^{\theta}$. Hence $(\sigma,\tau)\in\operatorname{Sym}\mathscr{A}$, hence $\operatorname{Sym}\mathscr{A}_{(C_{2}, O)} \circ \operatorname{Sym}\mathscr{A}_{(C_{1} , C_{2})} \circ \operatorname{Sym} \mathscr{A}_{(I, C_{1})} \leq \operatorname{Sym}\mathscr{A}$. 

To see that equality need not hold, consider the following counterexample. Let $V=\{i,c_{1},c_{2},o\}$ with $I=\{i\}$, $C_{1}=\{c_{1}\}$, $C_{2}=\{c_{2}\}$, $O=\{o\}$, and $E=\{(i,c_{1}),(c_{1},c_{2}),(c_{2},o)\}$. Let $Z_{v}=\mathbb{R}^{3}$ for every $v\in V$, let $\Gamma=S_{3}$, and let $\boldsymbol{\alpha}=(\alpha_{v}:\Gamma\curvearrowright Z_{v})_{v\in V}$ each be the standard representation of $S_{3}$ on $\mathbb{R}^{3}$. Take $J=\alpha_{I}(\Gamma)$ and $K=\alpha_{O}(\Gamma)$. Let $\Theta=S_{3}$ and define the realizations of the three subarchitectures by $a^{\pi}=P_{\pi}$, $\Phi^{\pi}=\operatorname{id}_{\mathbb{R}^{3}}$, and $e^{\pi}=P_{\pi}^{-1}$. Then $f^{\pi}$ realized by $\mathscr{A}$ satisfies $f^{\pi}=e^{\pi}\Phi^{\pi}a^{\pi}=\operatorname{id}_{\mathbb{R}^{3}}$, and hence $\operatorname{Sym}\mathscr{A}=\{(P_{\sigma},P_{\sigma}):\sigma\in S_{3}\}$. On the other hand, if $(P_{\sigma},P_{\tau})\in\operatorname{Sym}\mathscr{A}_{(I,C_{1})}$, then $P_{\pi}P_{\sigma}=P_{\tau}P_{\pi}$ for every $\pi\in S_{3}$. Taking $\pi=e$ gives $\tau=\sigma$, after which $\sigma$ must commute with every element of $S_{3}$. But $Z(S_{3})=\{e\}$, s0 $\operatorname{Sym}\mathscr{A}_{(I,C_{1})}=\{(\operatorname{id},\operatorname{id})\}$. The same argument gives $\operatorname{Sym}\mathscr{A}_{(C_{2},O)}=\{(\operatorname{id},\operatorname{id})\}$, while $\operatorname{Sym}\mathscr{A}_{(C_{1},C_{2})}=\{(P_{\sigma},P_{\sigma}):\sigma\in S_{3}\}$. Thus $\operatorname{Sym}\mathscr{A}_{(C_{2},O)}\circ\operatorname{Sym}\mathscr{A}_{(C_{1},C_{2})}\circ\operatorname{Sym}\mathscr{A}_{(I,C_{1})}=\{(\operatorname{id},\operatorname{id})\}\lneq\operatorname{Sym}\mathscr{A}$.
\end{proof}

\begin{proposition}
\label{prop:group-closure}

Let $H\lneq G\leq\Gamma$, let $\alpha_X:=\alpha|_G$, and let $G\overset{\alpha_Y}{\curvearrowright}Y$ with $\alpha_Y(G)\leq K$, and suppose $\mathrm{A}=\underline{H}_{\alpha_X}^{\alpha_Y}\lneq\underline{G}_{\alpha_X}^{\alpha_Y}$. Let $\mathscr{A}_{(C_{1},C_{2})}$ be a subarchitecture on which $\mathcal{C}$ closes $H$ to $G$. Suppose the node maps $\phi_{v}^{\theta}$ of $\mathscr{A}_{(C_{1},C_{2})}$ are $H$-equivariant, and suppose that $a^{\theta}$ and $e^{\theta}$ are $G$-equivariant for all \(\theta\). Then $\underline{H}_{\alpha_{X}}^{\alpha_{Y}}\lneq \underline{G}_{\alpha_{X}}^{\alpha_{Y}} \leq \operatorname{Sym}\mathscr{A}$, and group-closure therefore causes symmetry slippage.
\end{proposition}

\begin{proof}
Since $\mathcal C$ closes $H$ to $G$ on $\mathscr A_{(C_1,C_2)}$ and every $\phi_v^\theta$ is $H$-equivariant, every $\phi_v^\theta$ is $G$-equivariant. Hence the $\Phi^\theta:Z_{C_1}\to Z_{C_2}$ realized by $\mathscr A_{(C_1,C_2)}$ is $G$-equivariant. For every $g\in G$,
\begin{align}
f^\theta\alpha_X(g)&=e^\theta\Phi^\theta a^\theta\alpha_X(g) \\
&=e^\theta\Phi^\theta\alpha_{C_1}(g)a^\theta \\
&=e^\theta\alpha_{C_2}(g)\Phi^\theta a^\theta \\
&=\alpha_Y(g)e^\theta\Phi^\theta a^\theta \\
&=\alpha_Y(g)f^\theta
\end{align}
whence every $f^\theta\in\mathcal F_{\mathscr A}$ is $G$-equivariant, so $\underline G_{\alpha_X}^{\alpha_Y}\leq\operatorname{Sym}\mathscr A$. Since by assumption $\mathrm A=\underline H_{\alpha_X}^{\alpha_Y}\lneq\underline G_{\alpha_X}^{\alpha_Y}$, we get $\mathrm A\lneq\underline G_{\alpha_X}^{\alpha_Y}\leq\operatorname{Sym}\mathscr A$, and therefore $\mathscr A$ exhibits symmetry slippage.
\end{proof}

\groupclosureanalyticprop*

\begin{proof}
Fix a node $v$ of $\mathscr A_{(C_1,C_2)}$ and an $H$-equivariant $\phi\in C^\omega(X)$. Since the spaces are $G$-representations, $g0=0$ for all $g\in G$. Taylor-expanding $\phi$ at $0$ gives on an open neighborhod of $0$, $\phi(x)=\sum_{m=0}^{\infty}P_m(x)$, where $P_m(x):=\frac{1}{m!}D^m\phi(0)[x,\dots,x]$ is a homogeneous polynomial map of degree $m$. Using $H$-equivariance and expanding both sides at $0$ then gives $\sum_{m=0}^{\infty}P_m(hx)=\sum_{m=0}^{\infty}hP_m(x)$. Matching terms, $P_m(hx)=hP_m(x)$ for every $m$, from which we see each $P_m$ is $H$-equivariant. Because polynomials close $H$ to $G$ on $\mathscr A_{(C_1,C_2)}$, each $P_m$ is therefore $G$-equivariant.

Now fix $g\in G$. For $x$ in a small enough neighborhood $U$ of $0$ that the Taylor expansions of both $\phi(x)$ and $\phi(gx)$ converge, $\phi(gx)=\sum_{m=0}^{\infty}P_m(gx)=\sum_{m=0}^{\infty}gP_m(x)=g\sum_{m=0}^{\infty}P_m(x)=g\phi(x)$. Thus the real-analytic map $x\mapsto\phi(gx)-g\phi(x)$ vanishes on $U$. But by analytic continuation it must then vanish everywhere. Hence $\phi$ is $G$-equivariant. Since $v$ and $\phi$ were arbitrary, $C^\omega(X)$ closes $H$ to $G$ on $\mathscr A_{(C_1,C_2)}$.
\end{proof}

\overquotientingprop*

\begin{proof}
Suppose $\mathscr{A}$ over-quotients $\mathrm{A}$, witnessed by $s\in J_X$. Then for every $\theta\in\Theta$ and $x\in X$, $a^\theta s(x)\sim a^\theta(x)$. Since $\Phi^\theta$ factors through $Z_{C_1}/{\sim}$, we have $\Phi^\theta a^\theta s(x)=\Phi^\theta a^\theta(x)$. Applying $e^\theta$ gives $f^\theta s=f^\theta$ for every $\theta$. Hence $(s,I_Y)\in\operatorname{Sym}\mathscr A$. But Definition~\ref{def:over-quotienting} asserts $(s,I_Y)\notin\mathrm A$. Therefore $(s,I_Y)\in\operatorname{Sym}\mathscr A-\mathrm A$, so $\operatorname{Sym}\mathscr A\not\subset\mathrm A$ and $\mathscr A$ exhibits symmetry slippage.
\end{proof}

\extensionmistypingprop*

\begin{proof}
Let $H \lneq G$, and suppose $\mathscr{A}$ extension-mistypes $\widetilde{\rho}^{\text{p}}$ on $\mathscr{A}_{(C_{1}, C_{2})}$ as witnessed by some extension $\widetilde{\rho}_{}^{\text{r}}$ of $\rho$. Suppose there exist actions $\alpha_{X}^{\text{r}}:G \curvearrowright X$, $\alpha_{Y}^{\text{r}}: G \curvearrowright Y$ such that $a^{\theta}$ is $G$-equivariant with respect to $\alpha_{X}^{\text{r}}$ and $\widetilde{\rho}_{C_{1}}^{\text{r}}$, and $e^{\theta}$ is $G$-equivariant with respect to $\widetilde{\rho}_{C_{2}}^{\text{r}}$ and $\alpha_{Y}^{\text{r}}$. By extension-mistyping, every node map $\phi_v^\theta$ of $\mathscr A_{(C_1,C_2)}$ is $G$-equivariant with respect to $\widetilde{\rho}^{\mathrm r}$, and hence the realized map $\Phi^\theta:Z_{C_1}\to Z_{C_2}$ is $G$-equivariant with respect to $\widetilde{\rho}_{C_1}^{\mathrm r}$ and $\widetilde{\rho}_{C_2}^{\mathrm r}$. Thus, for every $g\in G$ and $\theta\in\Theta$,
\begin{align}
f^\theta\alpha_X^{\mathrm r}(g) &=e^\theta\Phi^\theta a^\theta\alpha_X^{\mathrm r}(g) \\
&=e^\theta\Phi^\theta\widetilde{\rho}_{C_1}^{\mathrm r}(g)a^\theta \\
&=e^\theta\widetilde{\rho}_{C_2}^{\mathrm r}(g)\Phi^\theta a^\theta \\
&=\alpha_Y^{\mathrm r}(g)e^\theta\Phi^\theta a^\theta \\
&=\alpha_Y^{\mathrm r}(g)f^\theta
\end{align}
Hence every $f^\theta$ is $G$-equivariant with respect to $\alpha_X^{\mathrm r}$ and $\alpha_Y^{\mathrm r}$. Since $\alpha_X^{\mathrm r}(G)\leq J$ and $\alpha_Y^{\mathrm r}(G)\leq K$, it follows that $\underline{G}{\alpha_X^{\mathrm r}}^{\alpha_Y^{\mathrm r}}\leq\operatorname{Sym}\mathscr A$. So if $\underline{G}{\alpha_X^{\mathrm r}}^{\alpha_Y^{\mathrm r}}\not\subset\mathrm A$, then some element of $\underline{G}_{\alpha_X^{\mathrm r}}^{\alpha_Y^{\mathrm r}}$ lives in $\operatorname{Sym}\mathscr A-\mathrm A$, and so $\operatorname{Sym}\mathscr A\not\subset\mathrm A$ and $\mathscr A$ exhibits symmetry slippage.
\end{proof}


\begin{proposition}
\label{prop:extension-mistyping-special}
Suppose $H \triangleleft G$ with $[G:H]=2$; choose $s \in G-H$. Suppose the node maps \(\phi_{v}^\theta\) are \(H\)-equivariant with respect to \(\boldsymbol \rho\), and suppose there exist linear maps $\boldsymbol S=\big(S_{v} \in \operatorname{GL}(Z_{v}) \big)_{v \in V}$ satisfying (i) $S_{v} = \eta_{v}(s)$ for all $v \in T$, $i_{S_{v}} \circ \rho_{v}=\rho_{v} \circ i_{s}$\footnote{Here $i_{g}(x)$ denotes conjugation of $x$ by $g$.}, (ii) $S_{v}^{2}=\rho_{v}(s^{2})$, and (iii) $\phi_{v}^{\theta}S_{\{ u \to v \}}=S_{v}\phi_{v}^{\theta}$ for all $v, \theta$. If $\widetilde{\rho}_{v}^{\text{p}}(s)\neq S_{v}$ for some $v \in T$, then there exists an extension $\widetilde{\boldsymbol \rho}^{\text{r}}$ of $\boldsymbol \rho$ to $G$ such that $\mathscr{A}$ extension-mistypes $\widetilde{\boldsymbol \rho}^{\text{p}}$ on $\mathscr{A}_{(C_{1},C_{2})}$. 
\end{proposition}

\begin{proof}
Since $[G:H]=2$, $G=H\sqcup sH$, so every $g\in G$ equals either $h$ or $sh$ for some $h\in H$. For each $v$, define $\widetilde{\rho}_v^{\mathrm r}:G \to \operatorname{Aut} Z_{v}$ by $\widetilde{\rho}_v^{\mathrm r}(h):=\rho_v(h)$ and $\widetilde{\rho}_v^{\mathrm r}(sh):=S_v\rho_v(h)$. We claim $\widetilde{\rho}_{v}^{\text{r}}$ witnesses that $\mathscr{A}$ extension-mistypes $\widetilde{\rho}^{\text{p}}$. This requires showing (Step 1) that $\widetilde{\rho}_{v}^{\text{r}}:G \to \operatorname{Aut}Z_{v}$ defines action extending $\rho_v$, (Step 2) that $\widetilde{\rho}^{\text{r}}_{v}=\eta_{v}$ forall $v \in T$, (Step 3) that $\eta_{v} \neq \widetilde{\rho}_{v}^{\text{p}}$ for some $v \in T$, (Step 4) the $\phi_{v}^{\theta}$ are equivariant with respect to $\widetilde{\rho}^{\text{r}}_{v}$. 

\emph{Step 1.} We first claim $\widetilde{\rho}_{v}^{\text{r}}:G \to \operatorname{Aut}Z_{v}$ defines action extending $\rho_v$. This requires checking (a) that $\widetilde{\rho}_{v}^{\text{r}} |_{H}=\rho_{v}$ and (b) that $\widetilde{\rho}^{\text{r}}_{v}$ is a homomorphism. (a) holds by construction; for (b), it suffices to check that $\widetilde{\rho}_{v}^{\text{r}}$ respects products of the form $h_{1}h_{2}$, $sh_{1}h_{2}$, $h_{1}sh_{2}$, $sh_{1}sh_{2}$. The first is immediate from (a). The second is immediate, too, by the definition of $\widetilde{\rho}_{v}^{\text{r}}$ and the fact that $\rho_{v}$ is homomorphism. For the case $h_1sh_2$, note that since $H\triangleleft G$ it holds $s^{-1}h_1s\in H$, and (i) gives $\rho_v(h_1)S_v=S_v\rho_v(s^{-1}h_1s)$. Therefore
\[
\begin{aligned}
\widetilde{\rho}_{v}^{\mathrm r}(h_1)\widetilde{\rho}_{v}^{\mathrm r}(sh_2)
&=\rho_v(h_1)S_v\rho_v(h_2)\\
&=S_v\rho_v(s^{-1}h_1s)\rho_v(h_2)\\
&=S_v\rho_v(s^{-1}h_1sh_2)\\
&=\widetilde{\rho}_{v}^{\mathrm r}(h_1sh_2),
\end{aligned}
\]
where the last equality we used $h_1sh_2=s(s^{-1}h_1sh_2)$. Lastly, for the case $sh_1sh_2$, condition (i) gives $S_v\rho_v(h_1)S_v^{-1}=\rho_v(sh_1s^{-1})$, while condition (ii) gives $S_v^2=\rho_v(s^2)$. Hence
\[
\begin{aligned}
\widetilde{\rho}_{v}^{\mathrm r}(sh_1)\widetilde{\rho}_{v}^{\mathrm r}(sh_2)
&=S_v\rho_v(h_1)S_v\rho_v(h_2)\\
&=\rho_v(sh_1s^{-1})S_v^2\rho_v(h_2)\\
&=\rho_v(sh_1s^{-1})\rho_v(s^2)\rho_v(h_2)\\
&=\rho_v(sh_1sh_2)\\
&=\widetilde{\rho}_{v}^{\mathrm r}(sh_1sh_2).
\end{aligned}
\]
Here $sh_1sh_2\in H$: normality gives $sh_1s^{-1}\in H$, while $[G:H]=2$ implies $s^2\in H$. Thus $\widetilde{\rho}_v^{\mathrm r}$ indeed defines a $G$-action extending $\rho_v$.

\emph{Step 2.} We next claim that $\widetilde{\rho}_v^{\mathrm r}=\eta_v$ for every $v\in T$. Indeed, both actions restrict to $\rho_v$ on $H$, while condition (i) gives $\widetilde{\rho}_v^{\mathrm r}(s)=S_v=\eta_v(s)$. Since every element of $G$ is either $h$ or $sh$ for some $h\in H$, for $sh\in sH$ we have
\[
\widetilde{\rho}_v^{\mathrm r}(sh)
=S_v\rho_v(h)
=\eta_v(s)\eta_v(h)
=\eta_v(sh).
\]
Thus $\widetilde{\rho}_v^{\mathrm r}=\eta_v$ on all of $G$ for every $v\in T$. 

\emph{Step 3.} This is clear since by hypothesis $\widetilde{\rho}_v^{\mathrm p}(s)\neq S_v=\widetilde{\rho}_v^{\mathrm r}(s)=\eta_v(s)$ for some $v\in T$, so $\eta_v\neq\widetilde{\rho}_v^{\mathrm p}$ for that $v$.

\emph{Step 4.} It remains to check that every node map is $G$-equivariant with respect to $\widetilde{\boldsymbol\rho}^{\mathrm r}$. By assumption, the node maps are already $H$-equivariant with respect to $\boldsymbol\rho$, and $\widetilde{\boldsymbol\rho}^{\mathrm r}|_H=\boldsymbol\rho$. Thus it remains only to check elements of $sH$. For $sh\in sH$, condition (iii) and $H$-equivariance give
\[
\begin{aligned}
\phi_v^\theta\widetilde{\rho}_{\{u\to v\}}^{\mathrm r}(sh)
&=\phi_v^\theta S_{\{u\to v\}}\rho_{\{u\to v\}}(h)\\
&=S_v\phi_v^\theta\rho_{\{u\to v\}}(h)\\
&=S_v\rho_v(h)\phi_v^\theta\\
&=\widetilde{\rho}_v^{\mathrm r}(sh)\phi_v^\theta.
\end{aligned}
\]
Since $G=H\sqcup sH$, every node map is therefore $G$-equivariant with respect to $\widetilde{\boldsymbol\rho}^{\mathrm r}$. We have shown that $\widetilde{\boldsymbol\rho}^{\mathrm r}$ is an extension of $\boldsymbol\rho$, that $\widetilde{\rho}_v^{\mathrm r}=\eta_v$ for every $v\in T$, that $\eta_v\neq\widetilde{\rho}_v^{\mathrm p}$ for some $v\in T$, and that every node map is $G$-equivariant with respect to $\widetilde{\boldsymbol\rho}^{\mathrm r}$. 

The result follows.
\end{proof}

\subsection{Elaboration and Proofs from Section~\ref{sec:computing}}
\label{sec:computation-proofs}

This section more precisely states, then proves, the statements claimed in Section~\ref{sec:computing}.

\begin{theorem}
\label{thm:hardness-app}
For every finite group \(H\), computing additional generators \(B\) satisfying \(\operatorname{Sym} \mathscr{A}= \langle \mathrm{A}, B \rangle\) is GI-hard, even when \(\mathrm{A}=\underline{H}_{\alpha_{X}}^{\alpha_{Y}} \leq \operatorname{Sym} \mathscr{A}\) is promised.
\end{theorem}

To prove Theorem~\ref{thm:hardness-app} we first prove the following, which will also be useful for Proposition~\ref{prop:AMLP-one-layer-app} below. Let $I,O$ be finite index sets. Recall from Subexample~\ref{subexample:aut-closure-finite} that the architecture $\mathscr{A}$ corresponding to a parameter-sharing linear $\operatorname{MLP}$ layer $X:=\mathbb{R}^{I} \to \mathbb{R}^{O}=:Y$ may be encoded as a coloring $c$ of the bipartite graph $(O, I, O \times I)$, and conversely that any colored bipartite graph $\Sigma=(O, I, O \times I, c)$ determines a parameter-sharing linear layer $W:\mathbb{R}^{I} \to \mathbb{R}^{O}$ as $W=\sum_{\gamma} w_{\gamma}A_{\gamma}$, where $(A_{\gamma})_{oi}=\boldsymbol 1_{\{  c(o,i)=\gamma\}}$.  

\begin{proposition}
\label{prop:graphaut-app}
Let $\mathscr{A}$ denote the architecture corresponding to a linear $\operatorname{MLP}$ layer $\mathbb{R}^{I} \to \mathbb{R}^{O}$, sharing parameters according to colored graph $\Sigma$. Then $\operatorname{Aut}\Sigma=\operatorname{Sym}\mathscr{A} \leq S_{I} \times S_{O}$. The result continues to hold for a nonlinear layer $x \mapsto \mathrm{a}(W x)$ provided $(r \in \mathbb{R}) \mapsto (t   \mapsto \mathrm{a}(tr))$ is injective.
\end{proposition}

\begin{proof}
We first consider the linear case. For each $\gamma \in \operatorname{im}c$, let $A_\gamma$ denote its indicator matrix and define $W_{\boldsymbol {w}}:=\sum_\gamma w_\gamma A_\gamma$ where $w_{\gamma}$ are trainable weights. Hence \begin{align}(P,Q)\in\operatorname{Sym}\mathscr A &\iff W_{\boldsymbol w}P=QW_{\boldsymbol w}\text{ for all }\boldsymbol w \\ &\iff A_\gamma P=QA_\gamma\text{ for all } \gamma \\ &\iff (P,Q)\in\operatorname{Aut}(\Sigma),\end{align}whence $\operatorname{Sym}\mathscr A=\operatorname{Aut}(\Sigma)$.

Now consider the nonlinear case $x\mapsto \mathrm a(W_{\boldsymbol w}x)$. The inclusion $\operatorname{Aut}(\Sigma) \subset \operatorname{Sym}\mathscr A$ follows from the linear case, since coordinate permutations commute with $\mathrm a$. For the reverse inclusion, if $(P,Q)\in\operatorname{Sym}\mathscr A$, then for every $\boldsymbol w,x$, $\mathrm a(W_{\boldsymbol w}Px)=Q\mathrm a(W_{\boldsymbol w}x)=\mathrm a(QW_{\boldsymbol w}x)$. Applying the same identity to $t\boldsymbol w$ gives $\mathrm a(tW_{\boldsymbol w}Px)=\mathrm a(tQW_{\boldsymbol w}x)$ for all $t \in \mathbb{R}$. Since $r\mapsto(t\mapsto\mathrm a(tr))$ is injective, this implies coordinatewise that $W_{\boldsymbol w}Px=QW_{\boldsymbol w}x$. Thus $W_{\boldsymbol w}P=QW_{\boldsymbol w}$ for every $\boldsymbol w$, and the linear case gives $(P,Q)\in\operatorname{Aut}(\Sigma)$. Thus $\operatorname{Sym}\mathscr A=\operatorname{Aut}(\Sigma)$, as claimed.
\end{proof}

\begin{proof}
Let $H$ a finite group be given. Given any finite simple graph $\mathcal{G}=(\mathcal{V}, \mathcal{E})$, take $I:=\mathcal{V} \times H=:O$. $H$ acts on $O$ and $I$ in the evident way, $h \cdot (o, h_{o})=(o, h h_{o})$, $h \cdot (i, h_{i})=(i, h h_{i})$, inducing permutation representations $\rho_{X}:H \to S_{I}\leq\operatorname{GL}(\mathbb{R}^{I})$, $\rho_{Y} :H \to S_{O}\leq  \operatorname{GL}(\mathbb{R}^{O})$. Take the family of actions $\alpha_{X}:=\rho_{X}$, $\alpha_{Y}:=\rho_{Y}$. We construct in polynomial time an architecture $\mathscr{A}$ for which $\underline{H}_{\alpha_{X}}^{\alpha_{Y}}=:\mathrm{A} \leq \operatorname{Sym}\mathscr{A}$, and show this reduces Graph Isomorphism to the problem of computing additional generators $B$ satisfying $\langle \mathrm{A}, B \rangle=\operatorname{Sym}\mathscr{A}$.

So let $\mathcal{G}=(\mathcal{V},\mathcal{E})$ be a finite simple graph. Put $I:=\mathcal{V} \times H=:O$. Define $\kappa: \mathcal{V} \times \mathcal{V} \to \{ \text{``self"}, \text{``edge''}, \text{``nonedge"} \}$ by $$\kappa(v_{o},v_{i})=\begin{cases}
 \text{``self"} & o=i \\
 \text{``edge"} & \{  o,i  \} \in \mathcal{E} \\
 \text{``nonedge"} & \text{else.}
\end{cases}$$
Define \begin{align}
c : O  \times I &\xrightarrow{} \{ \text{``self", ``edge", ``nonedge"} \} \times H \\
\big((o, h_{o} ), (i, h_{i}) \big) & \mapsto  \underbrace{ \big( \kappa(o,i) , h_{o} ^{-1} h_{i}  \big) }_{ =:\big( c_{1}(o, h_{0}), c_{2}(i, h_{i})\big) }
\end{align}
and put $\Sigma:=(O, I, O \times I, c)$. Let $\mathscr{A}$ be the corresponding linear parameter-sharing $\operatorname{MLP}$ layer (cf. Subexample~\ref{subexample:aut-closure-finite}). 

 For all $h,h_{o}, h_{i} \in H$ and $o,i \in \mathcal{V}$, one has
\begin{align}
c\big((o,hh_o),(i,hh_i)\big)
&=\big(\kappa(o,i),(hh_o)^{-1}(hh_i)\big)\\
&=\big(\kappa(o,i),h_o^{-1}h_i\big)\\
&=c\big((o,h_o),(i,h_i)\big)
\end{align}
from which it follows that $\underline{H}_{\alpha_X}^{\alpha_Y}=\mathrm A\leq\operatorname{Sym}\mathscr A$. By Proposition~\ref{prop:graphaut-app}, then, $\operatorname{Sym}\mathscr A=\operatorname{Aut} \Sigma$. We claim that $\operatorname{Aut}\Sigma\cong\operatorname{Aut}(\mathcal G)\times H$, and thus $\operatorname{Sym}\mathscr{A} \cong\operatorname{Aut}(\mathcal{G}) \times H$. 

To this end let $(P,Q)\in\operatorname{Aut}\Sigma$, $P=(P_{1}: \mathcal{V}\times H \to \mathcal{V}, P_{2}:\mathcal{V} \times H \to H)$, $Q=(Q_{1}:\mathcal{V} \times H\to \mathcal{V}, Q_{2}:\mathcal{V} \times H \to H)$. Given $v\in\mathcal V$, observe that $c_{1}\big( (v , \cdot), (v, \_) \big) \equiv \text{``self"}$. So since $(P,Q)$ preserves colors, $c_{1}\big( Q(v,\cdot), P(v, \_) \big)=\text{``self"}$, and hence $Q_{1}(v, \cdot) \equiv P_{1}(v, \_)=:\pi(v)$ for some $\pi \in S_{\mathcal{V}}$. Write $p_{v}(b):=P_{2}(v,b)$ and $q_{v}(a):=Q_{2}(v, a)$, so that $P(v, b)=( \pi(v), p_{v}(b) )$, $Q(v,a)=(\pi(v), q_{v}(a))$. Let $g_{v}=p_{v}(e)$. We would like to define a prospective group isomorphism  \begin{align}
 \Phi:\operatorname{Aut}\Sigma  &\to \operatorname{Aut} (\mathcal{G}) \times H \\
 (P,Q) &\mapsto ( \pi, g_{v} ).
\end{align}
Checking this is well-defined requires (1) checking $\pi$ is an automorphism of $\mathcal{G}$ and (2) checking independence of $v \in V$. We then must check (3) $\Phi$ is a group homomorphism and (4) $\Phi$ is invertible. We treat these in turn.

\emph{Step 1.} To see that $\pi \in \operatorname{Aut}\mathcal{G}$, observe that since $(P,Q)$ preserves colors, $c_{1}\big(Q(o,a),P(i,b)\big)=c_{1}\big((o,a),(i,b)\big)$ for all $o,i\in\mathcal V$ and $a,b\in H$. By the definition of $\pi$, the left-hand side is $\kappa(\pi(o),\pi(i))$, while the right-hand side is $\kappa(o,i)$. Hence $\kappa(\pi(o),\pi(i))=\kappa(o,i)$ for all $o,i\in\mathcal V$. In particular, $\{o,i\}\in\mathcal E$ if and only if $\{\pi(o),\pi(i)\}\in\mathcal E$. Since $\pi\in S_{\mathcal V}$, it follows that $\pi\in\operatorname{Aut}\mathcal G$.

\emph{Step 2.} To check that $\Phi$ is independent of $v\in\mathcal V$, first use preservation of second color component on $((v,a),(v,b))$. This gives $q_v(a)^{-1}p_v(b)=a^{-1}b$ for all $a,b\in H$. Taking $a=b$ gives $p_v(a)=q_v(a)$, while taking $a=e$ gives $p_v(b)=q_v(e)b$. Since $g_v:=p_v(e)=q_v(e)$, we therefore have $p_v(b)=q_v(b)=g_vb$ for all $b\in H$. Thus $P(v,a)=Q(v,a)=(\pi(v),g_va)$. Now let $u,v\in\mathcal V$. Preservation of the second color component on $((u,a),(v,b))$ gives $(g_ua)^{-1}(g_vb)=a^{-1}b$. Taking $a=b=e$ yields $g_u=g_v$. Hence $g_v$ is independent of $v$.

\emph{Step 3.} To check that $\Phi$ is a group homomorphism, let $\Phi(P_j,Q_j)=(\pi_j,g_j)$ for $j=1,2$. By Step 2, $P_j(v,a)=Q_j(v,a)=(\pi_j(v),g_ja)$. Hence $P_1P_2(v,a)=P_1(\pi_2(v),g_2a)=(\pi_1\pi_2(v),g_1g_2a)$, and likewise $Q_1Q_2(v,a)=(\pi_1\pi_2(v),g_1g_2a)$. Therefore $\Phi\big((P_1,Q_1)(P_2,Q_2)\big)=(\pi_1\pi_2,g_1g_2)=\Phi(P_1,Q_1)\Phi(P_2,Q_2)$.

\emph{Step 4.} We claim the map $\Psi:\operatorname{Aut}(\mathcal G)\times H\to\operatorname{Aut}(\Sigma)$ given by $\Psi(\pi,g):=(\Lambda_{\pi,g},\Lambda_{\pi,g})$, where $\Lambda_{\pi,g}(v,a):=(\pi(v),ga)$, is an inverse to $\Phi$. First, $\Psi$ is well-defined: for any $(\pi,g)\in\operatorname{Aut}(\mathcal G)\times H$, we have \begin{align}
c\big(\Lambda_{\pi,g}(o,h_o),\Lambda_{\pi,g}(i,h_i)\big)&\big(\kappa(\pi(o),\pi(i)),(gh_o)^{-1}(gh_i)\big) \\
&=\big(\kappa(o,i),h_o^{-1}h_i\big) \\
&=c\big((o,h_o),(i,h_i)\big),
\end{align}where the second equality uses $\pi\in\operatorname{Aut}(\mathcal G)$. Thus $(\Lambda_{\pi,g},\Lambda_{\pi,g})\in\operatorname{Aut}(\Sigma)$. By construction, $\Phi(\Psi(\pi,g))=(\pi,g)$ for every $(\pi,g)\in\operatorname{Aut}(\mathcal G)\times H$. Conversely, if $(P,Q)\in\operatorname{Aut}\Sigma$ and $\Phi(P,Q)=(\pi,g)$, then Step 2 gives $P(v,a)=Q(v,a)=(\pi(v),ga)=\Lambda_{\pi,g}(v,a)$ for all $v\in\mathcal V$ and $a\in H$, so $\Psi(\Phi(P,Q))=(P,Q)$. Hence $\Psi=\Phi^{-1}$, and therefore $\operatorname{Aut}\Sigma\cong\operatorname{Aut}(\mathcal G)\times H$.

We therefore see $\operatorname{Sym}\mathscr A\cong\operatorname{Aut}(\mathcal G)\times H$. Now let $\mathcal G_1=(\mathcal V_1,\mathcal E_1)$ and $\mathcal G_2=(\mathcal V_2,\mathcal E_2)$ be connected finite simple graphs, and put $\mathcal K:=\mathcal G_1\sqcup\mathcal G_2$. Apply the preceding construction to $\mathcal K$, producing an architecture $\mathscr A_{\mathcal K}$, and suppose additional generators $B=\{(P_\ell,Q_\ell)\}_\ell$ satisfying $\langle\mathrm A,B\rangle=\operatorname{Sym}\mathscr A_{\mathcal K}$ are computed. Together with the known generators of $\mathrm A$, the permutations $P_\ell$ generate the full input action of $\operatorname{Sym}\mathscr A_{\mathcal K}$.

Fix $v\in\mathcal V_1$, and let $e\in H$ denote the identity. Let us compute the orbit of $(v,e)$ under this generated input action. Under the isomorphism $\Phi:\operatorname{Sym}\mathscr A_{\mathcal K}\to\operatorname{Aut}(\mathcal K)\times H$, every input permutation has the form $P(w,a)=(\pi(w),ga)$ for some $(\pi,g)\in\operatorname{Aut}(\mathcal K)\times H$. Hence the orbit of $(v,e)$ intersects $\mathcal V_2\times H$ if and only if there exists $\pi\in\operatorname{Aut}(\mathcal K)$ with $\pi(v)\in\mathcal V_2$. Since $\mathcal G_1$ and $\mathcal G_2$ are the connected components of $\mathcal K$, this occurs if and only if $\mathcal G_1\cong\mathcal G_2$.

The orbit of a point under a permutation group given by generators can be computed in polynomial time~\citep{Seress2003}. The construction above is also polynomial-time. Therefore Graph Isomorphism reduces in polynomial time to computing additional generators $B$ satisfying $\langle\mathrm A,B\rangle=\operatorname{Sym}\mathscr A$, so this problem is GI-hard.
\end{proof}

Hardness makes us expect that deducing whether $\mathrm{A}=\operatorname{Sym}\mathscr{A}$ can necessitate the use of architecture-specific knowledge. Since general architectures are usually built from modular pieces, it is natural to ask when $\operatorname{Sym}\mathscr{A}$ can be exactly computed from subarchitectures $\operatorname{Sym} \mathscr{A}_{(C_{i}, C_{j})}$, that is, when a converse to Proposition~\ref{prop:decomposing-tlaws} holds. The following theorem offers such conditions. Let $\mathscr{A}=\mathscr{A}_{(C_{\ell-1}, O)} \circ \dots \circ \mathscr{A}_{(I, C_{1})}$ be a decomposition of $\mathscr{A}$ for which $\mathcal{F}_{\mathscr{A}}=\mathcal{F}_{\mathscr{A}(C_{\ell-1}, C_{O})} \circ \dots \circ \mathcal{F}_{\mathscr{A}_{(I, C_{1})}}$. Assume $\boldsymbol\beta=(\beta_v:G\curvearrowright Z_v)$ is an action on the spaces of $\mathscr A$ such that
$\operatorname{Sym}\mathscr A\subset\beta_I(G)\times\beta_O(G)$.
For the following result, \(\operatorname{Sym}\) on the space \(Z_{C}\) is taken with respect to \(J=\beta_{C}(G)\).

\begin{theorem}
\label{thm:composition-converse-app}
For each $\ell$ suppose (i) there exists $\mathcal{F}_{\ell}' \subset \mathcal{F}_{\mathscr{A}_{(C_{\ell-1}, C_{\ell})}} \cap \operatorname{Map}_{G}(Z_{C_{\ell-1}}, Z_{C_{\ell}})$ such that $\mathcal{F}^{'\top}_{\ell}$ is injective, and (ii) for $\ell>1$, $fg \equiv gf$ for all $g \in G$ whenever $fgp=gfp$ for all $f \in \mathcal{F}_{\mathscr{A}_{(C_{\ell-1}, C_{\ell})}}, p \in \mathcal{F}'_{\ell - 1} \circ \dots \circ \mathcal{F}'_{1}$. Then  $\operatorname{Sym}_{}\mathscr{A}=\operatorname{Sym}_{}\mathscr{A}_{(C_{L-1}, O)} \circ \dots \circ \operatorname{Sym}_{}\mathscr{A}_{(I, C_{1})}$.
\end{theorem}

If the $fg-gf$ were linear, the first condition is equivalent to $\bigcap_{f \in \mathcal{F}'_{\ell}}^{}\operatorname{ker } f=(0)$ and the second is $\operatorname{span}(\bigcup_{p \in \mathcal{F}'_{\ell-1} \circ \dots \circ \mathcal{F}_{1}'}\operatorname{im }p)=Z_{C_{\ell -1}}$. Of course, linearity basically never holds, and the stated hypotheses are the generalizations of these natural notions to the nonlinear setting. 

\begin{proof}
We note that conditions (i) and (ii) are complementary, with (i) being a monic condition and (ii) being an epic condition. 

The inclusion $\supset$ is nearly immediate and is Proposition~\ref{prop:decomposing-tlaws}. For $\subset$, let $(\sigma, \tau) \in \operatorname{Sym}\mathscr{A}$. Explicitly,

$\begin{aligned} &\operatorname{Sym}\mathscr{A}_{(C_{L-1}, O)} \circ \dots \circ \operatorname{Sym}\mathscr{A}_{(I, C_{1})} \\ &= \left\{ (\sigma,\tau)\in J \times K: \begin{array}{l} \text{there exist }\delta_0=\sigma,\delta_1,\dots,\delta_{L-1},\delta_L=\tau\\ \text{such that }(\delta_{\ell-1},\delta_\ell)\in \operatorname{Sym}\mathscr A_{(C_{\ell-1},C_\ell)} \text{ for all }\ell \end{array} \right\}. \end{aligned}$

So the goal is to find $g_{0},\dots, g_{L} \in G$ such that $\beta_{I}(g_{0})=\sigma$, $\beta_{O}(g_{L})=\tau$, and, for every $\ell$, $f_{\ell}\beta_{C_{\ell-1}}(g_{\ell-1})=\beta_{C_{\ell}}(g_{\ell})f_{\ell} \text{ for all } f_{\ell} \in \mathcal{F}_{\mathscr{A}_{(C_{\ell-1},C_{\ell})}}$.

We first claim that WLOG $g_{L}=g_{0}=:g$. Indeed, if $r=f_{L}' \circ \dots \circ f_{1}^{'}$, $f_{\ell}' \in \mathcal{F}_{\ell}'$, then by $(\sigma, \tau) \in \operatorname{Sym}\mathscr{A}$ one has $\beta_{O}(g_{L})r=r \beta_{I}(g_{0})$ and by $G$-equivariance of $f_{\ell}' \in \mathcal{F}_{\ell}'$ one further has $\beta_{O}(g_{L})r=r \beta_{I}(g_{L})$. Injectivity (monicness) of the $\mathcal{F}'_{\ell}{}^{\top}$ then lets us "cancel $r$" from $r\beta_{I}(g_{0})=r\beta_{I}(g_{L})$ to conclude $\beta_{I}(g_0)=\beta_{I}(g_{L})$. From here we split into cases based on $\ell$.

\textit{Case $\ell=1$.} Suppose $\ell=1$. Let $f_{1}\in\mathcal{F}_{\mathscr{A}_{(I,C_{1})}}$ and $q=f_{L}'\circ\dots\circ f_{2}'$, where $f_{\ell}'\in\mathcal{F}_{\ell}'$. Since $qf_{1}\in\mathcal{F}_{\mathscr{A}}$, we have $qf_{1}\beta_{I}(g)=\beta_{O}(g)qf_{1}=q\beta_{C_{1}}(g)f_{1}$, where the second equality follows from $G$-equivariance of $q$. By successive injectivity of the $\mathcal{F}_{\ell}'^{\top}$, we may cancel $q$ to obtain $f_{1}\beta_{I}(g)=\beta_{C_{1}}(g)f_{1}$ for all $f_{1}\in\mathcal{F}_{\mathscr{A}_{(I,C_{1})}}$. Hence $(\beta_{I}(g),\beta_{C_{1}}(g))\in\operatorname{Sym}\mathscr{A}_{(I,C_{1})}$.

\textit{Case $\ell > 1$.} Now let $\ell>1$. Fix $f_{\ell}\in\mathcal{F}_{\mathscr{A}_{(C_{\ell-1},C_{\ell})}}$ and $p=f_{\ell-1}'\circ\dots\circ f_{1}'$, where $f_{j}'\in\mathcal{F}_{j}'$, and let $q=f_{L}'\circ\dots\circ f_{\ell+1}'$, where $f_{j}'\in\mathcal{F}_{j}'$. Since $p$ and $q$ are $G$-equivariant and $qf_{\ell}p\in\mathcal{F}_{\mathscr{A}}$, we have $qf_{\ell}\beta_{C_{\ell-1}}(g)p=qf_{\ell}p\beta_{I}(g)=\beta_{O}(g)qf_{\ell}p=q\beta_{C_{\ell}}(g)f_{\ell}p$. By successive injectivity of the $\mathcal{F}_{j}'^{\top}$ for $j>\ell$, we may cancel $q$ to obtain $f_{\ell}\beta_{C_{\ell-1}}(g)p=\beta_{C_{\ell}}(g)f_{\ell}p$ for all $p\in\mathcal{F}_{\ell-1}'\circ\dots\circ\mathcal{F}_{1}'$. For $\ell=L$, the same equality follows directly, since there is no $q$. By condition (ii), $f_{\ell}\beta_{C_{\ell-1}}(g)=\beta_{C_{\ell}}(g)f_{\ell}$ for all $f_{\ell}\in\mathcal{F}_{\mathscr{A}_{(C_{\ell-1},C_{\ell})}}$. Hence $(\beta_{C_{\ell-1}}(g),\beta_{C_{\ell}}(g))\in\operatorname{Sym}\mathscr{A}_{(C_{\ell-1},C_{\ell})}$ for every $\ell>1$.

Taking $\delta_{\ell}:=\beta_{C_{\ell}}(g)$, we have $\delta_{0}=\sigma$, $\delta_{L}=\tau$, and $(\delta_{\ell-1},\delta_{\ell})\in\operatorname{Sym}\mathscr{A}_{(C_{\ell-1},C_{\ell})}$ for every $\ell$. Hence $(\sigma,\tau)$ belongs to the relational composition, proving $\subset$.
\end{proof}

To exemplify Theorem~\ref{thm:composition-converse-app}, we apply it in the classical finite setting described in Subexample~\ref{subexample:aut-closure-finite}. In particular, we use it to compute $\operatorname{Sym}_{}(A\operatorname{-MLP})$ in closed-form, where $A$-$\operatorname{MLP}$ is an $\operatorname{MLP}$ each of whose layers is equivariant to a free action by a finite group $H$, with parameter-sharing determined by $A \subset H$ (cf. Subexample~\ref{subexample:aut-closure-finite}). Before applying Theorem~\ref{thm:composition-converse-app}, we first compute $\operatorname{Sym}_{}\mathscr{A}$ in closed form for $\mathscr{A}$ a single equivariant layer. This is achieved by the following result.

\begin{proposition}
\label{prop:AMLP-one-layer-app}
Let $\mathscr{A}:\mathbb{R}^{I} \to \mathbb{R}^{O}$ be a linear $A$-$\operatorname{MLP}$ layer as in Section~\ref{sec:local-symmetry-slippage-creation-mechanisms}, $A\neq \emptyset$. Put $D_{A}:=\langle A A ^{-1} \rangle$, $k:=[H :D_{A}]$. Then $$\operatorname{Sym} \mathscr{A}\cong  D_{A} \wr S_{k}.$$
The conclusion continues to hold for a nonlinear layer $x \mapsto \mathrm{a}(W x)$ provided $(r \in \mathbb{R}) \mapsto (t   \mapsto \mathrm{a}(tr))$ is injective. 
\end{proposition}

Here, the notation $\wr$ denotes the \emph{wreath product} $D_{A} \wr S_{k}:=D_{A}^{k} \rtimes S_{k}$ where $S_{k}$ acts on $D_{A}^{k}$ by permuting its factors. In the dense special case $A=H$, $D_{A}=H$ and $k=1$ and so $\operatorname{Sym}\mathscr{A} \cong H$.  

\begin{remark}
Proposition~\ref{prop:AMLP-one-layer-app} may be viewed as generalization and strengthening of Theorem 3.3 in \citep{ravanbakhsh2017equivarianceparametersharing}, which amounts to proving one direction, \(\mathrm{A} \leq \operatorname{Aut}(\Sigma_{A})\), in the special case where \(A\) is a symmetric generating set. (The authors claim their proof shows equality, but their argument of the reverse direction appears to be flawed.) In our case, this direction already follows abstractly via Proposition~\ref{prop:graphaut-app}; the content the proof below lies in the reverse inclusion. 
\end{remark}

\begin{proof}
Assume first the linear case. By Proposition~\ref{prop:graphaut-app}, $\operatorname{Sym}\mathscr A=\operatorname{Aut}(\Sigma_A)$; it remains to compute the latter. We first claim $\operatorname{Aut}(\Sigma_{A})$ can be identified with the pairs $(s,t)$ of permutations of $H$ that satisfy $sg a=tga$ for all $g \in H$, $a \in A$. This object we can then identify with $D_{A} \wr S_{k}$.  

To this end choose representatives $n_p$ of the input $H$-orbits and $m_q$ of the output $H$-orbits. Since the $H$-actions are free, every input and output index can be written uniquely as $gn_p$ or $gm_q$. Recall that the sharing classes are $\Delta_{p,q,a}=\{(gm_q,gan_p):g\in H\}$. Let $(\sigma_I,\sigma_O)\in\operatorname{Aut}(\Sigma_A)$. Since every triple $(p,q,a)$ has its own color, $\sigma_I$ and $\sigma_O$ must preserve each input and output $H$-orbit individually, implying there are bijections $s_p,t_q:H\to H$ such that $\sigma_I(gn_p)=s_p(g)n_p$ and $\sigma_O(gm_q)=t_q(g)m_q$. Preservation of $\Delta_{p,q,a}$ means that $s_p(ga)=t_q(g)a$ for all $g\in H$, $a\in A$, $p$, and $q$. Fix $a_0\in A$. Then $s_p(ga_0)=t_q(g)a_0$ for all $p,q$, so the left-hand side is independent of $p$ and the right-hand side is independent of $q$. Since $g\mapsto ga_0$ is a bijection of $H$, all $s_p$ are in fact the same bijection $s$, and all $t_q$ are in fact the same bijection $t$. Thus $s(ga)=t(g)a$ for all $g\in H$ and $a\in A$. 

Now let $a,b\in A$. We have $s(ga)=t(g)a$, and and similarly $gab^{-1}$ gives $s(ga)=t(gab^{-1})b$. Therefore $t(gab^{-1})=t(g)ab^{-1}$. Since $D_A=\langle AA^{-1}\rangle$, it follows that $t(gd)=t(g)d$ for all $g\in H$ and $d\in D_A$. Conversely, suppose $t:H\to H$ is a bijection satisfying $t(gd)=t(g)d$ for all $d\in D_A$. Fix $a_0\in A$ and define $s(x):=t(xa_0^{-1})a_0$. Then for every $a\in A$, $s(ga)=t(gaa_0^{-1})a_0=t(g)aa_0^{-1}a_0=t(g)a$, since $aa_0^{-1}\in D_A$. Hence $(s,t)$ preserves every sharing class. Thus $\operatorname{Aut}(\Sigma_A)$ is exactly the group of bijections $t:H\to H$ satisfying $t(gd)=t(g)d$ for all $d\in D_A$.

The $D_A$-orbits under right multiplication are the $k=[H:D_A]$ right cosets of $D_A$. Thus $t$ induces an arbitrary permutation $\pi\in S_k$ of these cosets. Choose representatives $r_1,\dots,r_k$. For each $i$, there is a unique $d_i\in D_A$ such that $t(r_i)=r_{\pi(i)}d_i$. Then $D_A$-equivariance forces $t(r_id)=r_{\pi(i)}d_id$ for all $d\in D_A$. Hence $t$ is determined by an arbitrary permutation $\pi\in S_k$ together with independent choices $d_1,\dots,d_k\in D_A$. Therefore $\operatorname{Aut}(\Sigma_A)\cong D_A^k\rtimes S_k=D_A\wr S_k$.

For the nonlinear layer, Proposition~\ref{prop:graphaut-app} still gives $\operatorname{Sym}\mathscr A=\operatorname{Aut}(\Sigma_A)$ under the injectivity condition on $\mathrm a$, so the same computation applies.
\end{proof}

We note the injectivity condition on $\mathrm{a}$ in Propositions~\ref{prop:graphaut-app} and \ref{prop:AMLP-one-layer-app} is not stringent; it is satisfied e.g. by $\operatorname{ReLU}$, sigmoid, $\operatorname{\tanh}$, leaky $\operatorname{ReLU}$, $\operatorname{ELU}$, etc.

\begin{proposition}
\label{AMLP-multilayer-app}
Let $\mathscr{A}=\mathscr{A}_{(C_{L-1}, O)}\circ \dots \circ \mathscr{A}_{(I, C_{1})}$ be an $A$-$\operatorname{MLP}$ as in Section~\ref{sec:local-symmetry-slippage-creation-mechanisms} with activation $\mathrm{a}$, such that $(r \in \mathbb{R}) \mapsto (t   \mapsto \mathrm{a}(tr))$ is injective. Put $D_{A}:=\langle A A ^{-1}\rangle$, $k:=[H :D_{A}]$. Then $$\operatorname{Sym}\mathscr{A}\cong D_{A} \wr S_{k}.$$
\end{proposition}

\begin{proof}
Write $Z_{C_{\ell}} \cong (\mathbb{R}^{H})^{m_{\ell}}$ for the space of the $\ell$th layer of $\mathscr{A}$. Let $G:=S_{H}$ act on $H$ in the obvious way, inducing a permutation representation $\gamma:G \to \operatorname{GL}(\mathbb{R}^{H})$ and in turn product actions $(\gamma_{\ell}: G \curvearrowright Z_{C_{\ell}})$. Fix $a_{*} \in A$ and let $P_{a_{*}}:\mathbb{R}^{H} \to \mathbb{R}^{H}$ be the permutation matrix induced by right-multiplication by $a_{*}$. Define representations $\rho_{\ell}:G\to\operatorname{GL}(\mathbb R^H)$ recursively by $\rho_{0}:=\gamma$ and $\rho_{\ell}(g):=P_{a_*}\rho_{\ell-1}(g)P_{a_*}^{-1}$, inducing in turn product actions $\boldsymbol\beta=(\beta_{\ell}:G\curvearrowright Z_{C_{\ell}})_{\ell=0}^{L}$. 

We claim $\mathscr{A}$ and $\boldsymbol \beta$ satisfies the hypotheses of Proposition~\ref{thm:composition-converse-app}; combined with Proposition~\ref{prop:AMLP-one-layer-app} this proves the result. Showing the hypotheses amounts to checking (a) $\operatorname{Sym}\mathscr{A}\leq \beta_{I}(G)\times \beta_{O}(G)$ and that conditions (i) and (ii) hold. The checks are straightforward, though writing them out explicitly takes some bookkeeping. 

\emph{Verification of (a).} Let $(\sigma,\tau)\in\operatorname{Sym}\mathscr A$. For $p\in[m_0]$ and $q\in[m_L]$, choose indices $p=p_0,p_1,\dots,p_L=q$ and consider the realizations $F_{p,q,\boldsymbol t}=f_{L,t_L}\circ\dots\circ f_{1,t_1}$, where $f_{\ell,t_\ell}(x)=\mathrm a(t_\ell(E_{p_\ell p_{\ell-1}}\otimes P_{a_*})x)$. Each $F_{p,q,\boldsymbol t}$ depends only on the $p$th input factor, and among the output factors only the $q$th is nonconstant. Since $F_{p,q,\boldsymbol t}\sigma=\tau F_{p,q,\boldsymbol t}$ for every $p,q,\boldsymbol t$, varying one input factor at a time shows that $\sigma$ preserves each input factor, and varying the $q$th output shows that $\tau$ preserves each output factor. Hence $\sigma=\operatorname{diag}(\pi_1,\dots,\pi_{m_0})$ and $\tau=\operatorname{diag}(\eta_1,\dots,\eta_{m_L})$ for some $\pi_p,\eta_q\in S_H$. On the $q$th output factor, $F_{p,q,\boldsymbol t}$ has the form $\psi_{\boldsymbol t}^{\times H}R\operatorname{pr}_p$, where $R:=P_{a_*}^L$ and $\psi_{\boldsymbol t}:\mathbb R\to\mathbb R$ is the scalar map obtained by composing the scaled activations. Restricting $F_{p,q,\boldsymbol t}\sigma=\tau F_{p,q,\boldsymbol t}$ to this factor gives $\psi_{\boldsymbol t}^{\times H}R\pi_p=\eta_q\psi_{\boldsymbol t}^{\times H}R=\psi_{\boldsymbol t}^{\times H}\eta_qR$. As $\boldsymbol t$ varies, the family $(\psi_{\boldsymbol t})_{\boldsymbol t}$ separates points of $\mathbb R$: if $r\neq s$, at each successive layer the injectivity of $u\mapsto(t\mapsto\mathrm a(tu))$ allows a choice of $t_\ell$ for which the two resulting values remain distinct. We may therefore cancel the family $(\psi_{\boldsymbol t}^{\times H})_{\boldsymbol t}$ to obtain $R\pi_p=\eta_qR$ for every $p,q$. Thus all $\pi_p$ are equal to some $g\in S_H$, and all $\eta_q$ are equal to $RgR^{-1}$. By construction of $\boldsymbol\beta$, $\sigma=\beta_I(g)$ and $\tau=\beta_O(g)$, proving $\operatorname{Sym}\mathscr A\leq\beta_I(G)\times\beta_O(G)$.

\emph{Verification of condition (i).} For each $\ell$, let $\mathcal F'_\ell=\{x\mapsto\mathrm a(t(E_{qp}\otimes P_{a_*})x):t\in\mathbb R,\ p\in[m_{\ell-1}],\ q\in[m_\ell]\}$. Since $\rho_\ell(g)P_{a_*}=P_{a_*}\rho_{\ell-1}(g)$, we have $\beta_\ell(g)(E_{qp}\otimes P_{a_*})=(E_{qp}\otimes P_{a_*})\beta_{\ell-1}(g)$, and therefore every element of $\mathcal F'_\ell$ is $G$-equivariant. Now let $x\neq y\in Z_{C_{\ell-1}}$. Then $x_{p,h}\neq y_{p,h}$ for some $p\in[m_{\ell-1}]$ and $h\in H$. For any $q\in[m_\ell]$, the block $E_{qp}\otimes P_{a_*}$ sends these two values to the same output coordinate. Writing them as $r\neq s$, injectivity of $r\mapsto(t\mapsto\mathrm a(tr))$ gives $t\in\mathbb R$ such that $\mathrm a(tr)\neq\mathrm a(ts)$. Hence some $f\in\mathcal F'_\ell$ satisfies $f(x)\neq f(y)$, so $(\mathcal F'_\ell)^\top$ is injective.

\emph{Verification of condition (ii).} Let $\ell>1$ and suppose $fgp=gfp$ for every $f\in\mathcal F_{\mathscr A_{(C_{\ell-1},C_\ell)}}$ and $p\in\mathcal F'_{\ell-1}\circ\dots\circ\mathcal F'_1$. Write $f(x)=\mathrm a(Bx)$. Since $tB$ is again a weight matrix for every $t\in\mathbb R$, the hypothesis gives $\mathrm a(tB\beta_{\ell-1}(g)p(x))=\mathrm a(t\beta_\ell(g)Bp(x))$ for every $t,p,x$. By injectivity of $r\mapsto(t\mapsto\mathrm a(tr))$, $(B\beta_{\ell-1}(g)-\beta_\ell(g)B)p(x)=0$ for every $p$ and $x$. Moreover, the differences $p(x)-p(y)$ span $Z_{C_{\ell-1}}$. Indeed, fix a standard basis vector $e_{u,h}$. Choose indices $u_0,\dots,u_{\ell-1}=u$ and $h_0,\dots,h_{\ell-1}=h$ so that $E_{u_j u_{j-1}}\otimes P_{a_*}$ sends the coordinate $(u_{j-1},h_{j-1})$ to $(u_j,h_j)$. Choose $x,y\in Z_I$ differing only in coordinate $(u_0,h_0)$. Inductively, if the two states after layer $j-1$ differ only in coordinate $(u_{j-1},h_{j-1})$, with distinct values $r\neq s$, choose $t_j$ such that $\mathrm a(t_jr)\neq\mathrm a(t_js)$. Then after layer $j$ they differ only in coordinate $(u_j,h_j)$. Thus, for some $p\in\mathcal F'_{\ell-1}\circ\dots\circ\mathcal F'_1$, we have $p(x)-p(y)=c e_{u,h}$ with $c\neq0$. Since $e_{u,h}$ was arbitrary, these differences span $Z_{C_{\ell-1}}$. The resulting difference a nonzero multiple of $e_{u,h}$. Since $B\beta_{\ell-1}(g)-\beta_\ell(g)B$ is linear and vanishes on every $p(x)$, it vanishes on these differences and hence on all of $Z_{C_{\ell-1}}$. Therefore $B\beta_{\ell-1}(g)=\beta_\ell(g)B$, and since $\mathrm a$ commutes with coordinate permutations, $f\beta_{\ell-1}(g)=\beta_\ell(g)f$. Thus condition (ii) holds.
\end{proof}

\subsection{Elaboration and Proofs from Section~\ref{sec:GEs}}
\label{sec:GE-proofs}

We briefly elaborate upon the exposition given in Section~\ref{sec:GEs}, parsing apart the individual pieces of Proposition~\ref{prop:main-GE-prop} from the main text to give their precise assumptions and provide some commentary. We taken together, the results below prove Proposition~\ref{prop:main-GE-prop}.

\paragraph{GEs for Group-Closure.} Recall that group-closure creates symmetry slippage as over-equivariance $\underline{H}_{\alpha_{X}}^{\alpha_{Y}} \lneq \underline{G}_{\alpha_{X}}^{\alpha _{Y}} \leq\operatorname{Sym}\mathscr{A}$, so that slipped transformations $s \in \operatorname{Sym}\mathscr{A}-\mathrm{A}$ take the form $s =\big(\alpha_{X}(g), \alpha_{Y}(g) \big)$ for $g \in G-H$. Roughly, the next result states that $c:X \to Z$ is a GE for $s$ provided $s \not \in \operatorname{Sym}c$.

\begin{proposition}
\label{prop:GE-GC-app}
Let $S \subset \operatorname{Sym}\mathscr{A}-\mathrm{A}$ be a collection of transformations slipped due to group-closure. Let $c:X \to Z$ and $G$ act on $Z$. Write $\mathscr{A}':=\mathscr{A}\oplus_{C_{2}}c$. If $\mathscr{A}_{(C_{2}, O)}^{\prime\top}$ is injective and $G$-equivariant and $c$ is such that $c\alpha_X(g)\neq gc$ for every $s_g=(\alpha_X(g),\alpha_Y(g))\in S$, then $c$ is a GE for $S$. 
\end{proposition}

\begin{proof}
Fix $s_g=\bigl(\alpha_X(g),\alpha_Y(g)\bigr)\in S$. Since $c\alpha_X(g)\neq gc$, there exists $x\in X$ such that $c(\alpha_X(g)x)\neq gc(x)$. Since $b$ is $G$-equivariant, $b(\alpha_X(g)x)=gb(x)$, and hence
$$
\bigl(b(\alpha_X(g)x),c(\alpha_X(g)x)\bigr)\neq
g\bigl(b(x),c(x)\bigr).
$$
By injectivity of $\mathscr{A}_{(C_2,O)}^{\prime\top}$, there therefore exists $e'\in\mathcal F_{\mathscr A'_{(C_2,O)}}$ such that
$$
e'\bigl(b(\alpha_X(g)x),c(\alpha_X(g)x)\bigr)\neq e'\bigl(g(b(x),c(x))\bigr).
$$
Since $\mathscr{A}_{(C_2,O)}^{\prime\top}$ is $G$-equivariant, the right-hand side equals $\alpha_Y(g)e'\bigl(b(x),c(x)\bigr)$. Thus, $f':=e'\circ(b,c)\in\mathcal F_{\mathscr A'}$,
$$
f'\alpha_X(g)(x)\neq\alpha_Y(g)f'(x).
$$
Hence $s_g\notin\operatorname{Sym}\mathscr A'$. Since $s_g\in S$ was arbitrary, $\mathscr A'$ grips every $s\in S$, and therefore $c$ is a GE for $S$.
\end{proof}

\paragraph{GEs for Over-Quotienting.} Recall that slipped transformations $s \in \operatorname{Sym}\mathscr{A}-\mathrm{A}$ created by over-quotienting take the form $s=(\sigma, I_{Y})$ for some $\sigma \in J_X$. Roughly, the following result states that $c:X\to Z$ is a GE for $s$ provided $c\sigma \neq c$.

\begin{proposition}
\label{prop:GE-OQ-app}
Let $S \subset \operatorname{Sym}\mathscr{A}-\mathrm{A}$ be a collection of transformations slipped due to over-quotienting, let $c:X \to Z$, and write $\mathscr{A}':=\mathscr{A}\oplus_{C_{2}}c$. If $\mathscr{A}_{(C_{2},O)}^{\prime\top}$ is injective\footnote{More generally, it suffices to have injectivity on the pairs of representations witnessing $c\sigma \neq c$.\label{footnote:injectivity-weakening}} and $c$ is such that $c\sigma \neq c$ for all $s=(\sigma, I_{Y}) \in S$, then $c$ is a GE for $S$.
\end{proposition}

\begin{proof}
Fix $s=(\sigma,I_Y)\in S$. Since $c\sigma\neq c$, there exists $x\in X$ such that $c\sigma(x)\neq c(x)$. By over-quotienting, $a\sigma(x)\sim a(x)$, and since $\Phi$ is constant on $\sim$-classes, $b\sigma(x)=\Phi a\sigma(x)=\Phi a(x)=b(x)$. Hence
$\bigl(b\sigma(x),c\sigma(x)\bigr)\neq\bigl(b(x),c(x)\bigr)$. By injectivity of $\mathscr{A}_{(C_2,O)}^{\prime\top}$, there exists $e'\in\mathcal F_{\mathscr A'_{(C_2,O)}}$ separating these two representations. Thus, for $f':=e'\circ(b,c)\in\mathcal F_{\mathscr A'}$, $f'\sigma(x)\neq f'(x)$, so $f'\sigma\neq f'$. Therefore $(\sigma,I_Y)\notin\operatorname{Sym}\mathscr A'$. Since $s\in S$ was arbitrary, $\mathscr A'$ grips every $s\in S$, and hence $c$ is a GE for $S$.
\end{proof}

\paragraph{GEs for Extension-Mistyping.} Let $S \subset \operatorname{Sym}\mathscr{A}-\mathrm{A}$ be transformations slipped due to extension-mistyping; recall these take the form $s_{g}=\big( \alpha_{X}^{\text{r}}(g), \alpha_{Y}^{\text{r}}(g) \big) \in \operatorname{Sym} \mathscr{A}-\mathrm{A}$ for $g \in G$. Consider as before a modification $\mathscr{A}'=\mathscr{A}\oplus_{C}c$ of $\mathscr{A}$ for a cut $C$ and some $c:X \to Z$, with realized functions $f^{'}:X \xrightarrow{(b, c)} Z_{C} \times Z \xrightarrow{e'} Y$. Roughly, the following says $c:X \to Z$ is a GE if there exist $\kappa^{\text{p}},\kappa^{\text{r}}:G \curvearrowright Z$, "compatible" with $\mathrm{A}$ and $\mathscr{A}$ respectively, such that $c(x) \not \in \operatorname{Eq}(\kappa^{\text{p}}(g), \kappa^{\text{r}}(g))$ for some $x$.

\begin{proposition}
\label{prop:GE-EM-app}
Assume $Z$ admits an action $\kappa^{\text{p}}:G \curvearrowright Z$ with respect to which $c$ is equivariant. Assume $b$ is equivariant with respect to $\alpha_{X}^{\text{r}}$ and $\rho_{C}^{\text{r}}$, and that $Z$ admits an action $\kappa^{\text{r}}:G \curvearrowright Z$ such that $\mathscr{A}_{(C, O)}^{\prime\top}$ is equivariant with respect to $(\rho_{C}^{\text{r}} , \kappa^{r})$ and $\alpha_{Y}^{\text{r}}$. Finally, suppose $\mathscr{A}_{(C, O)}^{\prime\top}$ is injective\footnote{More generally, it suffices to have injectivity on the pairs of representations witnessing
$\kappa^{\mathrm{p}}(g)c\neq\kappa^{\mathrm{r}}(g)c$.}. Then if $\operatorname{im } c \not \subset \operatorname{Eq}\big( \kappa^{\text{p}}(g), \kappa^{\text{r}}(g) \big)$\footnote{Recall that the \emph{equalizer} of functions $f,g:X \to Y$ is $\operatorname{Eq}(f,g)=\{ x \in X: f(x)=g(x) \} \subset X$.} for all $s_{g} \in S$, $c$ is a GE for $S$. 
\end{proposition}

\begin{proof}
Fix $s_g=\bigl(\alpha_X^{\mathrm r}(g),\alpha_Y^{\mathrm r}(g)\bigr)\in S$. Since $\operatorname{im}c\not\subset\operatorname{Eq}\bigl(\kappa^{\mathrm p}(g),\kappa^{\mathrm r}(g)\bigr)$, there exists $x\in X$ such that $\kappa^{\mathrm p}(g)c(x)\neq\kappa^{\mathrm r}(g)c(x)$. By equivariance of $b$ and $c$,
$$
\bigl(b\alpha_X^{\mathrm r}(g)(x),c\alpha_X^{\mathrm r}(g)(x)\bigr)
=
\bigl(\rho_C^{\mathrm r}(g)b(x),\kappa^{\mathrm p}(g)c(x)\bigr)
\neq
\bigl(\rho_C^{\mathrm r}(g)b(x),\kappa^{\mathrm r}(g)c(x)\bigr).
$$
By injectivity of $\mathscr A_{(C,O)}^{\prime\top}$, there exists $e'\in\mathcal F_{\mathscr A'_{(C,O)}}$ separating these. Since $\mathscr A_{(C,O)}^{\prime\top}$ is equivariant, $e'\bigl(\rho_C^{\mathrm r}(g)b(x),\kappa^{\mathrm r}(g)c(x)\bigr)=\alpha_Y^{\mathrm r}(g)e'\bigl(b(x),c(x)\bigr)$. Thus, for $f':=e'\circ(b,c)\in\mathcal F_{\mathscr A'}$, $f'\alpha_X^{\mathrm r}(g)(x)\neq\alpha_Y^{\mathrm r}(g)f'(x),$ and hence $s_g\notin\operatorname{Sym}\mathscr A'$. Since $s_g\in S$ was arbitrary, $\mathscr A'$ grips every $s\in S$, and therefore $c$ is a GE for $S$.
\end{proof}

The equalizer condition dissolves in an important special case.

\begin{corollary}
\label{cor:ge-em-app-cor}
Under the hypothesis of Proposition~\ref{prop:GE-EM-app}, suppose additionally that $[G:H]=2$, that \(\kappa^\text{r}, \kappa^\text{p}\) are linear\footnote{Strictly speaking, we should also assume $Z$ is over a field of characteristic not $2$. } with $\kappa^{\text{r}}|_{H}=\kappa^{\text{p}}|_{H}$, and that $\kappa^{\text{p}}(s)=-\kappa^{\text{r}}(s)$ for some $s \in G-H$. Then any nonzero $c:X \to Z$ is a GE.
\end{corollary}

\begin{proof}
Fix $s_g\in S$. Since $\kappa^{\text{r}}|_{H}=\kappa^{\text{p}}|_{H}$, necessarily $g\in G-H$. As $[G:H]=2$, we may write $g=sh$ for some $h\in H$. Then from $\kappa^{\mathrm p}|_H=\kappa^{\mathrm r}|_H$,
$$
\kappa^{\mathrm p}(g)=\kappa^{\mathrm p}(s)\kappa^{\mathrm p}(h)=-\kappa^{\mathrm r}(s)\kappa^{\mathrm r}(h)=-\kappa^{\mathrm r}(g).
$$
Since the actions are linear and $\kappa^{\mathrm r}(g)$ is an isomorphism, $\operatorname{Eq}\bigl(\kappa^{\mathrm p}(g),\kappa^{\mathrm r}(g)\bigr)=\{0\}$. Thus if $c$ is nonzero, in particular $\operatorname{im}c\not\subset\{0\}$, and therefore $\operatorname{im}c\not\subset\operatorname{Eq}\bigl(\kappa^{\mathrm p}(g),\kappa^{\mathrm r}(g)\bigr)$. Since this holds for every $s_g\in S$, Proposition~\ref{prop:GE-EM-app} implies that $c$ is a GE for $S$.
\end{proof}

\paragraph{Transformation Behavior and Adaption Cost of GEs}{\color{blue}

Although GEs are simple, principled, general, and have additional applications (cf. Remark~\ref{rmk:ge-remark}), they are far from unique in their ability to correct symmetry slippage. In many settings, other simple repairs are available, whether they be canonicalization, direct retyping of extension-mistyping, data augmentation, or something bespoke. Furthermore, as noted in Section~\ref{sec:discussion}, GEs guarantee that the hypothesis class contains functions that correct the unprescribed symmetry, but do not guarantee that training will find such functions. To investigate whether transformation behavior in practice exploits this restored expressivity, we conduct an experiment comparing GEs to canonicalization, data augmentation, and a hybrid architecture combining an equivariant backbone with an unrestricted branch on a variant of the \(\operatorname{SO}(2) \subset \operatorname{O}(2)\) task of Figure~\ref{fig:motivational-figure}.

\begin{table}[t]
\centering
\caption{}
\label{tab:lambda-sweep}
\resizebox{\columnwidth}{!}{%
\begin{tabular}{llrrr}
\toprule
$\lambda$ & Architecture
& Test MSE ($\times 10^3$)
& Reflection error ($\times 10^3$)
& Epochs $\downarrow$ \\
\midrule
0.1 & Slipped MLP
    & 1.714 & 81.60 & -- \\
    & MLP + GE
    & \textbf{0.007} & \textbf{1.93} & \textbf{42} \\
    & Canonicalization
    & 0.017 & 2.06 & 102 \\
    & MLP + augmentation
    & 0.056 & 6.33 & 517 \\
    & Hybrid
    & 0.023 & 7.58 & 150 \\
\midrule
0.5 & Slipped MLP
    & 41.929 & 408.01 & -- \\
    & MLP + GE
    & \textbf{0.015} & \textbf{3.87} & \textbf{35} \\
    & Canonicalization
    & 0.033 & 5.18 & 96 \\
    & MLP + augmentation
    & 0.633 & 18.02 & 490 \\
    & Hybrid 
    & 0.065 & 11.86 & 125 \\
\midrule
0.9 & Slipped MLP
    & 135.087 & 734.41 & -- \\
    & MLP + GE
    & \textbf{0.026} & \textbf{6.68} & \textbf{71} \\
    & Canonicalization
    & 0.049 & 9.19 & 104 \\
    & MLP + augmentation
    & 0.103 & 12.11 & 704 \\
    & Hybrid 
    & 0.154 & 20.48 & 388 \\
\bottomrule
\end{tabular}%
}
\end{table}

Specifically, we consider a regression task on pairs of vectors $u,v\in\mathbb{R}^2$, with target
\[
f_\lambda(u,v)=\sin(u^\top v)\bigl(1+\lambda\det[u,v]\bigr).
\]
The idea is this target is $\operatorname{SO}(2)$-invariant but sensitive to reflections for $\lambda>0$. The slipped MLP takes only norms and inner products as input as is thus $\operatorname{O}(2)$-invariant. Appending the feature $\det[u,v]$ gives a GE. We compare these against an MLP using an $\operatorname{SO}(2)$-canonicalized representation and an unconstrained MLP trained with rotation augmentation.

Table~\ref{tab:lambda-sweep} reports the results averaged over \(3\) seeds, where ``Epochs" refers to the epochs required to reach validation MSE below $10^{-3}$.  GEs achieve the lowest predictive and reflection errors while reaching the validation threshold in fewer epochs than any of the alternatives. In particular, the learned functions exhibit the required reflection sensitivity while retaining exact prescribed $\operatorname{SO}(2)$-invariance. Thus, in this example, the expressivity restored by GEs is successfully exploited during training, rather than merely being available in the hypothesis class.}

\section{Case Studies}
\label{app:case-studies}
\subsection{Group-Closure Case Studies}

\subsubsection{Time-Orientation in Lorentz-Equivariant ML.}

Many physical tasks are equivariant to subgroups of the \emph{Lorentz group} $\operatorname{O}(1,d)$. The Lorentz group $\operatorname{O}(1,d)$ is a generalization of $\operatorname{O}(d)$, the group of linear automorphisms of Euclidean space, to linear automorphisms of \emph{Minkowski spacetime} $V=(\mathbb{R}^{1,d} ,\langle \_, \_ \rangle)$. Here, $\mathbb{R}^{1,d}:= \underbrace{ \mathbb{R} }_{ \text{ time }t } \times \underbrace{ \mathbb{R}^{d} }_{ \text{space }\boldsymbol x }$ and $\langle \overbrace{ (t, \boldsymbol x) }^{ =:\boldsymbol v }, \overbrace{ (s, \boldsymbol y) }^{ =:\boldsymbol w } \rangle=ts-\boldsymbol x \cdot \boldsymbol y=v^{0}w^{0}-\sum_{i=1}^{d}v^{i}w^{i}$. Like $\operatorname{O}(d)$, $\operatorname{O}(1,d)$ has a preferred subgroup of $\det$-$1$ transformations, called the \emph{proper Lorentz group} and denoted $G:=\operatorname{SO}(1,d)$. Unlike $\operatorname{O}(d)$, $\operatorname{O}(1,d)$ bifurcates into further components based on whether time-orientation is preserved; the subgroup of transformations preserving both space- and time-orientation is called the \emph{proper orthochronous Lorentz group} and denoted $H:=\operatorname{SO}^{+}(1,d)$. Explicitly,
$$H:=\operatorname{SO}^{+}(1,d)= \{  g \in \operatorname{SO}(1,d): gC^{+}=C^{+} \},$$
where $C^{+}:=\{ v \in V: \langle v,v \rangle>0 \text{ and }v^{0}>0 \}$ is called the \emph{future cone} of Minkowski space $V$; equivalently, $\operatorname{SO}^{+}(1,d)$ is the identity component of $G$. Further unlike $\operatorname{O}(d)$, $\operatorname{O}(1,d)$ is noncompact and so many "classical" equivariant learning techniques such as group-averaging don't easily apply. This motivates the implicit or explicit use of techniques from algebraic invariant theory which more readily extend \citep{blumsmith2023machine,villar2021scalars}. Because such techniques build exclusively polynomial features, and $\operatorname{SO}^{+}(1,d)$ is Zariski-dense in $\operatorname{SO}(1,d)$, this puts us in the setting of Proposition \ref{prop:group-closure} by way of Example~\ref{ex:zariski-closure}. This symmetry slippage occurs in each instrinsic Lorentz-equivariant architecture of which the authors are aware~\citep{gong2022lorentznet,bogatskiy2024pelican,bogatskiy2020lgn,spinner2024lgatr}.

Importantly, we do not suspect this symmetry slippage to injure expressivity within the published experiments for those architectures, as said experiments pertain to particle physics tasks whose inputs are all in $\overline{C^{+}}$. That said, any application involving displacements in $V$ (e.g. convolution) or more general directed relations (e.g. spacetime embeddings of directed graphs~\citep{saez2025neural}) may present an expressivity barrier. To demonstrate this, we compare the four architectures with and without GEs on a standard neural PDE task. The goal of neural PDE~\citep{LiEtAl2021,KovachkiEtAl2023} is to learn the solution operator $S: s \mapsto u$ of a PDE $Lu=s$. When $L$, $s$, and the boundary conditions are nice enough, $S$ is determined by convolution with a function $h$ (called the \emph{fundamental solution}), so that
$$u=S(s)= s  * h.$$
For constant-coefficient hyperbolic PDEs in Minkowski space (e.g. the wave equation), there is a canonical $h$ satisfying $\operatorname{supp}h \subset  \overline{C^{+}}$ called the retarded fundamental solution. Our experiment focuses on learning the retarded solution to the wave equation $\square u= s$ in $2+1$ dimensions. In this setting, $h$ admits a closed form given by $h(\Delta)= \frac{1}{2\pi}\frac{ \boldsymbol 1_{\{ \Delta^{0}>0 \}} \boldsymbol 1_{\{ \langle \Delta, \Delta \rangle>0 \}} }{\sqrt{ \langle \Delta, \Delta \rangle }}.$ Following standard discretization-invariant neural-operator setups~\citep{LiEtAl2021,KovachkiEtAl2023}, we sample irregular spacetime point clouds $P=(v_{1},\dots,v_{n}) \in V^{n}$ and scalar source fields $s=(s_{1},\dots,s_{n})$, and approximate the solution operator per the discrete convolution $\sum_{j}w_{j}h^{\theta }(v_{i}-v_{j})s_{j}$. With $X:=V^{n} \times \mathbb{R}^{n}$, $Y=\mathbb{R}^{n}$ a published model $\mathscr{M}$ gives rise to an architecture $\mathscr{A}_{\mathscr{M}}:X \rightsquigarrow Y$ parameterizing functions
$$f^{\theta}:(P, s) \xmapsto{b} (\{ \Delta_{ij} \} ,s) \xmapsto{} \left\{  h^{\theta}_{\mathscr{M}}\left( \Delta_{ij}\right)  \right\} \mapsto \hat{u},$$
where $\Delta_{ij}=v_{i}-v_{j}$, $\hat{u}_{i}=\sum_{j \neq i}w_{j}f^{\theta}(\Delta_{ij})s_{j}$, and $h^{\theta}_{\mathscr{M}}$ is a convolutional kernel parameterized by $\mathscr{M}$.

For $Z=\{ -1,0,1 \}^{n \times n}$, we define a GE $c:X \to Z$ given by $c\big((v_{1},\dots, v_{n}), s\big)_{ij}:=\boldsymbol 1_{\langle v_{i}-v_{j}, v_{i}-v_{j} \rangle>0}\operatorname{sign}(v_{i}^{0}-v_{j}^{0})$. Since $g \in G$ reverse time-orientation iff $g \not \in H$, we have $c(g(v_{1},\dots,v_{n}), s)=c(v_{1},\dots,v_{n}, s)$ if $g \in H$ and $c(g(v_{1},\dots,v_{n}))=-c(v_{1},\dots,v_{n},s)$ if $g \in G-H$, so $c$ is indeed a GE. Note that $c$ is clearly non-polynomial, as Example~\ref{ex:zariski-closure} would require. We append $c$ at the cut $C$ of $\mathscr{A}$ preceding the application of $h^{\theta}_{\mathscr{M}}$, so that $Z_{C}=V^{\{ (i,j) : i\neq j \}} \times \mathbb{R}^{n}$ and $\mathscr{A}_{\mathscr{M}}'=\mathscr{A}_{\mathscr{M}}\oplus_{C} c$ is defined by

\begin{align}
X & \xrightarrow{(b,c)}Z_{C} \times Z  \xrightarrow{ e ^{\theta'}} Y \\
x=(P, s) & \xmapsto{(b, c)} \big( ( \Delta_{ij} , s)_{i \neq j}, (c_{ij})_{i \neq j} ) \big) \xmapsto{} \{{h}^{\theta'}_{\mathscr{M}}(\Delta_{ij}; c_{ij}) \}_{i \neq j}\xrightarrow{}\hat{u},
\end{align}

where the notation ${h}^{\theta'}_{\mathscr{M}}(\Delta_{ij}; c_{ij})$ indicates that $c_{ij}$ has been appended as an input feature of $\mathscr{M}$.  

On this task we evaluate four architectures: LorentzNet~\citep{gong2022lorentznet}, PELICAN~\citep{bogatskiy2024pelican}, LGN~\citep{bogatskiy2020lgn}, and L-GATr~\citep{spinner2024lgatr}\footnote{L-GATr~\citep{spinner2024lgatr} also uses fixed reference multivectors such as $e_{0}=(1,0,0,0)$, for supplying time-orientation externally. Unlike a GE such as our $c$, however, a fixed $e_{0}$ need not preserve the full prescribed $\operatorname{SO}^{+}(1,d)$ symmetry.}. In practice, we encode the displacement $\Delta_{ij}$ as the tuple $\left( \frac{1}{2}\Delta_{ij}, -\frac{1}{2}\Delta_{ij} \right)$ when inputting into these architectures, as they can become trivial when operating on a single object. For each architecture, we compare performance when an arbitrary scalar feature $z$ has been appended $h^{\theta'}_{\mathscr{M}}(\Delta_{ij}; z)$ versus when the GE $c$ has been appended. The results are shown in Table~\ref{tab:lorentz}.

\begin{table}[t]

\centering
\small
\begin{tabular}{lcccc}
\toprule
Architecture & ID NMSE --- Raw & ID NMSE --- +GE & OOD NMSE --- Raw & OOD NMSE --- +GE \\
\midrule
\textbf{LorentzNet} \citep{gong2022lorentznet}
& $0.6399 \pm 0.0240$
& \textbf{$0.0688 \pm 0.0253$}
& $0.5886 \pm 0.0220$
& \textbf{$0.0576 \pm 0.0217$} \\
\textbf{PELICAN} \citep{bogatskiy2024pelican}
& $0.5154 \pm 0.0013$
& \textbf{$0.0459 \pm 0.0013$}
& $0.4784 \pm 0.0011$
& \textbf{$0.0343 \pm 0.0009$} \\
\textbf{LGN} \citep{bogatskiy2020lgn}
& $0.5401 \pm 0.0170$
& \textbf{$0.0335 \pm 0.0031$}
& $1.1434 \pm 0.8805$
& \textbf{$0.1271 \pm 0.0879$} \\
\textbf{L-GATr} \citep{spinner2024lgatr}
& $0.5129 \pm 0.0275$
& \textbf{$0.0545 \pm 0.0118$}
& $0.4847 \pm 0.0348$
& \textbf{$0.0542 \pm 0.0129$} \\
\bottomrule
\end{tabular}
\caption{}
\label{tab:lorentz}
\end{table}

\subsubsection{Signs in Units-Equivariant ML}

Most numerical quantities in scientific learning problems have units on which the underlying laws of physics do not depend. Firsthand experience of science's ``units-equivariance" is provided by the \emph{dimensional analysis} practiced in introductory science classes. Dimensionless representations and related data-driven dimensional methods have been used for discovery of dimensional structure \cite{Constantine2017,Evangelou2021}, dynamical-system identification \cite{Bakarji2022}, and symbolic regression \cite{villar2023dimensionless}, among other machine-learning settings \cite{Rudolph1998,FrisoneMisiti2019}.

To describe this symmetry, suppose an input $x=(x_{1},\dots,x_{d}) \in X$ consists of $d$-many measured quantities $x_{i}$, expressed in $k$ "base units" according to associated vectors $u_{i} \in \mathbb{Z}^{k}$, where the $j$th entry of $u_{i}$ is the exponent of the $j$th base unit in $x_{i}$. For instance, a classical dynamics setup might have $k=3$ base units ($\mathrm{m}$, $\mathrm{s}$, $\mathrm{kg}$); in this case,  a velocity $x_{i}$ would have $u_{i}=(1,-1, 0)$, while a force $x_{\ell}$ would have $u_{\ell}=(1, -2, 1)$. A change of base unit is specified by positive "conversion factors" $g=(g_{1},\dots,g_{k})\in H:= (\mathbb{R}^{+})^{k}$ and acts on inputs by
$$gx:=\big( g_{1}^{-u_{1 1}}\dots g_{k}^{-u_{1k}}x_{1}, \dots, g_{1}^{-u_{d 1}}\dots g_{k}^{-u_{dk}}x_{d} \big)$$
and on an output $y$ with associated vector $v \in \mathbb{Z}^{k}$ by $gy= g_{1}^{-v_{1}}\dots g_{k}^{-v_{k}}y .$ Units consistency then amounts to equivariance with respect to this $H$-action. If $y$ is dimensionless (e.g. in a classification task), then $v=0$ and the goal is $H$-invariance. 

In the setting of \citep{villar2023dimensionless}, a \emph{Buckingham $\Pi$-group} is a Laurent monomial $\Pi(x)=\prod_{i=1}^{d}x_{i}^{a_{i}}$ that is dimensionless, i.e., that has $Ua=\sum_{i}a_{i}u_{i}=0$. The \emph{Buckingham-$\Pi$ Theorem}~\citep{Buckingham1914} roughly states that if a physical relationship $F(x_{1},\dots,x_{d})=0$ holds independently of the choice of units then there exists a $\Psi$ and Buckingham $\Pi$-groups $\Pi_{1}(x),\dots,\Pi_{d-r}(x)$,  $r=\operatorname{rank}U$, such that $F(x)=0$ iff $\Psi(\Pi_{1}(x),\dots, \Pi_{d-r}(x))=0$. Leveraging this principle, \citep{villar2023dimensionless} perform unconstrained learning after applying the dimensionless feature map $x \mapsto \big(\Pi_{1}(x),\dots,\Pi_{d-r}(x)\big)$ for $\Pi_{a}$ obtained by solving the (integer) linear system $Ua=0$. 

Let $\mathcal{C}=K(X)$, and write $G:=(\mathbb{R}^{\times})^{k}$. $G$ acts on $X,Y$ by evident extension of the $H$-actions. Conceptually, the $G$-action permits nonphysical sign reversals of base units whereas $H$ consists only of legitimate, positive changes of base units. Now, $\overline{H}^{\text{Zariski}}=G$, and the $\Pi$-features used by \citep{villar2023dimensionless} are Laurent monomials, hence regular functions on $(\mathbb{R} ^{\times})^{d}$. Therefore, Proposition \ref{prop:group-closure} gives, by way of Example \ref{ex:zariski-closure}, symmetry slippage for any architecture built solely from $H$-equivariant features. While this symmetry slippage is innocuous in many settings, such as when the target only depends on magnitude, it can pose a problem in any application wherein the sign of a measured quantity matters to the target. For $Z\overset{}{=} \{ -1,0,1 \}^{d} \subset \mathbb{R}^{d}$ and $G$ acting trivially on $Z$, we define a GE $c:X \to Z$ by $c(x)_{i}=\operatorname{sign}(x_{i})$. Then $c$ is $H$-equivariant but not $G$-equivariant. A general architecture $\mathscr{A}$ of the form in \citep{villar2023dimensionless} realizes functions factoring as
$$X \xrightarrow{b=\phi_{\Pi}} Z_{C}=\mathbb{R}^{d-r}\xrightarrow{e^{\theta}}Y.$$
Appending $c$ at $C$, $\mathscr{A}'=\mathscr{A}\oplus _Cc$ is then
\begin{align}
X &\xrightarrow{(b, c) }Z_{C} \times Z \xrightarrow{e^{\theta'}} Y \\
x &\xmapsto{(b, c)} \big( (\Pi_{1}(x), \dots, \Pi_{d-r}(x)), c(x) \big) \xrightarrow{e^{\theta'}} \hat{y}.
\end{align}
For concreteness, we assume $\mathscr{A}_{(C, O)}$ is an $\operatorname{MLP}$, as in the experiments of \citep{villar2023dimensionless} and our own, in which case $e^{\theta'}$ is defined as in the MLP example in Section \ref{sec:GEs} and in which case we know $\mathscr{A}_{(C, O)}^{\prime\top}$ is injective. We conclude from Proposition~\ref{prop:GE-GC-app} that $c$ is indeed a GE for $\mathscr{A}$. 

To demonstrate the repair, we compare architectures with and without GEs on a natural time-series forecasting task from the Mauna Loa record of atmospheric $\mathrm{CO}_2$ concentration \citep{Keeling1976}. At each time $t$, the input $h_t=(h_{t,0},\ldots,h_{t,11})\in\mathbb R^{12}=X$ consists of the preceding twelve weekly changes in $\mathrm{CO}_2$ concentration, and the target $y_t=\boldsymbol 1_{q_{t+4}>q_t} \in Y=\{ 0,1 \}$ records whether the concentration $q$ four weeks later exceeds its current value. All input coordinates have the same unit $k=1$, so $H=\mathbb R^{+}$ acts by $h_t\mapsto ah_t$ and trivially on $Y$. The task is $H$-invariant, since a positive change of concentration units preserves whether future concentration is higher. It is not $G=\mathbb R^\times$-invariant, since $h_t\mapsto-h_t$ reverses increases and decreases and so need not preserve the target. 

We have $U=(1,\ldots,1)\in\mathbb Z^{1\times 12}$, with $\operatorname{ker }U$ having integer basis $a_{j}=e_{j}-e_{0}$. This gives $\Pi_{j}(h)=\frac{h_{j}}{h_{0}}$, $j=1,\dots,1 1$. Taking $e^{\theta}$ to be parameterized by a width-$32$ $\operatorname{ReLU}$ $\operatorname{MLP}$ with two hidden layers, we append GEs as above. We also evaluate against two controls given given by the GE alone and augmentation by a split-wise shuffled GE. Results are shown in Table~\ref{tab:units}. 
\medskip

\begin{remark}
    The same group-closure occurs in weight-space learning for ReLU networks \citep{neyshabur2015pathsgd,meng2019gsgd}. For a network with $k$ independently rescalable hidden neurons, positive homogeneity of ReLU entails the prescribed law $\mathrm{A}=\underline{H}_{\alpha}^{I}$ for $H:=(\mathbb{R}_{+})^{k}$, but, as above, the Zariski-closure of \(H\) is $G:=(\mathbb{R}^{\times})^{k}$. Since path and basis-path coordinates are polynomial in the network weights \citep{meng2019gsgd,gonon2025rescaling}, $H$-invariance of these coordinates extends automatically to $G$-invariance, causing symmetry slippage by Example~\ref{ex:zariski-closure}. This gives a group-closure interpretation of existing sign-based heuristics. For instance, \citet{meng2019gsgd} supplement basis-path values with signs of free skeleton weights. Letting $c:X\to Z$ denote the retained sign information and letting $G$ act trivially on $Z$, one has $c\alpha(h)=c$ for every $h\in H$, whereas $c\alpha(g)\neq c$ for each signed rescaling $g\in G-H$ detected by $c$. Thus, Proposition~\ref{prop:main-GE-prop} implies that $c$ is a GE for $S=G-H$. 
\end{remark}

\begin{table}[t]
\centering
\small
\begin{tabular}{lc}
\toprule
Architecture & Accuracy \\
\midrule
Baseline & $58.44\pm1.30\%$ \\
\textbf{Baseline $+\mathrm{GE}$} & \textbf{$82.04\pm0.56\%$} \\
GE only & $58.53\pm0.00\%$ \\
Baseline + Shuffled GE & $58.25\pm1.09\%$ \\
\bottomrule
\end{tabular}
\caption{}
\label{tab:units}
\end{table}

\subsubsection{Inner Automorphisms in Lie Algebra Learning}

Lie-algebra-valued representations arise naturally in geometric learning problems such as
dynamical-system modeling, registration, and homography-based vision
\citep{LinEtAl2024LieNeurons}. It is desirable for an architecture $\mathscr{A}$ for learning on Lie algebras to be coordinate-independent in an appropriate sense. More precisely, the Lie algebra $(\mathfrak{k}, [\_, \_])$ of a Lie group $K$ carries a linear action of $K$ by conjugation, known as the \emph{Adjoint representation}. $\operatorname{Ad}:K \to \operatorname{Aut}\mathfrak{k}$. Specifically, $g \xrightarrow{\operatorname{Ad}}dC_{g} |_{e}$, where $dC_{g} |_{e}$ is the differential of $C_{g}:h \mapsto gh g ^{-1}$ at $e$. The Lie algebra $\mathfrak{g}$ also acts on itself via the \emph{adjoint representation} ("small-a") $\operatorname{ad}:\mathfrak{k} \to \mathfrak{gl}(\mathfrak{k})$, $x \mapsto [x,\_]$; for $K$ connected,  $H:=\operatorname{Ad}(K)= \langle \exp(\operatorname{ad}_{x}) :x \in \mathfrak{k}\rangle=:\operatorname{Int}(\mathfrak{k})$ is the \emph{inner automorphism group} of $\mathfrak{k}$. With these definitions in place, the desired notion of equivariance in Lie algebra learning is $\operatorname{Ad}$-equivariance, or equivalently $\operatorname{Int}$-equivariance. 

To this end, the Lie neurons program~\citep{LinEtAl2024LieNeurons,KimEtAl2026ReLN} generalizes the well-known vector neurons approach~\citep{deng2021vector}, which may be viewed as developing $\operatorname{SO}(3)$-equivariant learning for $(\mathbb{R}^{3}, \times)\cong \mathfrak{so}(3)$, to more general matrix Lie algebras $\Sigma=(\mathfrak{k}, [\_,\_])$. Lie neuron-based architectures $\mathscr{A}$ are built from the function class $\mathcal{C}$ determined by linear channel mixing operations, $[\_, \_]$-operations, and $B(\_, \_)$-operations, where $B$ denotes the reductive invariant bilinear form on $\mathfrak{k}$ (e.g. the Killing form $\kappa$ when $\mathfrak{k}$ is semisimple). Such operations are equivariant not only to inner automorphisms $H=\operatorname{Int}(\Sigma)$, but to the full automorphism group $G=\operatorname{Aut}(\Sigma)$, leaving us in the setting of Example \ref{example:automorphism-closure} with $\mathcal{C}$ closing $H$ to $G$, and thus producing symmetry slippage whenever $\operatorname{Int}(\Sigma) \lneq \operatorname{Aut}(\Sigma)$. 

While this symmetry slippage does not pose an expressivity problem if the data and task refer to a single $H$-orbit, it can become problematic on tasks whose targets transform nontrivially under $G / H$ (called \emph{outer automorphisms} of $\Sigma$). For concreteness, we focus on the case $\mathfrak{k}=\mathfrak{sl}_{3}$. In this case $G / H \cong \mathbb{Z}_{2}$, represented by the outer automorphism $\tau(x)=-x^{\top}$. For $x=(x_{1},\dots,x_{d}) \in \mathfrak{sl}_{3}^{d}=:X$,  let $Z=\mathbb{R}^{d}$ with $G$ acting trivially, and define $c:X \to Z$ by $c(x)_{i}= \operatorname{tr}(x_{i}^{3})=3 \det x_{i}$. Then $c$ is $H$-equivariant, since trace is conjugation-invariant, hence inner automorphism-invariant, but not $G$-equivariant, since $c(\tau x)=-c(x)$. A Lie-Neuron architecture $\mathscr{A}$ realizes functions factoring as
$$X \xrightarrow{b^{\theta}} Z_{C}=\mathbb{R}^{m} \xrightarrow{e^{\theta}}Y$$
where $b^{\theta}$ is given by $X \xrightarrow{b_{1}} \mathfrak{sl}_{3}^{m} \xrightarrow{b_{2}}\mathbb{R}^{m}$ for $b_{1}$ the Lie Neuron equivariant backbone and $b_{2}( w )_{j}=\kappa(w_{j}, w_{j})$ and $e^{\theta}$ is parameterized by a $\mathscr{A}_{(C, O)}=\operatorname{MLP}$. Appending $c$ at $C$, $\mathscr{A}'=\mathscr{A}\oplus_{C}c$ gives
\begin{align}
X &\xrightarrow{(b^{\theta}, c)} Z_{C} \times Z \xrightarrow{e^{\theta'}}Y \\
x &\mapsto \big( b^{\theta}(x), c(x) \big) \mapsto \hat{y},
\end{align}
where $e^{\theta'}$ defined per the $\mathscr{A}_{(C, O)}=\operatorname{MLP}$ example in Section \ref{sec:GEs}. Since $\mathscr{A}_{(C, O)}'=\operatorname{MLP}$, we know $\mathscr{A}_{(C, O)}^{\prime\top}$ is injective. We conclude from Proposition~\ref{prop:GE-GC-app} that $c$ is indeed a GE for $\mathscr{A}$. 

To demonstrate the repair, we compare Lie-Neuron architectures with and without GEs on a natural continuum mechanics task: learning the strain-energy function $W=W(F)$ of a hyperelastic material. We use the classical experiments of Treloar \citep{treloar1944stress}, which measure the response of vulcanized rubber under several homogeneous deformation modes, including uniaxial tension (UT) and equi-biaxial-tension (BT). These data have become a standard benchmark for data-driven constitutive modeling \citep{dammass2025invariants,linka2021constitutive,holthusen2026treloar, linden2023neural}. For an incompressible deformation $F$, the Cauchy-Green tensor $F^{\top}F$ has $\det F^{\top}F=1$, meaning $F^{\top}F \in \operatorname{SL}(3)$, and so the true strain $\varepsilon=\frac{1}{2}log(F^{\top}F) \in \mathfrak{sl}_{3}$. For an isotropic material, $W$ is invariant under orthogonal conjugation, so that architectures equivariant to the restriction of $\operatorname{Ad}:\operatorname{SL}(3) \to \operatorname{Aut}\mathfrak{sl}_{3}$ to $\operatorname{SO}(3)$, such as Lie Neurons (LN) \citep{LinEtAl2024LieNeurons} or Reductive Lie Neureons (ReLN) \citep{KimEtAl2026ReLN}, are appropriate for the task. 

Following \citep{dammass2025invariants}, we fit $W$ to Treloar's measured UT and BT responses. The dataset contains $24$ UT and $16$ BT stress measurements. For each deformation $F$, we use the true strain $\varepsilon(F)= \frac{1}{2} \log (F^{\top}F)$ as input, and the network predicts $W(\varepsilon)$ by minimizing $\mathcal{L}=\operatorname{MSE}_{\operatorname{UT}}+3 \operatorname{MSE}_{\operatorname{BT}}$. We compare LN and ReLN with and without the GE $c(\varepsilon)=\operatorname{tr}(\varepsilon^{3})$ on the task. As controls, we also add an identical model with an added feature, and one with the even feature $c(\varepsilon)^{2}$. Results are shown in Table~\ref{tab:IntAut}.

\begin{table}[t]

\centering
\small
\begin{tabular}{llccc}
\toprule
Architecture & Arm & $\mathcal L_{\mathrm{UT}+3\mathrm{BT}}$ & UT $R^2$ & BT $R^2$ \\
\midrule
LN \citep{LinEtAl2024LieNeurons}
& Baseline
& $0.2817 \pm 0.0844$
& $0.9259 \pm 0.0198$
& $0.9874 \pm 0.0087$ \\
LN
& matched-zero
& $0.2776 \pm 0.0453$
& $0.9258 \pm 0.0115$
& $0.9900 \pm 0.0028$ \\
LN
& $+c^2$
& $0.2408 \pm 0.0214$
& $0.9352 \pm 0.0055$
& $0.9922 \pm 0.0013$ \\
LN
& \textbf{$+c$ (GE)}
& \textbf{$0.0093 \pm 0.0071$}
& \textbf{$0.9978 \pm 0.0015$}
& \textbf{$0.9992 \pm 0.0010$} \\
\midrule
ReLN \citep{KimEtAl2026ReLN}
& Baseline
& $0.2609 \pm 0.0456$
& $0.9301 \pm 0.0116$
& $0.9910 \pm 0.0029$ \\
ReLN
& matched-zero
& $0.2524 \pm 0.0254$
& $0.9323 \pm 0.0065$
& $0.9915 \pm 0.0015$ \\
ReLN
& $+c^2$
& $0.2443 \pm 0.0134$
& $0.9342 \pm 0.0034$
& $0.9922 \pm 0.0009$ \\
ReLN
& \textbf{$+c$ (GE)}
& \textbf{$0.0100 \pm 0.0033$}
& \textbf{$0.9975 \pm 0.0009$}
& \textbf{$0.9992 \pm 0.0004$} \\
\bottomrule
\end{tabular}
\caption{}
\label{tab:IntAut}
\end{table}

\subsection{Over-Quotienting Case Studies}

\label{app:over-quotienting-full-case-studies}
\subsubsection{Roots in Glycan Learning}
Glycans are carbohydrate molecules produced by organisms across the three domains of life (Eukarya, Bacteria, Archaea). A glycan is made of multiple simple sugars (monosaccharides) linked together by chemical bonds (called glycosidic linkages). Many glycans have a distinguished sugar, called the \emph{reducing end}, that contains a free anomeric carbon. These glycans are commonly abstracted as trees whose nodes are sugars and whose edges are bonds, who are rooted at the reducing end, and whose nodes are colored by sugar type and edges by bond type. Define $X_{n}$ to be the collection of rooted Glycan trees,
$$X_{n}:= \{ (T, r): T \text{ is a Glycan tree on }[n], r \in [n] \}.$$
The group $\Gamma:=H:=S_{n}$ acts on $X_{n}$ by simultaneously relabeling nodes and roots, $\alpha_{X}(\pi)(T, r)=(\pi T, \pi r)$. We may take $J:=\operatorname{Bij}(X_n)$.
Graph neural networks (GNNs) \citep{scarselli2009graph,gilmer2017neural,schlichtkrull2018modeling,vashishth2020composition} have recently been used extensively for glycan learning. Popular examples include GlycanML \citep{xu2025glycanml}, which employs a GNN backbone to aid in taxononimization tasks (e.g. Eukarya vs. Bacteria vs. Archaea), and GlycanAA \citep{xu2025glycanaa}, which refines such approaches with atom-level structure and hierarchical message passing. The prescribed transformation law of graph-based approaches to glycan learning is $\mathrm{A}=\{\big( \alpha_{X}(\pi), I_{Y} \big):\pi \in S_{n}\}$. In both aforementioned architectures $\mathscr{A}$, however, the GNN backbone operates on unrooted trees, in doing so over-quotienting the reducing end. To see this, take $Z_{C_{1}}=X_{n}$ and $Z_{C_{2}}=\{  T:(T, r) \in X_{n} \}$, whence $\mathscr{A}$ factors as
$$X_{n} \xrightarrow{b=I_{X_{n}}}Z_{C_{1}} \xrightarrow{\Phi^{}} Z_{C_{2}} \xrightarrow{d^{\theta}}Y$$
for $\Phi:(T, r) \mapsto T$ and $d^{\theta}$ parameterized by the GNN backbone. Define an equivalence relation $\sim$ on $Z_{C_{1}}$ by $(T_{1}, r_{1}) \sim (T_{2}, r_{2}) \iff T_{1} = T_{2}$ as colored graphs, and define $\sigma=\beta_{\kappa}$ for some $\kappa \neq e$, where $\beta_{\kappa}\in J$ is given by $\beta_{\kappa}(T,r)=(T,\kappa r)$. We see that $b \sigma(x) \sim b(x)$ for all $x \in X$. Furthermore, $\Phi^{}$ is by construction constant on $\sim$-classes and $(\sigma, I_{Y}) \not\in \mathrm{A}$, and so the over-quotienting conditions are satisfied and symmetry slippage occurs per Proposition \ref{prop:over-quotienting}. 

On many biological tasks, the reducing-end location carries relevant information. For an especially simple instance, in the dataset used by \citep{xu2025glycanml} for their domain-taxonomization tasks (e.g. predicting whether a glycan was produced by Eukarya, Bacteria, Archaea, or Virus),  $x_{B}=[\operatorname{Glc}(\beta 1 - 2)]\text{Glc}(\beta 1 - 3)\text{Glc}$ occurs in records labeled Bacteria while $x_{E}=\operatorname{Glc}(\beta 1 - 2)[\text{Glc}(\beta 1 - 3)]\text{Glc}$ occurs in records labeled Eukarya, despite the fact that over-quotienting implies every $f^{\theta}$ parameterized by $\mathscr{A}$ satifies $f^{\theta}(x_{E})=f^{\theta}(x_{B})$. 

Let $Z=\mathbb{R}^{[n]}$ and define $c:X_{n} \to Z$ by $c(T, r)_{v}:= \boldsymbol 1_{\{ v=r \}}$, so that $c(T, r)$ is the indicator vector for the tree's root node. Given a slipped transformation $s_{\kappa} = \big( \beta_{\kappa} , I_{Y}\big)$, $\kappa \neq e$, take $\sigma=\beta_{\kappa}$. Then $c \sigma(T ,r)=e_{\kappa r}$ and $e_{r}=c(T, r)$. Since the action by $S_{n}$ is faithful, it follows that $c \sigma \neq c$. Putting $C:=C_{2}$, $\mathscr{A}$ realizes functions factoring as
$$X_{n} \xrightarrow{b_{}} Z_{C} \xrightarrow{e^{\theta}}Y$$
where $b=\Phi \circ a$ and $e^{\theta}$ is the downstream GNN. (For GlycanAA, $e^{\theta}$ additionally contains a deterministic all-atom expansion before its hierarchical message-passing network.) Appending $c$ at $C$, $\mathscr{A}'=\mathscr{A} \oplus_{C} c$ gives
\begin{align}
X_{n} &\xrightarrow{(b ,c)}Z_{C} \times Z \xrightarrow{e^{\theta'}} Y \\
(T, r) &\mapsto (T, e_{r})  \mapsto \hat{y},
\end{align}
where $e^{\theta'}$ is implemented by supplying $c_{v}$ as an additional node feature to the GNN. To apply Proposition \ref{prop:GE-OQ-app} and thus verify $c$ is a GE, it remains to check that $\mathscr{A}_{(C, O)}^{\prime\top}$ is injective on the pairs of representations witnessing $c\sigma \neq c$, that is, that  $\mathscr{A}_{(C,O)}^{\prime\top}$ separates a pair of representations witnessing $c\sigma\neq c$.  Indeed, let $\sigma=\beta_\kappa$ with $\kappa\neq e$, and choose $r\in[n]$ such that $\kappa r\neq r$. Choose $T$ so that $\deg_T(r)\neq \deg_T(\kappa r)$. Then the representations at $Z_{C}$ are $z=(T,e_r)$ and $z'=(T,e_{\kappa r})$. a usual sum-aggregation message-passing layer can propagate the root indicator to its neighbors and, after pooling, realize
$$\psi(T,e_r)=\sum_{u\in[n]}\sum_{v\in N_T(u)}(e_r)_v=\deg_T(r).$$
Hence $\psi(z)=\deg_T(r)\neq\deg_T(\kappa r)=\psi(z')$, so $\mathscr{A}_{(C,O)}^{\prime\top}$ separates this pair.

To demonstrate the repair, we compare GlycanML and GlycanAA with and without GEs on the proposed domain-level taxonomy prediction task benchmarked in both papers. Given a glycan, the task is to predict the domain (Eukarya, Bacteria, Archaea, or Virus) of the organism with which it is associated. We compare the released RGCN and CompGCN of GlycanML and the richer GlycanAA architecture, each with and without GEs, on their released $13209$ domain-taxonomy records using $5$-fold cross-validation with records having the same rootless graph placed in the same group. 
The results are shown in Table \ref{tab:Glycan}.

\begin{table}[t]
\centering
\small
\begin{tabular}{lcc}
\toprule
Architecture & Macro-F1 & + GE Macro-F1 \\
\midrule
GlycanML, RGCN \citep{xu2025glycanml}
& $0.6001 \pm 0.0250$
& \textbf{$0.6358 \pm 0.0193$} \\
GlycanML, CompGCN \citep{xu2025glycanml}
& $0.6250 \pm 0.0242$
& \textbf{$0.6710 \pm 0.0212$} \\
GlycanAA \citep{xu2025glycanaa}
& $0.6409 \pm 0.0420$
& \textbf{$0.6871 \pm 0.0067$} \\
\bottomrule
\end{tabular}
\caption{}
\label{tab:Glycan}
\end{table}

Note that this case study is over-quotienting, but not over-equivariance.

\subsubsection{Normalization in ECG ML}

Amplitude normalization is a common practice in ML for ECG analysis \citep{ma2021ecg,weimann2024federated,zeng2026stabilizing,hu2026one,li2025ecgfounder}. Let $\mathbb{R}^{m \times T}$ denote the space of $m$-lead ECG signals. In the broadest sense, by a \emph{normalization rule} we mean any function $q:X \subset \mathbb{R}^{m \times T} \to X$ satisfying $q(\lambda x)=q(x)$ for all $x \in X$, $\lambda>0$. Equivalently, with $\Gamma:=\mathbb{R}^{\times}\overset{\alpha}{\curvearrowright}{X}$ acting by scaling $\alpha_{\lambda}(x)=\lambda x$, a normalization rule $q$ is an $G:=\mathbb{R}^{+}$-invariant function on $X$. While many normalization rules exist, each imposes at least this symmetry. Let ${H}=(e) \leq G \leq \Gamma$. For certain tasks, a trivial transformation law $\mathrm{A}=\underline{H}_{\alpha}^{I}=\underline{( e )}_{\alpha}^{I}=\{ (I_{X}, I_{Y}) \}$ is prescribed, and a normalizing architecture over-quotients it, incurring symmetry slippage. To see this, consider an architecture realizing functions $d^{\theta}:X \to Y$ and let $\mathscr{A}$ be the architecture obtained by imposing the normalization rule $q$, so that $\mathscr{A}$ realizes functions factoring as

\[
\begin{tikzcd}
X \arrow[r, "a=I"] \arrow[rrr, "b", bend left]
& Z_{C_1} \arrow[r, "q"] \arrow[rr, "\Phi^\theta"', bend right]
& Z_{C_1} \arrow[r, "d^\theta"]
& Z_{C_2} \arrow[r, "e^\theta"]
& Y
\end{tikzcd}
\]

where $Z_{C_{1}}=X$, $Z_{C_{2}}$ is the original architecture's pooled feature space before its affine readout $e^{\theta}$. Let $\sim_{G}$ be the orbit relation of $G \curvearrowright Z_{C_{1}}$. For $\sigma=\alpha_{\lambda}$, $\lambda \neq 1$, we clearly have $a\sigma(x) \sim_{G} a(x)$ for all $x \in X$. Furthermore, $\Phi$ is by construction constant on $\sim_{G}$-classes and $(\sigma, I_{Y}) \not \in \mathrm{A}$, and so the over-quotienting conditions are satisfied and symmetry slippage occurs per Proposition \ref{prop:over-quotienting}, with $\{ (\alpha_{\lambda}, I_{Y}): \lambda>0, \lambda \neq 1 \} \subset \operatorname{Sym}\mathscr{A}-\mathrm{A}$. We examine two common ECG normalization rules. One rule, found e.g. in the recent ECG foundation model ECGFounder~\citep{li2025ecgfounder}, applies the $z$-score
$$q_{\text{glob}}(x)= \frac{x - \texttt{mean}(x) \boldsymbol  1}{\texttt{stdev}(x)}.$$
A more refined rule, found e.g. in \citep{weimann2024federated}, standardizes leadwise thus:
$$q_{\text{lead}}(x)_{j, :}= \frac{x_{j, :} - \texttt{mean}(x_{j, :})}{\texttt{stdev}(x_{j, :})}.$$
While normalization is of course deliberate and often desirable, this over-quotienting can also erase task-relevant information. Two particularly direct examples may be found in the well-known PTB-XL benchmark \citep{strodthoff2021ptbxl}, which include e.g. the targets LVOLT (low QRS voltage) and VCLVH (voltage criteria for left-ventricular hypertrophy) that depend on amplitude. 

Let $Z:=\mathbb{R}$ and define $c:X \subset \mathbb{R}^{m \times T} \to Z$ by $c(x)=\log\frac{\|x\|_{2}}{\sqrt{ mT }}$. Given a slipped transformation $s_{\lambda}=(\alpha_{\lambda}, I_{Y})$, $\lambda>0$, $\lambda \neq 1$, take $\sigma=\alpha_{\lambda}$. Then $c \sigma(x)=\log \frac{\| \lambda x \|_{2}}{\sqrt{ mT }}=\log \lambda+ c(x) \neq c(x)$. Putting $C:=C_{2}$, $\mathscr{A}$ realizes functions factoring as $X \xrightarrow{ b} Z_{C} \xrightarrow{e^{\theta}}Y$; appending $c$ at $C$, $\mathscr{A}'=\mathscr{A} \oplus_{C} c$ gives
\begin{align}
X &\xrightarrow{(b,c)} Z_{C} \times Z \xrightarrow{e^{\theta'}} Y \\
x &\mapsto \big(b(x), \log \frac{\|x\|_{2}}{\sqrt{ m T }}\big) \mapsto \hat{y},
\end{align}
where $e^{\theta'}$ is defined per the $\mathscr{A}_{(C,O)}=\operatorname{MLP}$ example in Section \ref{sec:GEs}. 

To demonstrate the repair, we compare the two architectures \citep{li2025ecgfounder} and \citep{weimann2024federated} with and without GEs on the PTB-XL benchmark \citep{wagner2020ptbxl,strodthoff2021ptbxl}. We first take the PTB-XL multilabel classification task and restrict to the two labels LVOLT and VCLVH which depend explicitly on amplitude. We then compare the architectures in their published settings. For \citep{li2025ecgfounder}, we follow its released PTB-XL setting, retaining the pretrained normalized architecture while training parameter-matched readouts $e^{\theta}$ with and without GEs. For Weimann and Conrad, we follow their released PhysioNet Challenge 2021 diagnostic setting \citep{weimann2024federated}. 
Results are shown in Tables \ref{tab:ecgnormalization} and \ref{tab:ecgnormalization2}.


\begin{table*}[t]
\centering
\small
\begin{tabular}{llcc}
\toprule
Architecture
& Metric
& Matched control
& $+\log\rho$ GE \\
\midrule
\multirow{2}{*}{\citet{weimann2024federated}}
& AUROC
& $0.8646 \pm 0.0122$
& \textbf{$0.8792 \pm 0.0159$} \\
& AUPRC
& $0.6839 \pm 0.0283$
& \textbf{$0.7142 \pm 0.0301$} \\
\addlinespace
\multirow{2}{*}{ECGFounder \citep{li2025ecgfounder}}
& AUROC
& $0.8832 \pm 0.0039$
& \textbf{$0.9493 \pm 0.0040$} \\
& AUPRC
& $0.6869 \pm 0.0206$
& \textbf{$0.8310 \pm 0.0275$} \\
\bottomrule
\end{tabular}
\caption{}
\label{tab:ecgnormalization}
\end{table*}

\begin{table*}[t]
\centering
\small
\begin{tabular}{llcc}
\toprule
Architecture / task
& Metric
& Matched control
& $+\log\rho$ GE \\
\midrule
\multirow{2}{*}{\citet{weimann2024federated} native multilabel}
& AUROC
& $0.944762 \pm 0.001881$
& \textbf{$0.945808 \pm 0.002091$} \\
& AUPRC
& $0.608922 \pm 0.005248$
& $0.608099 \pm 0.006468$ \\
\addlinespace
\multirow{2}{*}{ECGFounder PTB-XL}
& AUROC
& $0.947557 \pm 0.008134$
& \textbf{$0.948464 \pm 0.008039$} \\
& AUPRC
& $0.644723 \pm 0.020212$
& \textbf{$0.646601 \pm 0.020199$} \\
\bottomrule
\end{tabular}
\caption{
}
\label{tab:ecgnormalization2}
\end{table*}



\subsubsection{Polarity in Time-Frequency Signal Processing}

The Choi-Williams distribution (CWD) is a classical quadratic time-frequency representation widely applied to nonstationary signal analysis, including radar recognition~\citep{zhang2016lpi}, machinery fault diagnosis~\citep{lee2001directional}, EEG~\cite{alazrai2017eeg}, and ECG~\citep{kayikcioglu2017timefrequency, kayikcioglu2020time} classification~\citep{choi1989improved, sejdic2009timefrequency}. Being quadratic, the CWD of a signal is invariant under sign-reversal, $\operatorname{CWD}(-x)=\operatorname{CWD}(x)$. Although such a property is often harmless, it is problematic for any task in which signs carry information. For example, a typical heartbeat is reflected in an ECG signal by a \emph{P wave}, followed by a \emph{QRS complex} (including the 'spike'), followed by a \emph{T wave}. The constant stretch between the end of the QRS complex and start of the T wave is the \emph{ST segment}. The time $R \subset \mathbb{R}$ between successive spikes is called an \emph{$R\text{-}R$ interval}. When the ST segment is not on the baseline, its sign relative to the baseline has physiological meaning, being used to help deduce what anomaly has occurred and where \citep{wagner2009aha}. For simplicity, we focus on this CWD-in-ECG learning setting, following in particular the papers \citep{kayikcioglu2017timefrequency} and \citep{kayikcioglu2020time}. 

Let $X$ consist of pairs $(r,i)$, where $r$ is a single-lead ($m=1$ in the previous ECG setting) ECG signal and $i$ indexes an $R$-$R$ interval $[R_{i}, R_{i+1}) \subset \mathbb{R}$. Let $\Gamma:=\operatorname{Aff}(1, \mathbb{R}) \cong \mathbb{R}^{\times} \ltimes \mathbb{R}$ act on $X$ as $\alpha_{\lambda,\delta}(r, i)=(\lambda r+ \delta \boldsymbol 1, i)$. Consider the subgroup $H:= \operatorname{Aff}^{+}(1, \mathbb{R}) \cong \mathbb{R}^{+} \ltimes \mathbb{R}$ of $\Gamma$. In the ECG setting, $H$ models common gain and DC offset changes; $Y$ is desired to be invariant under such nuisance transformations, giving the prescribed law
$$\mathrm{A} := \underline{H}_{\alpha}^{I} = \{ (\alpha_{\lambda,\delta} , I_{Y}): \lambda>0, \delta \in \mathbb{R} \}.$$
Attaining $\mathrm{A} \leq \operatorname{Sym}\mathscr{A}$ can be done by standardizing each $R$-$R$ interval as preprocessing. Indeed, for an interval $x_{i}(r):=r |_{[R_{i}, R_{i+1})}$, define $a(r,i):= \frac{x_{i}(r) - \texttt{mean}\big({x_{i}(r)}\big) \boldsymbol 1}{\,  \texttt{stdev}\big( x_{i}(r) \big)}$, and observe that $a\big( \alpha_{\lambda}, \delta (r,i)\big)=a(r,i)$ for all $(r,i) \in X$. Put $Z_{C_{1}}:=a(X)$. Following this preprocessing, the architecture $\mathscr{A}$ proposed in \citep{kayikcioglu2020time} applies CWD feature extractor $\Phi:Z_{C_{1}} \to Z_{C_{2}}=:\mathbb{R}^{2 \ell}$ given by $\Phi(u)=\big( E\big( \operatorname{CWD}(u) \big), E\big( \operatorname{CWD}(Du) \big) \big)$ where $Du$ is the first-difference of $u$ and $E$ extracts $\ell$-many pieces of summary information (e.g. min and max) from its input. Importantly, $\Phi(-u)=\Phi(u)$, invariance inherited from $\operatorname{CWD}$. These features are fed into a downstream classification model parameterizing functions $e^{\theta}:Z_{C_{2}} \to Y$. Thus $\mathscr{A}$ realizes functions that factor as
$$X \xrightarrow{ a}Z_{C_{1}} \xrightarrow{\Phi} Z_{C_{2}}\xrightarrow{e^{\theta}}Y.$$
Define an equivalence relation $\sim$ on $Z_{C_{1}}$ by $u \sim v \iff v \in \{ \pm u \}$. Note $\Phi$ is constant on $\sim$-classes because $\operatorname{CWD}$ is. Take $\sigma=\alpha_{-1,0}$. Then for every $(r,i) \in X$, $a \sigma(r,i)=a(-r, i)=-a(r,i)$, so that $a \sigma(r,i) \sim a(r,i)$. Lastly, $(\sigma, I_{Y})=(\alpha_{-1,0}, I_{Y}) \not \in \mathrm{A}$ since $-1 \not \in \mathbb{R}^{+}$. Hence the conditions of Definitoin \ref{def:over-quotienting} are satisfied and $\mathscr{A}$ over-quotients $\mathrm{A}$, with Proposition \ref{prop:over-quotienting} giving symmetry slippage as $(\alpha_{-1,0}, I_{Y}) \in \operatorname{Sym}\mathscr{A}-\mathrm{A}$.  

We would like to define GEs repairing this symmetry slippage. Let $Z=\mathbb{R}$. Given $x=(r,i)$, define
$$s_{i}(r):= \frac{r(t_{i}^{\text{ST}})- \texttt{mean}(r |_{B_{i}})}{\texttt{stdev}\big( x_{i}(r) \big) }$$
where $t_{i}^{ST}$ is the time at which the ST segment corresponding to heartbeat $i$'s height was measured, and $B_{i}$ is a short stretch of signal immediately before that heartbeat used to estimate the baseline. The idea is that $s_{i}(r)$ measures the ST height relative to the baseline, in units of the intervals standard deviation. We then define $c$ as the comparison of this quantity to a reference value computed from beats near the beginning of $r$, $s_{r}^{\text{ref}}:=\texttt{median}\{ s_{j}(r):j \text{ is a beat in the first 30s} \}$,
$$c(r, i):=s_{i}(r)-s_{r}^{\text{ref}}.$$
Thus $c(r,i)$ is positive when the ST segment of beat $i$ lies above the recording's reference level and negative when it lies below it. 
Given the slipped transformation $s=(\alpha_{-1,0},I_Y)$, take $\sigma=\alpha_{-1,0}$. Then, whenever $c(r,i)\neq 0$,
$$c\sigma(r,i)=c(-r,i)=-c(r,i)\neq c(r,i).$$

Putting $C:=C_{2}$, $\mathscr{A}$ realizes functions factoring as
$$
X \xrightarrow{b} Z_{C} \xrightarrow{e^{\theta}} Y.
$$
where $b=\Phi \circ a$. Appending $c$ at $C$, $\mathscr{A}'=\mathscr{A}\oplus_{C}c$ gives
\begin{align}
X &\xrightarrow{(b,c)} Z_{C}\times Z \xrightarrow{e^{\theta'}} Y,\\
(r,i) &\mapsto \big(\Phi(a(r,i)),c(r,i)\big)\mapsto \hat y,
\end{align}
where $e^{\theta'}$ is just the classifier of $\mathscr{A}$ with $c$ appended as a feature. 

To demonstrate the repair, we evaluate the CWD representation on the natural task of distinguishing ST elevation from ST depression in the European ST-T Database (EDB) \citep{taddei1992european}. We compare the architecture $\mathscr{A}$ proposed in \citep{kayikcioglu2020time} with and without GEs, with the non-GE version receiving instead the additional sign-invariant feature $|c|$. The cohort contains 68 subjects and 288 expert-annotated ST episodes. Following \citep{kayikcioglu2020time}, we train the same logistic and 5NN classifiers on both representations under leave-one-subject-out evaluation, taking 25 $R$-$R$ intervals from each episode and averaging their predicted probabilities to obtain an episode-level prediction. Results are shown in Table \ref{tab:cwd}.

\begin{table}[t]
\centering
\small
\begin{tabular}{lc}
\toprule
Architecture & AUROC \\
\midrule
Baseline Logistic Regression \citep{kayikcioglu2020time}
& $0.8505$ \\
Logistic Regression + GE
& \textbf{$0.9671$} \\
Baseline 5-NN \citep{kayikcioglu2020time}
& $0.8157$ \\
5-NN + GE
& \textbf{$0.9427$} \\
\bottomrule
\end{tabular}
\caption{}
\label{tab:cwd}
\end{table}

\subsection{Extension-Mistyping Case Studies}

\subsubsection{(Pseudo)scalar Mistyping in Neural Operators}

The action on $\mathbb{R}^{2}$ of the dihedral group $G:=D_{4}$ induces two natural actions on functions $\omega:\mathbb{R}^{2}\to \mathbb{R}^{}$. The first action is $g \omega(x):=\omega(g ^{-1} x)$, in which case $\omega$ is called a \emph{scalar (field)}. The second action is $g\omega(x):=\det(g)\omega(g ^{-1} x)$, in which case $\omega$ is called a \emph{pseudoscalar (field)}. The two actions, which we denote $\widetilde{\alpha}^{\text{r}}, \widetilde{\alpha}^{\text{p}}:G \curvearrowright \mathbb{R}^{\mathbb{R}^{2}}$, restrict to the same action $\alpha:C_{4}=:H \curvearrowright \mathbb{R}^{\mathbb{R}^{2}}$, and so the distinction is not relevant when the only actions of interest are by rotations. It becomes relevant under reflections, however. For instance, two-dimensional vorticity (i.e., $\operatorname{curl}$) is a pseudoscalar.

\citet{helwig2023group} introduce the \textit{Group Equivariant Fourier Neural Operator (G-FNO)}, extending the Fourier neural operator of \citep{LiEtAl2021} to impose equivariance to discrete rotations and reflections. They introduce two architectures, G-FNO-p4 and G-FNO-p4m. The first is $H=C_{4}$-equivariant, while the second extends the action to $G=D_{4}$. The architectures have the form
$$
X\xrightarrow{P_G}Z_0\xrightarrow{L_1}\cdots\xrightarrow{L_L}Z_L\xrightarrow{Q_G}Y
$$
where the $L_\ell$ are group Fourier layers, $Z_{\ell} \cong (\mathbb{R}^{n_{\ell}} \otimes \mathbb{R}^{K})^{\mathbb{R}^{2}}$ carries the representation $\lambda_{\ell}$ on functions $\mathbb{R}^{2} \to \mathbb{R}^{n_{\ell}}\otimes \mathbb{R}^{K}$ induced by the tensor product of the trivial representation $K \curvearrowright \mathbb{R}^{n_{\ell}}$ and the regular representation on $\mathbb{R}^{K}$. Explicitly, $(\rho_{\ell}(k)z)(x)=(I_{n_{\ell}} \otimes \operatorname{Reg}_{K}(k))z(k ^{-1} x)$. G-FNO-p4m, however, extends the action from $\alpha$ to $\widetilde{\alpha}^{\text{r}}$ on vorticity rather than $\widetilde{\alpha}^{\text{p}}$, thereby typing vorticity as an ordinary scalar rather than a pseudoscalar.

Let us verify this is extension-mistyping. Let $\mathscr A$ denote G-FNO-p4m, let $\lambda_\ell:G\curvearrowright Z_\ell$ denote the $D_4$-action on each $Z_{\ell}$ described above, and let $\boldsymbol\rho=(\rho_v:H\curvearrowright Z_v)_{v\in V}$ denote the $H$-actions obtained by restricting the $G$-actions realized by the architecture; in particular, $\rho_\ell=\lambda_\ell|_H$ on the $Z_{\ell}$ for $\ell =1,\dots,L$ and $\rho_I=\rho_O=\alpha$ on $X$ and $Y$. Let $\widetilde{\boldsymbol\rho}^{\mathrm p}$ be the prescribed extension of $\boldsymbol\rho$ to $G=D_4$: on $X$ and $Y$ it is the pseudoscalar action $\widetilde\alpha^{\mathrm p}$, while on $Z_0,\ldots,Z_L$ it is $\lambda_\ell$. Choose a reflection $s\in G-H$, and define $S_I=\widetilde\alpha^{\mathrm r}(s)$, $S_\ell=\lambda_\ell(s)$, and $S_O=\widetilde\alpha^{\mathrm r}(s)$. Since these operators arise from $D_4$-actions, they satisfy conditions (i) and (ii) of Proposition~\ref{prop:extension-mistyping-special}, and since each map in G-FNO-p4m is equivariant for these actions, condition (iii) is also satisfied. On the vorticity carriers, however, $\widetilde\rho_I^{\mathrm p}(s)=-\widetilde\alpha^{\mathrm r}(s)=-S_I$ and $\widetilde\rho_O^{\mathrm p}(s)=-\widetilde\alpha^{\mathrm r}(s)=-S_O$, so $\widetilde\rho_v^{\mathrm p}(s)\neq S_v$ for $v=I,O$. Since $H\triangleleft G$ and $[G:H]=2$, Proposition~\ref{prop:extension-mistyping-special} therefore gives an extension $\widetilde{\boldsymbol\rho}^{\mathrm r}$ of $\boldsymbol\rho$ for which G-FNO-p4m extension-mistypes $\widetilde{\boldsymbol\rho}^{\mathrm p}$.

To see this extension-mistyping creates symmetry slippage, take $\mathrm A:=\underline{D_4}_{\widetilde\alpha_X^{\mathrm p}}^{\widetilde\alpha_Y^{\mathrm p}}$. In G-FNO-p4m, however, $P_G$ is equivariant with respect to $\widetilde\alpha_X^{\mathrm r}$ and $\lambda_0$, each $L_\ell$ is equivariant with respect to $\lambda_{\ell-1}$ and $\lambda_\ell$, and $Q_G$ is equivariant with respect to $\lambda_L$ and the $\widetilde\alpha_Y^{\mathrm r}$. Proposition~\ref{prop:extension-mistyping} therefore gives $\underline{D_4}_{\widetilde\alpha_X^{\mathrm r}}^{\widetilde\alpha_Y^{\mathrm r}}\leq\operatorname{Sym}\mathscr A$. For every reflection $g\in D_4-C_4$, one has $\widetilde\alpha_X^{\mathrm p}(g)=-\widetilde\alpha_X^{\mathrm r}(g)$ and $\widetilde\alpha_Y^{\mathrm p}(g)=-\widetilde\alpha_Y^{\mathrm r}(g)$, so the realized transformation $s_g:=\bigl(\widetilde\alpha_X^{\mathrm r}(g),\widetilde\alpha_Y^{\mathrm r}(g)\bigr)$ is not contained in the prescribed law $\mathrm A$. Thus $S:=\{s_g:g\in D_4-C_4\}\subset\operatorname{Sym}\mathscr A-\mathrm A$, and G-FNO-p4m exhibits symmetry slippage by extension-mistyping.

To repair, put $Z:=(\mathbb{R}^{G})^{\mathbb{R}^{2}}$ and define $c:X\to Z$ by $c(x)_{g}=\det(g)x$ for $g\in G$ (applied channelwise). Let $\kappa^{\mathrm r}:G\curvearrowright Z$ be the restriction of $\lambda_0$ to one channel of $Z_0$, and put $\kappa^{\mathrm p}(k):=\det(k)\kappa^{\mathrm r}(k)$. Then $c(\widetilde\alpha_X^{\mathrm r}(k)x)=\kappa^{\mathrm p}(k)c(x)$. Since $\det(h)=1$ for $h\in H=C_4$, we have $\kappa^{\mathrm p}|_H=\kappa^{\mathrm r}|_H$, while for every reflection $s\in G-H$, $\kappa^{\mathrm p}(s)=-\kappa^{\mathrm r}(s)$. Putting $C:=Z_0$, $\mathscr A$ realizes functions factoring as $X\xrightarrow{b}Z_C\xrightarrow{e^\theta}Y$, where $b=P_G$ and $e^\theta=Q_G\circ L_L\circ\cdots\circ L_1$. Appending $c$ at $C$, $\mathscr A'=\mathscr A\oplus_Cc$, gives

\begin{align}
X &\xrightarrow{(b,c)} Z_C\times Z\xrightarrow{e^{\theta'}}Y,\\
x &\longmapsto \bigl(P_Gx,c(x)\bigr)\longmapsto\hat y,
\end{align}

where $e^{\theta'}$ is obtained in the usual way (Section~\ref{sec:GEs}).

Since $c \not \equiv 0$, to apply Corollary~\ref{cor:ge-em-app-cor} and thus verify that $c$ is a GE it only remains check that $\mathscr{A}_{(C,O)}^{\prime\top}$ is injective on pairs of representations witnessing $\kappa^{\mathrm p}(s)c(x)\neq\kappa^{\mathrm r}(s)c(x)$. One example to see this is as follows. Choose any frequency $\xi$ whose stabilizer under $G$ is trivial, for example $\xi=(1,2)$, and define $\psi$ such that the only nonzero Fourier coefficient of $\psi$ is $\xi$. Since the stabilizer of $\xi$ is trivial, $\psi_-:=\sum_{g\in G}\det(g)g\psi \neq 0$. A $G$-Fourier convolution followed by $Q_G$ gives an $e$ satisfying

\begin{align}
e(0,\kappa^{\mathrm p}(s)c(x))
&=-e(0,\kappa^{\mathrm r}(s)c(x))
\neq e(0,\kappa^{\mathrm r}(s)c(x)),
\end{align}

so the condition indeed holds.

To demonstrate the repair, we compare G-FNO-p4m with and without the GE on two vorticity-forecasting tasks. The first is the NS-SYM benchmark of \citep{helwig2023group}, governed by $\partial_t\omega+u\cdot\nabla\omega=\nu\Delta\omega+f$ with $\nu=10^{-4}$ and $f=0.1(\cos 4\pi x_1+\cos 4\pi x_2)$; we refer the reader to \citep{helwig2023group} for details. Results are shown in Table~\ref{tab:GFNO-ns-sym} We also experiment on an unforced decaying-turbulence benchmark~\citep{cao2025spectralrefiner}. We use random peaked-spectrum initial conditions following the classical decaying-turbulence setup of~\cite{mcwilliams1984emergence} We generate the trajectories using a pseudospectral vorticity solver on the periodic domain $[0,2\pi]^2$, following the modern \texttt{torch-cfd} implementation~\citep{cao2025spectralrefiner}. We generate 768 examples at $64\times64$ resolution with peak wavenumber $4$ and evolve them with viscosity $\nu=10^{-3}$, time step $\Delta t=10^{-3}$, and $T=0.5$. The task is to approximate the finite-time solution operator
$\omega(\_,0)\mapsto\omega(\_,T)$. Given an initial vorticity
field, the model predicts the full vorticity field after evolution to
$T=0.5$. For each sampled initial condition $\omega_0$, we also include
the sign-reversed initial condition $-\omega_0$ and evolve the two
initial conditions separately. The idea is that the data contain pairs
$\omega_0$, $-\omega_0$, which we see causes symmetry slippage to become especially apparent empirically. Performance is measured by NMSE on held-out
final-time vorticity fields. Results are shown in Table~\ref{tab:GfNO-mcwilliams}.

\begin{table}[t]
\centering
\small
\begin{tabular}{lc}
\toprule
Architecture & Rel. MSE (\%) \\
\midrule
G-FNO-p4m Baseline \citep{helwig2023group}
& $2.363 \pm 0.240$ \\
G-FNO-p4m + GE
& $1.701 \pm 0.173$ \\
G-FNO-p4m Retyped
& $1.302 \pm 0.130$ \\
\bottomrule
\end{tabular}
\caption{}
\label{tab:GFNO-ns-sym}
\end{table}

\begin{table}[t]
\centering
\small
\begin{tabular}{lc}
\toprule
Architecture & NMSE \\
\midrule
G-FNO-p4m Baseline \citep{helwig2023group}
& $0.294442 \pm 0.005787$ \\
G-FNO-p4m + GE
& $0.008005 \pm 0.000336$ \\
G-FNO-p4m Retyped
& $0.005520 \pm 0.000207$ \\
\bottomrule
\end{tabular}
\caption{}
\label{tab:GfNO-mcwilliams}
\end{table}

\subsubsection{Tetrahedral Chirality in Molecular Graph Learning}

Chiral molecules can occur in two mirror-image forms, called \emph{enantiomers}, that have the same atoms/bonds but different handedness. A common source of chirality is a tetrahedral atom bonded to four distinct ligands, whose spatial arrangement can occur in two opposite configurations. While chemists conventionally describe these configurations using the Cahn-Ingold-Prelog (CIP) priority rules \citep{cahn1966specification}, which assign the four ligands fixed priorities and label the resulting configuration $R$ or $S$, the ligand ordering is in general arbitrary and $G:=S_4$ acts by permuting it. Even permutations $H:=A_4$ preserve handedness and odd permutations reverse it. Any representation of tetrahedral stereochemistry defined relative to an ordering of the four ligands should therefore be equivariant to the sign representation $\pi\mapsto\operatorname{sign}(\pi)\in\mathbb R$ of $S_4$~\citep{pattanaik2020message}.

For a concrete example, consider the architecture $\mathscr A$ given by QGeoGNN~\citep{xu2023retention}, a graph neural network for chromatographic retention-time prediction. QGeoGNN represents molecular chirality using an atomic \emph{chiral tag} appended to ordinary atom features and applies a graph isomorphism network (GIN) \citep{xu2018how}. Encoding the two tetrahedral tag values at a stereocenter $v$ by $\tau_v\in\{\pm1\}$, the prescribed action is $\widetilde\rho_v^{\mathrm p}(\pi)\tau_v=\operatorname{sign}(\pi)\tau_v$: after permuting the four ligand positions by $\pi\in S_4$, the tag should flip iff $\pi$ is odd in order to represent the same stereochemical configuration. QGeoGNN instead treats the tag as it does other node features, giving the realized action $\widetilde\rho_v^{\mathrm r}(\pi)\tau_v=\tau_v$. Thus $\widetilde\rho_v^{\mathrm p}|_{A_4}=\widetilde\rho_v^{\mathrm r}|_{A_4}$, while $\widetilde\rho_v^{\mathrm p}(s)=-\widetilde\rho_v^{\mathrm r}(s)$ for every $s\in S_4-A_4$.

For every node map $\phi_v^\theta$ in the GIN, $\phi_v^\theta\widetilde\rho_{\{u\to v\}}^{\mathrm r}(\pi)=\widetilde\rho_v^{\mathrm r}(\pi)\phi_v^\theta$ for $\pi\in S_4$, while $\widetilde\rho_v^{\mathrm p}(s)\neq\widetilde\rho_v^{\mathrm r}(s)$ on the chiral-tag carrier for $s\in S_4-A_4$. Hence QGeoGNN extension-mistypes $\widetilde{\boldsymbol\rho}^{\mathrm p}$. Write $\widetilde\rho_I^{\mathrm p},\widetilde\rho_I^{\mathrm r}:S_4\curvearrowright X$ for the corresponding actions on molecular representations: both permute the four ligand positions by $\pi$, while $\widetilde\rho_I^{\mathrm p}$ sends $\tau_i\mapsto\operatorname{sign}(\pi)\tau_i$ and $\widetilde\rho_I^{\mathrm r}$ sends $\tau_i\mapsto\tau_i$. Taking $\mathrm A:=\underline{S_4}_{\widetilde\rho_I^{\mathrm p}}^{1}$, every $f^\theta\in\mathcal F_{\mathscr A}$ satisfies $f^\theta\widetilde\rho_I^{\mathrm r}(\pi)=f^\theta$, so Proposition~\ref{prop:extension-mistyping} gives $\underline{S_4}_{\widetilde\rho_I^{\mathrm r}}^{1}\leq\operatorname{Sym}\mathscr A$. Since $\widetilde\rho_I^{\mathrm r}(s)\neq\widetilde\rho_I^{\mathrm p}(s)$ for every $s\in S_4-A_4$, $(\widetilde\rho_I^{\mathrm r}(s),1)\notin\mathrm A$, and QGeoGNN exhibits symmetry slippage.

This symmetry slippage can pose a problem when enantiomers have different targets, as can occur in chromatographic enantioseparation. To repair, let $\chi_i(x)\in\{R,S,\varnothing\}$ denote the CIP configuration at atom $i$. Put $Z=(\mathbb R^4)^V$ and define $c:X\to Z$ by $c_i(x)=\left(\mathbf 1_{\{\chi_i(x)=R\}},\mathbf 1_{\{\chi_i(x)=S\}},\mathbf 1_{\{\tau_i(x)=+1\}},\mathbf 1_{\{\tau_i(x)=-1\}}\right)$. Let $P_{RS}$ exchange the first two coordinates of $\mathbb R^4$ and fix the latter two, and define $\kappa^{\mathrm r},\kappa^{\mathrm p}:S_4\curvearrowright\mathbb R^4$ by $\kappa^{\mathrm r}(\pi)=I_4$ and $\kappa^{\mathrm p}(\pi)=I_4$ for $\pi\in A_4$, while $\kappa^{\mathrm p}(\pi)=P_{RS}$ for $\pi\in S_4-A_4$. It is readily checked that $\operatorname{im}c\not\subset\operatorname{Eq}(\kappa^{\mathrm p}(s),\kappa^{\mathrm r}(s))$.\footnote{Specifically: $\tau_i(\widetilde\rho_I^{\mathrm r}(\pi)x)=\tau_i(x)$, while $\chi_i(\widetilde\rho_I^{\mathrm r}(\pi)x)=\chi_i(x)$ for $\pi\in A_4$ and $\chi_i(\widetilde\rho_I^{\mathrm r}(\pi)x)=\overline{\chi_i(x)}$ for $\pi\in S_4-A_4$, where $\overline R=S$ and $\overline S=R$. Hence $c(\widetilde\rho_I^{\mathrm r}(\pi)x)=\kappa^{\mathrm p}(\pi)c(x)$. Moreover, $\kappa^{\mathrm p}|_{A_4}=\kappa^{\mathrm r}|_{A_4}$, while for every $s\in S_4-A_4$, $\operatorname{Eq}(\kappa^{\mathrm p}(s),\kappa^{\mathrm r}(s))=\{(a,a,b,d):a,b,d\in\mathbb R\}$. Since at an assigned stereocenter the first two coordinates of $c_i(x)$ are either $(1,0)$ or $(0,1)$, so $\operatorname{im}c\not\subset\operatorname{Eq}(\kappa^{\mathrm p}(s),\kappa^{\mathrm r}(s))$.}

Let $a_i(x)\in\mathbb R^d$ denote the ordinary atom features at atom $i$ immediately before the first GIN layer, so that the input to the GIN is $b(x)=((a_i(x),\tau_i(x)))_{i\in V}\in Z_C$. Thus $\mathscr A$ factors as $X\xrightarrow{b}Z_C\xrightarrow{e^\theta}Y$. Appending $c$ at this cut gives $\mathscr A'=\mathscr A\oplus_Cc$ with $X\xrightarrow{(b,c)}Z_C\times Z\xrightarrow{e^{\theta'}}Y$ and $(b,c)(x)=((a_i(x),\tau_i(x),c_i(x)))_{i\in V}$. Since $\operatorname{im}c\not\subset\operatorname{Eq}(\kappa^{\mathrm p}(s),\kappa^{\mathrm r}(s))$ for every $s\in S_4-A_4$, together with the injectivity condition of Proposition~\ref{prop:GE-EM-app}, which we verified for this type of MPNN in Section~\ref{app:over-quotienting-full-case-studies} this gives that $c$ is a GE.

We demostate the repair on QGeoGNN's all-column retention-time benchmark \citep{xu2023retention} with a fixed $90/5/5$ IID split containing $23{,}020$ training, $1{,}278$ validation, and $1{,}278$ test molecules. We evaluate the model with and without GEs, comparing as well to a 'retyped' model that just replaces $\tau_i$ at assigned tetrahedral stereocenters by $\chi_i\in\{R,S\}$. Results are in Table~\ref{tab:qgeognn}. 

\begin{table}[t]
\centering
\small
\begin{tabular}{lc}
\toprule
Architecture &  RMSE \\
\midrule
Baseline \citep{xu2023retention}
& $5.3713 \pm 0.0971$ \\
Retyped
& \textbf{$5.1707 \pm 0.1258$} \\
+ GE
& $5.2268 \pm 0.0454$ \\
\bottomrule
\end{tabular}
\caption{}
\label{tab:qgeognn}
\end{table}

\subsubsection{Pseudovector Mistyping in Magnetic Equivariant ML}

Magnetic moments and spins are pseudovectors $w$, acted on by $G:=\operatorname{O}(3)$ as $\widetilde{\rho}^{\text{p}}(Q)w:=\det(Q) Qw$. This is in contrast to (ordinary) vectors, which are of course acted on as $\eta(Q)w= Qw$ with no determinant factor. The distinction is not relevant when only $\operatorname{SO}(3) :=H \triangleleft G$ is acting as $\rho=\widetilde{\rho}^{p} |_{H}=\eta |_{H}$, but it is relevant to equivariant learning for magnetic materials, where spin and magnetic-moment pseudovectors are often represented alongside ordinary vectors such as atomic positions and displacements~\citep{yuan2024magnet,yu2024spingnnpp,cho2026sparsitypromoting}. In the atomistic equivariant learning literature \citep{geiger2022e3nn}, $\widetilde{\rho}^{\text{p}}$ is denoted $1e$ and $\eta$ is denoted $1o$ respectively. More generally, the notation $V_{\ell,p}=\ell p$ denotes the $G$-irrep obtained by extending the unique real degree-$\ell$ $H$-irrep \citep{hitchin2015liegroups} to $G$ with $p \in \{ e,o \}$ recording the action of $G-H$. 

Suppose there is a node $v$ in $\mathscr{A}$ whose representation space $Z_{v}$ consists of spin or magnetic moment features and comes equipped with the standard action $\eta_{v}(Q)z=Qz$. With $s:= -I \in G-H$, $S_{v}:=-I_{Z_{v}}$, one has $\widetilde{\rho}_{v}^{\text{p}}(s) \neq S_{v}$. Therefore, since $[G:H]=2$, extension-mistyping occurs if $S_{v}$ can be extended to a family $\boldsymbol S=(S_{u})_{u \in V}$ of linear maps satisfying hypotheses (i)-(iii) of Proposition~\ref{prop:extension-mistyping-special}. For a straightforward example, let $\mathscr{A}$ denote the spin-aware augmentation of MACE \citep{batatia2022mace} proposed in \citep{cho2026sparsitypromoting}. The authors propose a framework for augmenting equivariant materials foundation models, and instantiate the framework by augmenting MACE with additional inputs given by spin directions $s_{i} \in Z_{v} \cong \mathbb{R}^{3}$ where $Z_{v}$ is equipped with the action $\eta_{v}=1o$. With $X=\{ (\zeta_{i}, r_{i}, s_{i}, \mathcal{G}_{i}) \}$ denoting tuples of atomic number, position, and spin direction and atomistic graphs $\mathcal{G}_{i}=(\mathcal{V}_{i}, \mathcal{E}_{i})$ for which they are node features, the architecture $\mathscr{A}$ then has computation graph

\[
\begin{tikzcd}
                                              & Z_v=Z_{u_0}^{\text{spin}} \arrow[rd] &                   &                   &                  &         \\
{\{(\zeta_i, r_i, s_i)\}=X} \arrow[ru] \arrow[rd] &                                      & Z_{u_1} \arrow[r] & Z_{u_2} \arrow[r] & \cdots \arrow[r] & Z_{u_k} \\
                                              & Z_{u_0}^{\text{MACE}} \arrow[ru]     &                   &                   &                  &
\end{tikzcd}
\]

where $Z_{u_{i}}=\bigoplus_{ \ell ,p} m_{, \ell_{i} ,p_{i}} V_{\ell, p}$ for some multiplicities $m_{u_{i}, \ell, p}$ \citep{batatia2022mace}. Let $S_{\ell, p}:V_{\ell, p} \to V_{\ell, p}$ be defined as $I$ if $p=e$ and $-I$ if $p=o$. Define $\boldsymbol S=S_{v} \cup (S_{u_{i}})_{i=1}^{k}$ via $S_{u_{i}}=\bigoplus_{\ell_{i}, p_{i}} S_{\ell_{i}, p_{i}}$. With $\mathscr{A}$ parameterizing functions factoring as $X \xrightarrow{a}Z_{C_{1}} \xrightarrow{\Phi^{\theta}}Z_{C_{2}} \xrightarrow{e^{\theta}} Y$ where $Z_{C_{1}}=Z_{v}\oplus Z_{u_{0}}^{\text{MACE}}$ and $Z_{C_{2}}=Z_{u_{2}}$, one has $S_{v} \in \boldsymbol S$ and it is readily checked that $\boldsymbol S$ satisfies the conditions (i)-(iii) of Proposition~\ref{prop:extension-mistyping-special}; we conclude that extension-mistyping occurs on $\mathscr{A}_{(C_{1},C_{2})}$. For another example, let $\mathscr A$ denote MagNet, the $E(3)$-equivariant force-field architecture for magnetic materials proposed in \citep{yuan2024magnet}. MagNet first processes atomic structure through a sequence of atomic interaction blocks, and then combines the resulting atomic features with additional inputs given by magnetic-moment directions $s_i\in Z_v\cong\mathbb R^3$, where $Z_v$ is again equipped with the action $\eta_v=1o$. The resulting features are processed through two magnetic interaction blocks. The architecture $\mathscr A$ therefore has computation graph of the form

\[
\begin{tikzcd}
                                              & Z_v=Z_{u_0}^{\mathrm{mag}} \arrow[rd] &                   &                   &                  &         \\
{\{(\zeta_i,r_i,s_i)\}=X} \arrow[ru] \arrow[rd]  &                                      & Z_{u_1} \arrow[r] & Z_{u_2} \arrow[r] & \cdots \arrow[r] & Y \\
                                              & Z_{u_0}^{\mathrm{atom}} \arrow[ru]   &                   &                   &                  &
\end{tikzcd}
\]

where $Z_{u_0}^{\mathrm{atom}}$ is the output of the atomic interaction stage and $Z_{u_1},Z_{u_2}$ are the outputs of the two magnetic interaction blocks, where $Z_{u_i}=\bigoplus_{\ell,p}m_{u_i,\ell,p}V_{\ell,p}$ for some multiplicities $m_{u_i,\ell,p}$. For the purposes of our experiments below, we augment MagNet with an equivariant readout that first forms edge features $v_{ij}\in V_{1,o}$ and then combines each $v_{ij}$ with the $r_{ij}\in V_{1,o}$ through a $1o\otimes1o\to1e$ tensor product, and regard this as the architecture $\mathscr{A}$. Let $S_{\ell,p}:V_{\ell,p}\to V_{\ell,p}$ be defined as before, and define $\boldsymbol S$ likewise as before. With $\mathscr A$ parameterizing functions factoring as $X\xrightarrow{a}Z_{C_1}\xrightarrow{\Phi^\theta}Z_{C_2}\xrightarrow{e^\theta}Y$, where $Z_{C_1}=Z_v\oplus Z_{u_0}^{\mathrm{atom}}$ and $Z_{C_2}=Z_{u_2}$, one has $S_v\in\boldsymbol S$, and it is readily checked that $\boldsymbol S$ satisfies conditions (i)-(iii) of Proposition~\ref{prop:extension-mistyping-special}; we conclude that extension-mistyping occurs on $\mathscr A_{(C_1,C_2)}$.Moreover, since $a$ and $e^{\theta}$ are both equivariant in both cases above, Proposition~\ref{prop:extension-mistyping} applies to give that this extension-mistyping creates symmetry slippage.

While this symmetry slippage does not pose an expressivity problem for tasks invariant under spin reversal (e.g. zero-field scalar energy prediction), it can be problematic otherwise. We define GEs for each aforementioned architecture in turn. Let $Z= \bigoplus_{i \neq j}\mathbb{R}^{3}$, and write $r_{ij}:=r_{i}-r_{j}$. For the architecture $\mathscr{A}$ of \citep{cho2026sparsitypromoting}, define $c:X \to Z$ by
$$
c(x)_{ij}= \begin{cases}
 (r_{ij} \times s_{i}) \times r_{ij} & (i, j) \in \mathcal{E}_{x} \\
 0 & (i, j) \not \in \mathcal{E}_{x}.
\end{cases}
$$
Note $c\big( \alpha_{X}^{\text{r}}(Q)x \big)_{ij}=Q c(x)_{ij}$. Now define actions $\kappa^{\text{p}},\kappa^{\text{r}}:G \curvearrowright Z$ as $\kappa^{\text{p}}(Q):= \bigoplus_{i \neq j}\eta(Q)$ and $\kappa^{\text{r}}(Q):= \bigoplus_{i \neq j} \widetilde{\rho}^{\text{p}}(Q)$. Certainly $c \alpha_{X}^{\text{r}}(Q)=\kappa^{\text{p}}(Q) \circ c$. Putting $C:=C_{2}$, $\mathscr{A}$ realizes functions as $X \xrightarrow{b}Z_{C} \xrightarrow{e^{\theta}}Y$; appending $c$ at $C$, $\mathscr{A}'=\mathscr{A}\oplus_{C}c$ gives
$$
X \xrightarrow{ (b, c)}Z_{C} \times Z \xrightarrow{e^{\theta'}}Y,
$$
where $e^{\theta'}$ is defined as with the $\mathscr{A}_{(C,O)}=\operatorname{MLP}$ example in Section \ref{sec:GEs}. For the MagNet architecture $\mathscr{A}$ of \citep{yuan2024magnet}, instead define $c:X\to Z$ by
$$
c(x)_{ij}=
\begin{cases}
r_{ij}\times s_i &(i,j)\in\mathcal E_x,\\
0 &(i,j)\notin\mathcal E_x.
\end{cases}
$$
Then $c\bigl(\alpha_X^{\mathrm r}(Q)x\bigr)_{ij}=\det(Q)Q\,c(x)_{ij}$. Define $\kappa^{\mathrm p}(Q):=\bigoplus_{i\neq j}\widetilde\rho^{\mathrm p}(Q)$ and $\kappa^{\mathrm r}(Q):=\bigoplus_{i\neq j}\eta(Q)$. Hence $c\alpha_X^{\mathrm r}(Q)=\kappa^{\mathrm p}(Q)c$. Putting $C$ immediately before our appended readout, $\mathscr A$ realizes functions as $X\xrightarrow{b}Z_C\xrightarrow{e^\theta}Y$; appending $c$ at $C$, $\mathscr A'=\mathscr A\oplus_Cc$ gives $X\xrightarrow{(b,c)}Z_C\times Z\xrightarrow{e^{\theta'}}Y$, with $e^{\theta'}$ defined as above.

In both constructions, $\kappa^{\mathrm p}|_H=\kappa^{\mathrm r}|_H$ since $\eta|_H=\widetilde\rho^{\mathrm p}|_H$. Moreover, for $s=-I\in G-H$, $\kappa^{\mathrm p}(s)=-\kappa^{\mathrm r}(s)$. Indeed, for the first architecture one has $\kappa^{\mathrm p}(s)=-I_Z$ and $\kappa^{\mathrm r}(s)=I_Z$, while for the second these are reversed. Since in each case $c\not\equiv0$, and $\mathscr{A}_{(C,O)}$ satisfies the injectivity condition of Proposition~\ref{prop:GE-EM-app}, Corollary~\ref{cor:ge-em-app-cor} gives that both $c$'s are GEs for their respective architectures. 

To demonstrate the repair, we compare the two architectures architectures with and without GEs on the noncollinear anomalous-Hall prediction task of \citep{kipp2024hall}. Each input is a magnetic honeycomb configuration specified by two sublattice spin directions $s_A,s_B$, and the target is the symmetrized noncollinear Hall response $\sigma^{\mathrm{nc}}(s_A,s_B)=\frac{1}{2}\big(\sigma(s_A,s_B)+\sigma(s_B,s_A)\big)$. For each architecture, we compare the GE against a parameter-matched control appended at the same cut and processed by the same downstream architecture: for \citep{cho2026sparsitypromoting} we compare $c_{\mathrm{GE},ij}=(r_{ij}\times s_i)\times r_{ij}$ against $c_{\mathrm{ctl},ij}=r_{ij}\times s_i$, while for MagNet \citep{yuan2024magnet} we compare $c_{\mathrm{GE},ij}=r_{ij}\times s_i$ against $c_{\mathrm{ctl},ij}=(r_{ij}\times s_i)\times r_{ij}$. Results are shown in Table~\ref{tab:magnets}.


\begin{table}[t]
\centering
\small
\begin{tabular}{lccc}
\toprule
Architecture & Baseline RMSE & GE RMSE & GE $R^2$ \\
\midrule
\citep{cho2026sparsitypromoting}
& $0.071233 \pm 0.000208$
& $0.030384 \pm 0.002293$
& $0.8151 \pm 0.0284$ \\
{MagNet} \citep{yuan2024magnet}
& $0.070866 \pm 0.000002$
& $0.032227 \pm 0.001077$
& $0.7930 \pm 0.0140$ \\
\bottomrule
\end{tabular}
\caption{}
\label{tab:magnets}
\end{table}

\section{Experimental Details} 
\label{app:experimental-details}

We provide specific details to aid in reproducing the experiments described in Sections~\ref{sec:experiments} and \ref{app:case-studies}. All training was done on an NVIDIA A100 GPU. In each case study, we report mean and standard deviation over three seeds of the evaluation metric(s) used by the papers proposing the architectures we analyze, unless stated otherwise. 
\paragraph{Experimental Details for Case Study~\ref{case:lorentz}}

We use \(64/16/24\) base geometries for training/validation/ID test, respectively, together with a time-orientation-reversed version of every geometry, and an independent \(24\)-base-geometry OOD test set. Each contains \(n=64\) points sampled independently with
\(t\sim\operatorname{Unif}[-2,2]\) and \(x,y\sim\operatorname{Unif}[-1,1]\), and four source fields, each generated as a sum of \(4\)-\(7\) anisotropic Gaussian bumps with amplitudes \(a\sim\mathcal N(0,1)\), centers \(t_c\sim\operatorname{Unif}[-1.5,1.5]\), \(x_c,y_c\sim\operatorname{Unif}[-0.8,0.8]\), widths \(\sigma_t\sim\operatorname{Unif}[0.22,0.55]\), \(\sigma_x,\sigma_y\sim\operatorname{Unif}[0.18,0.45]\), and normalized to unit standard deviation. We use uniform quadrature weights \(w_j=16/64=0.25\) and \(q_{\min}=0.005\). ID geometries are transformed by a boost with direction \(\theta\sim\operatorname{Unif}[0,2\pi)\) and rapidity \(\eta\sim\operatorname{Unif}[-2,2]\); OOD uses the same random directions with an equiprobable rapidity sign and \(|\eta|\sim\operatorname{Unif}[4,4.5]\). We use the released LorentzNet, PELICAN, LGN, and L-GATr models, modified for the task as described in Appendix~\ref{app:case-studies} The control uses an identically zero scalar slot wherever the GE is supplied in the repaired arm, giving equal parameter counts. 

\paragraph{Experimental Details for Case Study~\ref{case:units}}
We use the weekly Mauna Loa $\mathrm{CO}_2$ series from \texttt{statsmodels.datasets.co2} (March 1958-December 2001 \cite{KeelingWhorf2004}). Missing weekly levels are linearly interpolated before forming first differences and the 12-week input windows. Because the literal Buckingham-$\Pi$ features $\Pi_j(h)=h_j/h_0$ are undefined when $h_0=0$, we discard such windows, leaving $2100$ examples. The resulting chronological train/validation/test split contains $1244/434/422$ examples. We repeat the experiment with a $16$-sample purge after each split boundary, so that no weekly observation is shared across train, validation, and test windows; the purged test set contains $406$ examples. All arms use full-batch Adam with learning rate $3\times 10^{-3}$, weight decay $10^{-4}$, at most $2000$ epochs.

\paragraph{Experimental Details for Case Study~\ref{case:intaut}}
Following the Treloar protocol of Holthusen et al.~\citep{holthusen2026treloar}, we construct incompressible deformation gradients from the reported stretches and fit jointly to the 24 uniaxial-tension (UT) and 16 biaxial-tension (BT) stress measurements, reserving pure shear for extrapolation. We parameterize the input by the true strain $\varepsilon=\frac12\log(F^\top F)$, predict a strain energy $W_\theta(\varepsilon)$, and obtain the corresponding nominal stresses by automatic differentiation through the deformation parametrization. We optimize $\mathrm{MSE}_{\mathrm{UT}}+3\,\mathrm{MSE}_{\mathrm{BT}}$.

\paragraph{Experimental Details for Case Study~\ref{case:glycan}}

For all three architectures, we use the same five outer stratified group folds over the 13,209 records, grouping records by exact rootless attributed graph. In each outer fold, the held-out group is used only for testing, while a second grouped stratified split of the remaining records supplies validation data for checkpoint selection. We otherwise retain the released domain-classification configurations and architecture-specific hierarchy; the common training configuration uses three 128-dimensional message-passing layers, dual readout, batch size 256, 50 epochs, Adam with learning rate $5\times 10^{-4}$ and weight decay $10^{-3}$, and unweighted multiclass cross-entropy. RGCN and CompGCN checkpoints are selected by validation domain accuracy, while GlycanAA checkpoints are selected by validation Macro-F1. For each seed, predictions from the five held-out folds are concatenated to give one out-of-fold prediction over all 13,209 records; Table~\ref{tab:Glycan}  reports Macro-F1 on the pooled OOF predictions, with mean and standard deviation taken over the three seeds rather than over folds.

\paragraph{Experimental Details for Case Study~\ref{case:ecg-normalization}}

For the targeted LVOLT/VCLVH experiments, we use the standard PTB-XL split \citep{wagner2020ptbxl,strodthoff2021ptbxl} of folds 1-8 train, fold 9 validation, fold 10 test. We construct balanced target-aware cohorts by pairing each positive ECG with one fold-matched negative control, excluding target-positive patients from the control pool and preferentially matching sex, device, site, and age. Using all 182 LVOLT and 200 VCLVH positives gives 758 unique ECGs, we split 606/78/74. We report macro AUROC/AUPRC over the two labels. For ECGFounder \citep{li2025ecgfounder} and \citet{weimann2024federated}, we freeze respectively 1024-dimensional exact-global-$z$ Net1D and 512-dimensional exact-$z$ ResNet-34 representations from three validation-selected backbones trained on this cohort. 
For the broader native settings, ECGFounder uses the released pretrained model with ten-fold patient-disjoint OOF evaluation on 21,799 usable PTB-XL records: each fold is test, the next cyclic fold validation, and the remaining eight folds train the matched affine readout. Metrics are pooled OOF macro AUROC/AUPRC. For these entries, the central value is the pooled OOF metric and $\pm$ is patient-cluster-bootstrap standard deviation. For Weimann-Conrad, we train the released central ResNet-34 on the PTB client using the released splits, yielding 16,667/1,198/4,488 train/validation/test records, for 100 epochs under the documented central-training configuration. We then freeze the 512-dimensional pooled representation and fit matched affine readouts with zero auxiliary input or standardized $\log\rho$; regularization is selected by validation macro AUROC. Final metrics are mask-aware macro averages over scored diagnoses, with $\pm$ denoting standard deviation across three independently trained seeds.

\paragraph{Experimental Details for Case Study~\ref{case:cwd}}
We compute the Choi-Williams distribution using the \texttt{tfrcw} implementation in the Time-Frequency Toolbox (TFTB) \citep{auger1998tftb}, with the number of frequency bins equal to the R-R interval length, kernel parameter $\sigma=1$, and odd Hamming time- and frequency-smoothing windows of lengths $\lfloor N/10\rfloor$ and $\lfloor N/4\rfloor$. Each R-R interval is standardized and converted to its analytic signal; after removing $3\%$ of CWD time columns from each boundary, we divide the remainder into five contiguous regions and record the maximum and minimum coefficient in each, repeating the procedure for the first-differenced waveform to obtain 20 CWD features. Although we exactly recovered the MIT-BIH healthy/arrhythmia portion of the dataset used by \citet{kayikcioglu2020time}, the released annotations do not determine their ST-elevation/depression sample-selection procedure. We therefore instead devised the discussed experiment using expert-annotated ST episodes from the European ST-T Database \citep{taddei1992european}. For the GE, the ST sample is taken $120$, $112$, $104$, or $100$\,ms after $R_i$ for heart rates $<100$, $100$-$110$, $110$-$120$, or $\geq120$\,bpm, respectively; the local baseline is the flattest 40-ms window in $[R_i-200\,\mathrm{ms},R_i-40\,\mathrm{ms}]$, and the recording reference $s_r^{\mathrm{ref}}$ is the median ST score over clean N-N intervals ending in the first 30\,s. The final cohort contains 68 subjects and 288 episodes (207 depression, 81 elevation); we retain 25 adjacent clean N-N intervals wholly within each episode and nearest its annotated apex, requiring R-R duration in $[0.25,2.0]$\,s. Evaluation is leave-one-subject-out, holding out all recordings of each subject together and averaging the 25 interval probabilities for each episode. Features are standardized using training-fold statistics only. The repaired arm appends $c(r,i)=s_i(r)-s_r^{\mathrm{ref}}$, while the matched control appends $|c(r,i)|$. We use $L_2$-regularized logistic regression ($C=1$) and 5-NN with Euclidean distance.

\paragraph{Experimental Details for Case Study~\ref{case:G-FNO}}
For the first experiment, we follow the released G-FNO protocol of Helwig et al.~\citep{helwig2023group}. The baseline is the original scalar-typed G-FNO-$p4m$; the retyped arm changes the input/output vorticity representations to the pseudoscalar action, while the GE arm retains the scalar pathway and appends the GE at the first regular-carrier cut. For the decaying-turbulence experiment, we generate initial conditions following Cao et al.~\citep{cao2025spectralrefiner} and their \texttt{torch-cfd} implementation using the McWilliams spectrum~\citep{mcwilliams1984emergence}: Gaussian white noise is spectrally filtered with density proportional to $[k(1+(k/k_{\mathrm{peak}})^4)]^{-1}$, the resulting streamfunction is normalized to unit kinetic energy, and vorticity is obtained spectrally. We generate $768$ base initial conditions with $k_{\mathrm{peak}}=4$ on a $64^2$ periodic grid and evolve each together with its sign reversal, giving $1{,}536$ trajectories, with $\nu=10^{-3}$, $\Delta t=10^{-3}$, no forcing or drag, and $T=0.5$. Base initial conditions are split $512/128/128$ into train/validation/test sets pre-pairing, so that each $\omega_0,-\omega_0$ pair remains within one split.

\paragraph{Experimental Details for Case Study~\ref{case:QGeoGNN}}
For QGeoGNN, we use the authors' $90/5/5$ all-column split and retain the released protocol. CIP assignments are computed in RDKit \citep{rdkit} by explicitly assigning stereochemistry and reading the atom-level \texttt{\_CIPCode}; assigned tetrahedral centers are mapped to $R$ or $S$, while unassigned or unsupported stereochemistry is left unassigned. For the GE arm, $c_i=(\mathbf 1_R,\mathbf 1_S,\mathbf 1_{\tau_i=+1},\mathbf 1_{\tau_i=-1})$ is zero for atoms without assigned tetrahedral stereochemistry and is injected immediately after the atom-feature encoder and before the first GIN layer through a zero-initialized bias-free $4\!\to\!128$ linear map, adding $512$ trainable parameters. The Retyped arm changes only the stored chiral-tag feature to the corresponding absolute CIP $R/S$ type.

\paragraph{Experimental Details for Case Study~\ref{case:magnet}}
 We follow the anomalous-Hall setup of Kipp et al.~\citep{kipp2024hall}, using the $\xi$-even target at $E_F=-1.5\,\mathrm{eV}$, $\Gamma=0.025\,\mathrm{eV}$, $t=1$, $\alpha_R=0.4\,\mathrm{eV}$, and $\lambda_{\mathrm{ex}}=1.4\,\mathrm{eV}$. Following their random-spin setup, we sample $s_A,s_B$ independently and uniformly on $S^2$ and apply the $12$ $C_{6v}$ spatial operations, including the identity, together with time reversal alone. We independently generate $2{,}048$ base configurations and, unlike the original protocol, split them $1{,}536/256/256$ into train/validation/test sets before augmentation so that symmetry-related copies cannot cross splits, yielding $19{,}968/3{,}328/3{,}328$ augmented examples. Table~\ref{tab:magnets} reports performance on the $3{,}328$ augmented held-out examples; we additionally evaluate on the $256$ untouched held-out base configurations as a robustness check. For Cho et al.~\citep{cho2026sparsitypromoting}, we reconstruct their published magnetic architecture, the parameter-matched control and GE each contain $3{,}560$ trainable parameters. For MagNet \citep{yuan2024magnet}, we follow the published architectural specification; its matched control and GE each contain $58{,}378$ parameters. 

\end{document}